\documentclass{article} 
\usepackage{iclr2027_conference,times}

\usepackage{amsmath,amsfonts,bm}

\def\eqref#1{equation~\ref{#1}}

\def\1{\bm{1}}

\DeclareMathAlphabet{\mathsfit}{\encodingdefault}{\sfdefault}{m}{sl}
\SetMathAlphabet{\mathsfit}{bold}{\encodingdefault}{\sfdefault}{bx}{n}

\usepackage{hyperref}
\usepackage{url}
\usepackage{amsfonts,amssymb,amsmath,amsthm}
\usepackage{bm}
\usepackage{bbm}
\usepackage{multirow}
\usepackage{rotating}
\usepackage{algpseudocode}
\usepackage{algorithm}
\usepackage{xcolor}
\usepackage{graphicx}
\usepackage{booktabs}
\usepackage[table]{xcolor}
\newcommand{\sd}[1]{{\tiny$\pm$#1}}
\usepackage{enumitem}
\usepackage{wrapfig}

  \definecolor{hdr}{RGB}{225,236,248}      
  \definecolor{grn}{RGB}{226,242,229}      
  \newcommand{\streamarrow}{\makebox[0.78\linewidth]{%
    \leaders\hrule height 0.12pt\hfill\kern-1pt\raisebox{-0.32ex}{\tiny$\blacktriangleright$}}}
\newcommand{\rst}[1]{\textsubscript{\tiny(#1)}}

\theoremstyle{definition}

\title{The Batch Size Can Be One: Test-Time Adaptation from Classifier Geometry}

\author{Jaeho Hwang$^{1}$ \qquad Yujin Oh$^{2}$ \qquad Juhyeon Shin$^{3}$ \\
	\textbf{Juhye Lee$^{2}$ \qquad Sungroh Yoon$^{3,*}$ \qquad Uiwon Hwang$^{2,*}$} \\
	$^1$Human-Centered Artificial Intelligence Research Institute, Ewha Womans University\\
	$^2$Division of Artificial Intelligence \& Software, Ewha Womans Univeristy\\
	$^3$Interdisciplinary Program in Artificial Intelligence, Seoul National University
}

\iclrfinalcopy 
\begin{document}

\maketitle
{\renewcommand{\thefootnote}{*}%
	\footnotetext{Corresponding authors (sryoon@snu.ac.kr, uiwon.hwang@ewha.ac.kr)}}
	
\begin{abstract}
Test-time adaptation updates a deployed model on an unlabeled test stream. In practice this stream often arrives one sample at a time, with the classes in no particular order. A single sample then provides limited information for estimating distribution-level statistics, and selecting samples by confidence can admit predictions that are wrong yet confident. A corruption, however, shifts the features while leaving the decision regions of the frozen classifier unchanged. We therefore propose Origin-recentered Test-Time Adaptation (OriTTA), which constructs fixed geometric targets from the source model and adapts selected features toward them. Seen from the origin of the feature space, the classes lean toward a common direction. We choose a new origin to bring class scores closer together while keeping it near the mean feature of one synthetic prototype per class. From this origin, prototype directions spread more evenly, and each class obtains an axis and an ideal point before any test data arrives. During adaptation, OriTTA aligns selected features with their predicted class axis and penalizes excessive radius, keeps one input per predicted class in memory to limit feature changes for classes absent from the stream, and resets the model when re-classifying this memory reveals collapse. Without tuning for each setting, OriTTA achieves the highest mean accuracy at batch size 1 on CIFAR-10-C, CIFAR-100-C and ImageNet-C in all sixteen combinations of four backbones, two protocols and two stream orders. On ImageNet-C fog, recent methods collapse in up to two thirds of one hundred runs, while OriTTA never does.
\end{abstract}

\section{Introduction}
\label{sec:intro}

In real-world deployment, a model may encounter test data from distributions that differ from its training data. Test-time adaptation (TTA) addresses this by updating the model on the unlabeled test stream itself, most commonly by minimizing the entropy of the model's own predictions with respect to its normalization parameters~\citep{wang2021tent}. Later methods refine which samples to learn from and how large a step to take~\citep{niu2022efficient, niu2023towards, lee2024entropy, hu2025beyond}.


Existing TTA methods face two difficulties in realistic test streams. First, many of them estimate normalization statistics from the current test batch, and these estimates become unreliable when samples arrive one at a time or in a non-shuffled class order \citep{niu2023towards}. A single sample then provides limited information for estimating distribution-level statistics, and when a class arrives as one long run, even a large batch is dominated by that class, so its statistics capture the class rather than the distribution shift. Second, methods that update the model with its own predictions risk reinforcing their own mistakes. Confidence-based selection \citep{niu2022efficient} filters out uncertain samples but can still admit confidently incorrect predictions, which subsequent updates then reinforce. When adaptation continues across a sequence of shifts \citep{wang2022continual}, these errors compound over the stream until predictions collapse onto a few classes \citep{press2023rdumb}.

Since the classifier is frozen during TTA, its decision regions stay unchanged and a corruption only displaces the features. The geometry of these fixed regions then indicates where the features of each class should lie, without depending on the test stream. Using this geometry directly is difficult, however, because of three properties of the classifier and its feature space. First, the classifier rows share a component that shifts every logit equally without changing predictions. Second, some feature directions affect neither predicted labels nor class probabilities. Third, class prototypes point toward a common direction when measured from the feature-space origin, making their angular arrangement sensitive to the choice of origin.

We therefore propose Origin-recentered Test-Time Adaptation (OriTTA), which places an ideal point for each class using the frozen classifier and synthetic class prototypes, and adapts selected features toward the ideal point of their predicted class. OriTTA centers the classifier rows and projects features onto their span and moves the origin to a point where class scores are close together while staying near the mean feature of one synthetic prototype per class. From this origin, prototype directions spread more evenly, and each class obtains an axis, its ideal point on the axis and a cone around it, all fixed before adaptation begins. On CIFAR-100, the fraction of prototypes inside the cone of their own class rises from 47\% to 99\%.




During adaptation, OriTTA applies angular and radial losses to the current test feature only if it lies inside the cone of its predicted class and near its ideal point. These losses encourage class-axis alignment and penalize excessive radius. A memory initialized with synthetic prototypes holds one input per predicted class and is updated with low-entropy test inputs. OriTTA periodically pulls stored features toward their ideal points to limit changes while their classes are absent from the stream. Reclassifying this memory detects concentration of predictions on a few classes and triggers restoration of the source model. Without per-setting tuning, OriTTA achieves the highest mean accuracy at batch size 1 on CIFAR-10-C, CIFAR-100-C and ImageNet-C in all sixteen combinations of four backbones, episodic and continual adaptation, and shuffled and class-blocked streams.

Our contributions are the following:
\begin{itemize}[leftmargin=1.2em,itemsep=1pt,topsep=2pt]
\item We propose OriTTA, which takes the target of adaptation from the frozen classifier, measured from a recentered origin computed before any test data arrives.
\item We design per-sample losses requiring neither batch statistics nor classifier softmax probabilities, and a collapse detector that evaluates prediction concentration across class-indexed memory slots.
\item We analyze classifier and prototype geometry across four source models and demonstrate accuracy gains from projection and recentering in the evaluated ablations.
\item We show that OriTTA achieves the highest mean accuracy at batch size 1 in all sixteen settings and never collapses in 100 runs on ImageNet-C fog.
\end{itemize}
\section{Related Work}
\label{sec:related}

\paragraph{Single-sample and non-i.i.d.\ test-time adaptation.}
TENT~\citep{wang2021tent} introduced entropy minimization over normalization affine parameters and EATA~\citep{niu2022efficient} added sample filtering and anti-forgetting. Single-sample adaptation was addressed by augmentation in MEMO~\citep{zhang2022memo} and SITA~\citep{khurana2021sita}. SAR~\citep{niu2023towards} formalized the \emph{wild} setting, which combines batch size 1, online label shift and mixed shifts, and proposed reliable-sample selection with sharpness-aware updates, followed by DeYO~\citep{lee2024entropy}, ReCAP~\citep{hu2025beyond} and SaTeen~\citep{sateen}. Temporally correlated streams are handled by adjusting normalization statistics (NOTE~\citep{gong2022note}, DELTA~\citep{zhao2023delta}, UnMIX-TNS~\citep{tomar2024unmix}), by class-balanced memories (NOTE, RoTTA~\citep{yuan2023robust}) or by output correction (LAME~\citep{boudiaf2022parameter}). OriTTA selects and adapts test features by their position relative to a fixed ideal point for each class, without requiring batch statistics.

\paragraph{Continual adaptation and collapse.}
CoTTA~\citep{wang2022continual}, RMT~\citep{dobler2023robust} and PETAL~\citep{brahma2023probabilistic} counter drift with teacher-student schemes and weight restoration. RDumb~\citep{press2023rdumb}
uses fixed-interval source resets, while PeTTA~\citep{bohdal2024persistent} detects collapse by measuring divergence from source statistics. OriTTA reclassifies a memory with one input per predicted class and restores the source weights when predictions concentrate on a few classes. This trigger does not directly use the class frequencies of the incoming stream.

\paragraph{Feature geometry.}
Neural collapse~\citep{papyan2020prevalence} has motivated geometric approaches to adaptation. TCA~\citep{ni2025maintaining} promotes uniformity among pseudo-label centroids, while NEO~\citep{murphy2026neo} recenters target embeddings without updating the model. Both estimate class structure from test batches, whereas OriTTA derives its origin and ideal points from the source model before any test data arrive. Further related work is discussed in Appendix~\ref{app:related}.

\section{Method}
\label{sec:method}
\begin{figure}[t!]
    \centering
    \includegraphics[width=\textwidth]{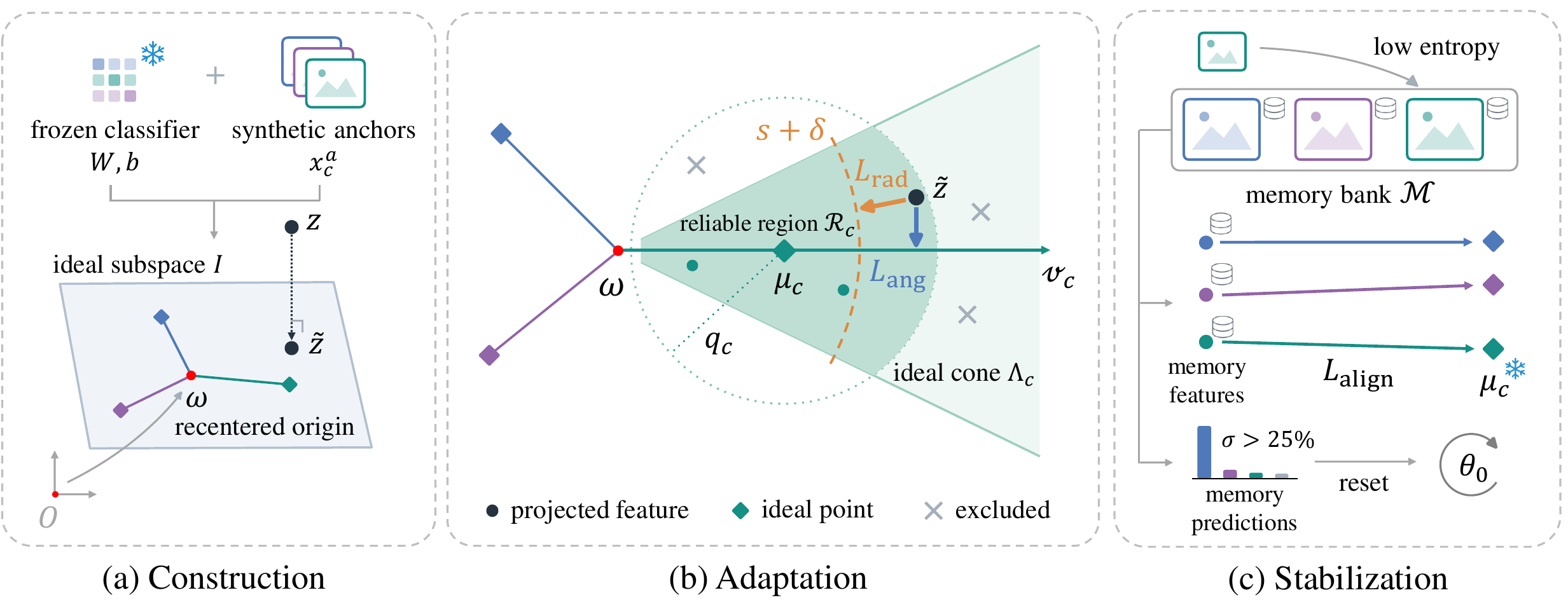}
    \caption{Overview of OriTTA. (a) The frozen classifier and synthetic anchors define a fixed reference frame. (b) Reliable projected features undergo angular alignment and radial regularization. (c) Stored images are re-encoded for alignment to fixed ideal points, while concentrated current predictions over the memory trigger resets. Snowflakes denote frozen quantities.}
    \label{overview}
    \vspace{-5pt}
\end{figure}
\subsection{Notation and the Recentered Frame}
\label{sec:frame}

\paragraph{Why recenter.}
Much of the recent geometric analysis of deep features measures distances and angles from the origin of the feature space: cosine similarity, feature-norm statistics, contrastive objectives on the unit sphere and the simplex-ETF picture of neural collapse all take the origin as the reference point. However, the origin is a property of the parameterization rather than of the data. After a ReLU the features occupy the non-negative orthant, so the origin sits at its corner. After a LayerNorm they are shifted by the affine parameters. In either case the center of the data has no reason to coincide with it. Angles measured from a point that is not the center are dominated by the offset, not by the arrangement of the classes. We therefore recover a center of the source feature distribution and define every quantity of our method in the frame recentered at that point.

\paragraph{The classifier.}
Let $f_\theta:\mathcal{X}\to\mathbb{R}^D$ be the backbone up to its penultimate layer, producing the feature $z=f_\theta(x)$ of an input $x$, and let the classifier be linear: for class $c\in\{1,\cdots,K\}$, $w_c\in\mathbb{R}^D$ is the $c$-th row of $W$, $b_c$ its bias, and $\ell_c(z)=w_c\cdot z+b_c$ its logit, so that $\ell(z)=Wz+b$ and $\hat y(x)=\arg\max_c\ell_c(f_\theta(x))$. Write $\bar w=\frac1K\sum_c w_c$ for the mean row and
\begin{equation}
  v_c=\frac{w_c-\bar w}{\|w_c-\bar w\|},\qquad
  \beta_c=\frac{b_c-\bar b}{\|w_c-\bar w\|},\qquad c=1,\dots,K,
  \label{eq:V}
\end{equation}
collect the rows into $V\in\mathbb{R}^{K\times D}$ and $\beta\in\mathbb{R}^K$. Then the affine hyperplane
\begin{equation}
  H_c=\{\,z\in\mathbb{R}^D:\ v_c\cdot z+\beta_c=0\,\}=\{\,z:\ \ell_c(z)=\bar\ell(z)\,\},\qquad \bar\ell=\frac1K\sum_c\ell_c
\end{equation}
is the locus on which class $c$ scores exactly the average logit and $|v_c\cdot z+\beta_c|$ is the Euclidean distance of $z$ from it, since $\ell_c(z)-\bar\ell(z)=(w_c-\bar w)\cdot z+(b_c-\bar b)$. The unit normals $v_c$ span the row space of the centered classifier, whose dimension we denote $r=\operatorname{rank}V=\min(K-1,D)$.

\paragraph{Recentered origin.}
We define the center of the feature space as the point that is simultaneously closest, in the least-squares sense, to all $K$ hyperplanes $H_c$. i.e.\ the point at which the classifier is
maximally ambiguous:
\begin{equation}
  \mathcal{J}(z)=\sum_{c=1}^{K}\operatorname{dist}(z,H_c)^2=\|Vz+\beta\|^2 .
  \label{eq:J}
\end{equation}
$\mathcal{J}$ always admits a minimizer (not unique in general), but it constrains $z$ only through its position relative to the hyperplanes and says nothing about where the features themselves lie.

A center must also be \emph{where the source features live}, and in test-time adaptation that location has to be recovered without the source data. We obtain it from the classifier itself. For each class $c$ we synthesize a single \emph{anchor image} induced by the cross-entropy,
\begin{equation}
  x_{c}^{a}=\arg\min_{x\in[0,1]^{C\times H\times W}}\ \mathrm{CE}\big(\ell(f_{\theta_0}(x)),\,c\big),
  \label{eq:anchor}
\end{equation}
obtained by gradient descent on the pixels from a random initialization. This is the same cross-entropy the source model was trained with, minimized over the input rather than the weights, so $x_c^a$ is the input the model is most confident belongs to class $c$, and it depends on the source model alone. Their feature centroid $\bar{z}^{a}=\frac1K\sum_{c=1}^{K} f_{\theta_0}(x_{c}^{a})$ is a source-free estimate of where class-prototypical features gather. The \emph{recentered origin} combines the two requirements,
\begin{equation}
  \omega=\arg\min_{z}\ \|Vz+\beta\|^2+\lambda\|z-\bar{z}^{a}\|^2
   =(V^{\!\top}V+\lambda I)^{-1}\big(\lambda\bar{z}^{a} -V^{\!\top}\beta\big),
  \qquad \lambda=\frac{K}{100},
  \label{eq:origin}
\end{equation}
the point closest to the decision hyperplanes among those near the prototype centroid. The regularizer is weighted at one hundredth of a single hyperplane constraint, so the ridge acts as $K/100$ additional constraints pulling toward $\bar{z}^{a}$. Appendix~\ref{app:lambda} discusses the choice.

\paragraph{Ideal subspace.}
Let $\mathcal{I}=\omega+\operatorname{span}\{v_1,\dots,v_K\}$ be an ideal subspace which is the $r$-dimensional affine subspace through the recentered origin spanned by the class normals, and let
\begin{equation}
  \tilde{z}:=\text{proj}_{\mathcal{I}}(z)=\omega+P\,(z-\omega),\qquad P=V^{+}V,
  \label{eq:proj}
\end{equation}
be the orthogonal projection onto it, where $V^{+}$ is the Moore-Penrose pseudoinverse of $V$ so that $P$ is the orthogonal projector onto $\operatorname{span}\{v_1,\dots,v_K\}$. Every logit \emph{difference} is preserved by the projection, since $\ell_i(z)-\ell_j(z)=\ell_i(\tilde z)-\ell_j(\tilde z)$ for all $i,j$. Hence $\hat y$ and every softmax probability are unchanged, and the component of $z$ orthogonal to $\mathcal{I}$ is invisible to the decision. Thus, we work with $\tilde{z}$ throughout and for a class $c$ decompose $u=\tilde z-\omega$ into a radial and a perpendicular part,
\begin{equation}
  t_c=u\cdot v_c,\qquad d_{\perp,c}=\sqrt{\|u\|^2-t_c^2},
  \label{eq:decomp}
\end{equation}
so that $\|\tilde z-\mu_c\|^2=(t_c-s)^2+d_{\perp,c}^2$ splits the distance to the ideal point into a radius error and an angular error. We note that if $K-1\ge D$, then $\mathcal{I}$ is the whole feature space.

\paragraph{Ideal points.}
Displacing $\omega$ along the normal $v_c$ increases $\ell_c-\bar\ell$ faster than any other direction, so the axis $\{\omega+t\,v_c:\ t>0\}$ is the natural home of class $c$. We place the \emph{ideal point} of class $c$ at a common radius $s$ along its axis,
\begin{equation}
  \mu_c=\omega+s\,v_c,\qquad
  s=\operatorname{med}_{c\in\{1,\cdots,K\}}\ \big\|\operatorname{proj}_{\mathcal{I}}(z_{c}^{a})-\omega\big\|,
  \label{eq:ideal}
\end{equation}
where $\operatorname{med}$ denotes the median over classes and $z_{c}^{a}$ is the feature vector of $x_{c}^{a}$. Hence, $s$ is the distance at which the model's own prototypical inputs typically land and it is clear that $\mu_{c}\in\mathcal{I}$. Each anchor depends on its random initialization, but the frame does not. Regenerating the IN-RN50 anchors from ten other seeds moves $\omega$ by at most $0.03\,s$ and changes $s$ by at most $0.4\%$ (see Appendix~\ref{app:anchors} for more details).

\paragraph{Ideal cone.}

Within $\mathcal{I}$ we define the region in which the class-$c$ distribution is expected to lie as a cone around the class axis,
\begin{equation}
  \Lambda_c=\big\{\tilde z\in\mathcal{I}:\ t_c>t_{\min},\ \ d_{\perp,c}<\kappa\,t_c\big\},
  \qquad \kappa=\kappa_0\sqrt{r},\ \ \kappa_0=0.11,\ \ t_{\min}=0.5 .
  \label{eq:cone}
\end{equation}
The half-angle grows with $\sqrt{r}$ because the perpendicular offset of a class-$c$ feature is spread over the $r-1$ directions orthogonal to its axis and concentrates at a radius proportional to $\sqrt{r}$. The ratio $\kappa_0$ of that spread to the prototype radius is the same across backbones whose feature scales differ by an order of magnitude, so a single constant serves all of them. The lower bound $t_{\min}$ excludes the immediate neighborhood of $\omega$, where the classifier is ambiguous by construction.

\paragraph{Reliable region.}
A test feature enters the per-sample losses only if it is \emph{reliable} for its predicted class $c=\hat y(x)$: it must lie in the ideal cone $\Lambda_c$ and, among the cone samples of class $c$, be among the closest to $\mu_c$,
\begin{equation}
  \mathcal{R}_c = \big\{\tilde z\in\Lambda_c:\ \|\tilde z-\mu_c\|\le q_c\big\},
  \label{eq:reliable}
\end{equation}
where $q_c$ is a running estimate of the $0.2$-quantile of $\|\tilde z-\mu_c\|$ over the samples that have entered $\Lambda_c$ as class $c$ so far. A fixed radius would not serve, since the distances depend on the corruption.

\subsection{Losses on the Reliable Region}
\label{sec:losses}
Two losses act on the current feature, and only when it is reliable, $\tilde z\in\mathcal{R}_{\hat y}$.
Since they update normalization parameters shared by every input, moving the reliable features
toward their ideal geometry also moves the features of the remaining inputs, on which no loss is
computed.
\paragraph{Angular loss.}
With $c=\hat y(x)$, we treat the negative perpendicular distances to the $K$ class axes as logits and apply a cross-entropy toward the predicted class,
\begin{equation}
  L_{\mathrm{ang}}(\tilde z)=-\log\frac{\exp(-d_{\perp,c}/\tau)}{\sum_{j=1}^{K}\exp(-d_{\perp,j}/\tau)},
  \qquad \tau=\sqrt{s}.
  \label{eq:ang}
\end{equation}
This is a cross-entropy toward the predicted class, with the logits replaced by negative distances to the class axes. The temperature scales with $s$ because $d_\perp$ does.

\paragraph{Radial loss.}
A reliable feature that lies far from the recentered origin is pulled inward with a one-sided hinge,
\begin{equation}
  L_{\mathrm{rad}}(\tilde z)=\frac{1}{r}\max\big(0,\ \|u\|-(s+\delta)\big)^2,
  \,\, \delta=\tfrac13\,d_{\mathrm{nn}},\,\,
   d_{\mathrm{nn}}=\operatorname{med}_{i\in\{1,\cdots,K\}}\ \min_{j\ne i}\|\mu_i-\mu_j\|.
  \label{eq:rad}
\end{equation}
It acts only on the radius and leaves the angle to $L_{\mathrm{ang}}$ (see Appendix~\ref{app:lossablation} for the ablation study).

\subsection{Memory Bank and Alignment Loss}
\label{sec:bank}

The two losses above act on whichever class the current input belongs to. On a class-blocked stream, many classes go unseen for thousands of steps, and nothing prevents the model from drifting away from them. We keep the most recent real input of every class in a memory bank and periodically pull each one back to its ideal point, so that no class is forgotten.

\paragraph{Memory bank.}
The memory bank $\mathcal{M}=\{m_1,\dots,m_K\}$ holds one input per class, the most recently admitted one. It is initialized with the anchor images, $m_c\leftarrow x_c^{a}$, and an anchor is replaced by the first real input admitted for its class. The current input replaces $m_{\hat y}$ if its predictive entropy is below $0.4\ln K$. 
\paragraph{Alignment loss.}
Every $\lceil 30/B\rceil$ steps, we take $\min\{20,K\}$ classes $\mathcal S_t$ in round-robin order, pass their bank inputs through the current backbone, and pull each projected feature to its ideal point:
\begin{equation}
  L_{\mathrm{align}}=\frac{1}{|\mathcal S_t|}\sum_{c\in\mathcal S_t}\frac{1}{r}\,
  \big\|\operatorname{proj}_{\mathcal{I}}\big(f_\theta(m_c)\big)-\mu_c\big\|^2 .
  \label{eq:align}
\end{equation}
where $B$ is the batch size. The bank sample is used rather than the anchor because the anchors drift off $\mu_c$ under adaptation as well, and constraining a mixture of anchor and real features would push the real ones to the far side of $\mu_c$. The target is the seed label rather than a fresh prediction, since re-predicting it would let a drifting model relabel its own bank (Appendix~\ref{app:abl_align} for the discussion).

\subsection{Collapse Detection and Reset}
\label{sec:reset}
Online adaptation can still collapse onto a few classes. We detect this by periodically re-classifying the bank with the current model and measuring the \emph{bank share}
\begin{equation}
  \sigma=\frac1K\max_{j}\ \big|\{c:\ \hat y(m_c)=j\}\big|,
  \label{eq:share}
\end{equation}
the fraction of the bank sent to a single class. Since the bank holds one representative of every class, $\sigma=1/K$ for a healthy model and rises only under collapse, independently of the label order of the stream. When $\sigma>0.25$, the model is restored to its source weights and the bank to its anchors (details in Appendix~\ref{app:reset}).

\medskip
The complete objective is
\begin{equation}
  L_{\mathrm{total}}=L_{\mathrm{ang}}+L_{\mathrm{align}}+2\,L_{\mathrm{rad}}.
  \label{eq:total}
\end{equation}
Nothing above is estimated from source data, and nothing is tuned per setting: every quantity is either derived from the source weights, through the classifier and its anchors, or a single constant shared across all datasets, backbones and stream orders.
\section{Experiments}
\begin{table}[t]
\centering\footnotesize
\setlength{\tabcolsep}{3pt}
\caption{IN-RN50 under the four protocols. Per-corruption standard
deviations are in Appendix~\ref{app:full}. The best result in each column is shown in bold and the second best is underlined.}
\label{tab:in_gn}
\resizebox{\linewidth}{!}{%
\begin{tabular}{>{\columncolor{hdr}}c l ccccccccccccccc | c}
\toprule
\rowcolor{hdr} & episodic & gaus & shot & impu & defo & glas & moti & zoom & snow & fros & fog & brit & cont & elas & pixe & jpeg & Average \\
\midrule
& TENT   & 12.35 & 20.22 & 13.91 & 13.16 & 6.46 & 18.07 & 18.23 & 16.00 & 20.65 & 1.48 & 69.80 & 42.04 & 6.22 & 49.38 & 53.75 & 24.11\sd{0.40} \\
& EATA   & 29.49 & 31.55 & 30.23 & 17.93 & 19.32 & 30.45 & 32.62 & 42.73 & 41.21 & 45.55 & 69.40 & 43.66 & 32.98 & 52.13 & 54.02 & 38.22\sd{0.21} \\
& SAR    & 28.63 & 31.01 & 29.40 & 18.46 & 15.23 & 28.82 & 30.43 & 44.75 & 44.39 & 33.60 & 71.95 & 44.61 & 13.03 & 47.83 & 56.24 & 35.89\sd{0.11} \\
& DeYO   & 40.87 & 43.33 & 42.08 & 22.07 & 23.69 & 38.63 & 28.12 & 52.26 & 50.21 & 18.88 & 73.19 & 51.27 & 42.97 & 57.31 & 58.54 & 42.89\sd{1.95} \\
& ReCAP  & \underline{42.07} & \underline{44.35} & \underline{43.17} & 24.69 & 25.05 & 41.73 & \underline{39.64} & \underline{54.67} & 52.16 & \underline{46.54} & \underline{73.33} & 53.52 & 41.54 & \underline{60.27} & \underline{60.20} & \underline{46.86}\sd{1.52} \\
& SaTeen & \textbf{43.20} & \textbf{45.44} & \textbf{44.25} & \underline{25.80} & \underline{26.25} & \textbf{42.44} & 26.68 & \textbf{55.29} & \underline{52.69} & 19.30 & \textbf{73.34} & \textbf{54.00} & \underline{48.71} & \textbf{60.77} & 60.13 & 45.22\sd{2.20} \\
\rowcolor{grn}\cellcolor{hdr}\multirow{-7}{*}{\rotatebox[origin=c]{90}{\textit{shuffled}}} & OriTTA & 41.11 & 43.15 & 42.22 & \textbf{33.53} & \textbf{31.93} & \underline{42.38} & \textbf{46.33} & 54.32 & \textbf{53.27} & \textbf{59.04} & 72.80 & \underline{53.79} & \textbf{49.64} & 59.77 & \textbf{60.62} & \textbf{49.59}\sd{0.05} \\
\midrule
& TENT   & 15.59 & 22.05 & 17.00 & 13.21 & 7.87 & 18.40 & 18.53 & 17.01 & 20.43 & 1.50 & 69.84 & 42.12 & 6.22 & 49.45 & 53.80 & 24.87\sd{0.80} \\
& EATA   & 25.27 & 25.71 & 25.28 & 15.68 & 14.93 & 21.35 & 22.70 & 33.11 & 32.90 & \underline{37.47} & 65.32 & 34.82 & 18.24 & 38.49 & 50.43 & 30.78\sd{0.19} \\
& SAR    & 28.53 & 31.11 & 29.44 & 18.56 & 15.14 & 28.83 & 30.53 & 44.16 & 44.36 & 33.30 & 71.93 & 44.67 & 13.87 & 47.73 & 56.22 & 35.89\sd{0.15} \\
& DeYO   & 40.89 & 43.18 & 42.19 & 21.84 & 23.44 & 38.64 & 33.34 & 52.08 & 50.35 & 18.67 & 73.19 & 51.26 & 42.93 & 57.44 & 58.51 & 43.20\sd{1.74} \\
& ReCAP  & \underline{42.13} & \underline{44.41} & \underline{43.39} & 24.72 & 24.72 & 41.78 & \underline{39.92} & \underline{54.62} & 52.25 & 35.69 & \textbf{73.44} & \underline{53.63} & 42.67 & \underline{60.32} & \textbf{60.22} & \underline{46.26}\sd{2.10} \\
& SaTeen & \textbf{43.13} & \textbf{45.48} & \textbf{44.31} & \underline{25.58} & \underline{25.96} & \textbf{42.34} & 27.82 & \textbf{55.28} & \textbf{52.74} & 13.48 & \underline{73.41} & \textbf{54.00} & \textbf{48.85} & \textbf{61.00} & \underline{60.14} & 44.90\sd{1.84} \\
\rowcolor{grn}\cellcolor{hdr}\multirow{-7}{*}{\rotatebox[origin=c]{90}{\textit{class-blocked}}} & OriTTA & 40.79 & 42.55 & 41.81 & \textbf{33.32} & \textbf{31.09} & \underline{41.83} & \textbf{45.80} & 53.69 & \underline{52.67} & \textbf{57.92} & 72.50 & 53.12 & \underline{48.79} & 58.73 & 60.05 & \textbf{48.98}\sd{0.10} \\
\midrule\midrule
\rowcolor{hdr} & & gaus & shot & impu & defo & glas & moti & zoom & snow & fros & fog & brit & cont & elas & pixe & jpeg & Average \\
\rowcolor{hdr} & continual & \multicolumn{15}{c}{\streamarrow} & \\
\midrule
& TENT   & 12.35 & 0.74 & 0.22 & 2.16 & 0.26 & 0.22 & 0.31 & 0.57 & 0.18 & 0.20 & 1.48 & 0.31 & 0.10 & 0.16 & 0.79 & 1.34\sd{0.28} \\
& EATA   & 29.49 & 29.30 & 25.44 & 12.94 & 14.77 & \underline{18.41} & \underline{23.08} & \underline{26.36} & \underline{27.85} & \underline{31.02} & \underline{54.08} & \underline{24.44} & \underline{21.43} & \underline{32.12} & \underline{38.49} & \underline{27.28}\sd{1.91} \\
& SAR    & 28.63 & 41.23 & 44.45 & 15.66 & \underline{19.02} & 9.66 & 10.66 & 5.55 & 3.04 & 1.21 & 6.33 & 1.39 & 0.48 & 0.98 & 1.94 & 12.68\sd{0.59} \\
& DeYO   & 40.87 & 48.80 & \underline{47.90} & 10.76 & 13.74 & 13.96 & 6.48 & 1.53 & 0.73 & 0.44 & 1.76 & 0.26 & 0.20 & 0.19 & 0.23 & 12.52\sd{0.55} \\
& ReCAP  & \underline{42.03} & \textbf{49.33} & \textbf{48.41} & \textbf{24.02} & 3.50 & 3.91 & 2.70 & 3.55 & 3.85 & 5.13 & 6.98 & 4.96 & 1.72 & 4.31 & 4.65 & 13.94\sd{8.28} \\
& SaTeen & \textbf{43.20} & \underline{48.90} & 47.61 & 19.60 & 1.17 & 0.19 & 0.19 & 0.16 & 0.12 & 0.11 & 0.12 & 0.11 & 0.11 & 0.11 & 0.11 & 10.79\sd{0.67} \\
\rowcolor{grn}\cellcolor{hdr}\multirow{-7}{*}{\rotatebox[origin=c]{90}{\textit{shuffled}}} & OriTTA & 41.09 & 43.78 & 38.40 & \underline{22.63} & \textbf{21.81} & \textbf{36.39} & \textbf{35.26} & \textbf{50.72} & \textbf{48.64} & \textbf{42.97} & \textbf{63.05} & \textbf{39.22} & \textbf{40.62} & \textbf{33.09} & \textbf{54.44} & \textbf{40.81}\sd{1.02} \\
\midrule
& TENT   & 15.59 & 1.08 & 0.27 & 2.73 & 0.36 & 0.27 & 0.38 & 0.63 & 0.19 & 0.22 & 1.97 & 0.41 & 0.10 & 0.17 & 0.97 & 1.69\sd{0.39} \\
& EATA   & 25.27 & 11.04 & 7.49 & 6.19 & 5.24 & 5.26 & 6.21 & 4.71 & 5.98 & 5.36 & 12.14 & 2.29 & 4.36 & 5.29 & 6.47 & 7.55\sd{1.19} \\
& SAR    & 28.53 & 41.14 & 44.47 & 15.76 & \underline{19.17} & 11.52 & \underline{11.18} & \underline{15.79} & \underline{18.37} & \underline{18.04} & \underline{34.59} & \underline{20.35} & \underline{5.68} & \underline{18.72} & \underline{20.27} & \underline{21.57}\sd{10.86} \\
& DeYO   & 40.89 & 48.73 & \underline{47.86} & 11.55 & 11.50 & \underline{11.82} & 6.89 & 1.88 & 0.77 & 0.47 & 2.02 & 0.27 & 0.20 & 0.20 & 0.24 & 12.35\sd{0.84} \\
& ReCAP  & \underline{42.15} & \textbf{49.28} & \textbf{48.43} & 18.17 & 3.23 & 0.81 & 0.31 & 0.26 & 0.17 & 0.15 & 0.24 & 0.14 & 0.13 & 0.15 & 0.17 & 10.92\sd{1.02} \\
& SaTeen & \textbf{43.13} & \underline{48.78} & 47.57 & \underline{24.45} & 3.17 & 3.59 & 2.04 & 2.06 & 0.95 & 0.24 & 0.16 & 0.10 & 0.12 & 0.10 & 0.10 & 11.77\sd{1.78} \\
\rowcolor{grn}\cellcolor{hdr}\multirow{-7}{*}{\rotatebox[origin=c]{90}{\textit{class-blocked}}} & OriTTA & 40.74 & 44.45 & 39.26 & \textbf{25.80} & \textbf{24.91} & \textbf{33.45} & \textbf{32.44} & \textbf{52.02} & \textbf{48.77} & \textbf{45.97} & \textbf{66.77} & \textbf{45.32} & \textbf{39.96} & \textbf{41.61} & \textbf{48.27} & \textbf{41.98}\sd{0.78} \\
\bottomrule
\end{tabular}}
\vspace{-5mm}
\end{table}

\subsection{Experimental Settings}
\label{sec:settings}

\paragraph{Datasets and backbones.}
We evaluate on CIFAR-10-C, CIFAR-100-C and ImageNet-C, on all 15
corruptions at severity 5 (10,000 images per corruption on CIFAR, 50,000 on ImageNet), with
WideResNet-28-10 on CIFAR-10, WideResNet-40-2 trained with AugMix on CIFAR-100, and ResNet-50 with GroupNorm and ViT-B/16 on ImageNet, following
SAR~\citep{niu2023towards}. We refer to the four settings as C10, C100, IN-RN50 and IN-ViT.

\paragraph{Protocols.}
All methods adapt online at batch size 1 with one update per image. In the \emph{episodic}
protocol the model is restored to its source weights at each corruption and in the \emph{continual}
protocol it adapts through the fifteen corruptions in sequence without any reset. Each protocol
is run on a \emph{shuffled} stream and a \emph{class-blocked} one, in which each class arrives as a
contiguous block. The stream is redrawn for every seed and shared across methods, and we report
accuracy (\%) averaged over ten seeds (2021--2030), with the standard deviation as a subscript.

\paragraph{Baselines.}
We compare against TENT~\citep{wang2021tent}, EATA~\citep{niu2022efficient},
SAR~\citep{niu2023towards}, DeYO~\citep{lee2024entropy}, ReCAP~\citep{hu2025beyond} and
SaTeen~\citep{sateen}, using their official code and published batch-size-1 settings
(see Appendix~\ref{app:baselines} for more details).

\paragraph{Implementation.}
OriTTA updates only the normalization affine parameters with Adam, at a learning rate of
$2.5\times10^{-4}$ for LayerNorm and $2.5\times10^{-4}/8$ for BatchNorm and GroupNorm (see Appendix~\ref{app:optim} for the discussion of Adam and SGD). The anchors
and the frame are computed once per backbone, \textbf{before} the stream starts, and are never updated. All constants are listed in Appendix~\ref{app:details}.

\subsection{Main Results}
\label{sec:main}

\paragraph{IN-RN50 results.}
Table~\ref{tab:in_gn} reports the representative setting under all four protocols. In the episodic protocol, OriTTA leads the strongest baseline by about $2.7$ on either stream order, and the one-standard-deviation intervals of the two methods do not overlap. On the three noise corruptions, which come first in the sequence and which every method handles from the source weights, OriTTA trails the best baseline by two to three points. On defocus and glass blur, on zoom blur and on fog, it leads by five to twelve points, and by twenty on fog under the class-blocked stream. The standard deviation of OriTTA is the smallest, at $0.05$ and $0.1$, which indicates high stability of OriTTA. In the continual protocol the picture changes in kind rather than in degree. Every baseline collapses somewhere in the sequence: TENT within the first two corruptions, DeYO, ReCAP and SaTeen by the fourth or fifth, and the most resilient baseline, EATA on the shuffled stream and SAR on the class-blocked one, retains three quarters and three fifths of its episodic accuracy respectively. OriTTA gives up about nine points relative to its own episodic accuracy on the shuffled stream and seven on the class-blocked one, and ends at $1.5$ and $1.9$ times the accuracy of the best baseline.

\paragraph{Stability over one hundred seeds.}
\begin{wraptable}{r}{0.39\linewidth}
\vspace{-\intextsep}
\centering\scriptsize
\setlength{\tabcolsep}{3pt}
\caption{One hundred seeds ($2001$--$2100$) on the three corruptions with the least stable
baselines for IN-RN50, episodic, shuffled. \emph{Collapse} is the fraction of seeds in which
accuracy of fog corruption falls below $5\%$.}
\label{tab:seeds100}
\begin{tabular}{l c c c c}
\toprule
\rowcolor{hdr} & zoom & elastic & fog & collapse \\
\midrule
TENT   & 18.2\sd{0.6} & 6.1\sd{0.5} & 1.5\sd{0.1} & 100\% \\
EATA   & 32.9\sd{1.6} & 33.5\sd{1.9} & 46.2\sd{2.1} & 0\% \\
SAR    & 30.4\sd{0.3} & 14.1\sd{1.4} & 33.2\sd{2.2} & 0\% \\
DeYO   & 33.1\sd{6.8} & 43.0\sd{0.5} & 21.0\sd{25.2} & 66\% \\
ReCAP  & 40.4\sd{3.6} & 45.1\sd{8.2} & 39.2\sd{25.4} & 33\% \\
SaTeen & 27.2\sd{9.1} & 48.8\sd{0.4} & 23.6\sd{27.7} & 62\% \\
\rowcolor{grn}OriTTA & \textbf{46.5}\sd{0.2} & \textbf{49.7}\sd{0.3} & \textbf{59.0}\sd{0.2} & 0\% \\
\bottomrule
\end{tabular}
\vspace{-2mm}
\end{wraptable}
In fact, the large seed-to-seed deviations of DeYO, ReCAP and SaTeen in Table~\ref{tab:in_gn}
come almost from three corruptions, fog, zoom, and elastic corruption (Table~\ref{tab:full_rn50}). On the other twelve their deviations stay under a point. On these three corruptions the methods behave differently from run to run, and on
fog some runs fail outright, with accuracy falling close to zero within the corruption. All three
methods minimize prediction entropy, which reinforces the model's own errors once it becomes
confidently wrong. $10$ seeds are too few to characterize failures of this kind, so we repeat the
three corruptions over $100$ seeds each (Table~\ref{tab:seeds100}). The deviations of DeYO,
ReCAP and SaTeen remain large on all three. Crucially, on fog they collapse to below $5\%$ accuracy in
$66\%$, $33\%$ and $62\%$ of the seeds respectively. Across the same three hundred runs OriTTA never
collapses, its deviation stays at $0.2$--$0.3$.

\paragraph{Other datasets and backbones.}
\begin{table}[h]
\centering\footnotesize
\setlength{\tabcolsep}{2.5pt}
\caption{Mean accuracy over the 15 corruptions on C10, C100, and IN-ViT. Full results
are in Appendix~\ref{app:full}.}
\label{tab:main}
\resizebox{\linewidth}{!}{%
\begin{tabular}{l ||c|c|c|c c|| c|c|c|c c ||c|c|c|c}
\toprule
\rowcolor{hdr} & \multicolumn{4}{c}{C10} & & \multicolumn{4}{c}{C100} & & \multicolumn{4}{c}{IN-ViT} \\
\cmidrule(lr){2-5}\cmidrule(lr){7-10}\cmidrule(lr){12-15}
\rowcolor{hdr} & \multicolumn{2}{c}{episodic} & \multicolumn{2}{c}{continual} & & \multicolumn{2}{c}{episodic} & \multicolumn{2}{c}{continual} & & \multicolumn{2}{c}{episodic} & \multicolumn{2}{c}{continual} \\
\rowcolor{hdr} & shuffled & class-blk & shuffled & class-blk & & shuffled & class-blk & shuffled & class-blk & & shuffled & class-blk & shuffled & class-blk \\
\midrule\midrule
TENT   & 46.70\sd{2.77} & 43.52\sd{6.84} & 15.66\sd{0.77} & 11.04\sd{0.78} & & 25.83\sd{1.73} & 18.18\sd{1.18} & 1.24\sd{0.11} & 1.17\sd{0.12} & & 50.95\sd{0.13} & 51.14\sd{0.23} & 47.85\sd{2.96} & 48.85\sd{4.09} \\
EATA   & \underline{56.66}\sd{0.08} & 56.63\sd{0.05} & \underline{56.96}\sd{0.23} & \underline{56.94}\sd{0.17} & & \underline{54.49}\sd{0.27} & \underline{48.98}\sd{0.44} & \underline{29.22}\sd{1.37} & 18.05\sd{0.68} & & 62.18\sd{0.07} & 45.63\sd{0.10} & 59.90\sd{0.13} & 9.06\sd{0.31} \\
SAR    & 56.33\sd{0.03} & 56.96\sd{0.10} & 55.99\sd{0.87} & 55.49\sd{2.39} & & 44.58\sd{0.24} & 47.10\sd{0.56} & 20.36\sd{7.07} & \underline{31.24}\sd{6.53} & & 57.79\sd{0.32} & 44.58\sd{0.42} & 59.20\sd{0.10} & 50.41\sd{0.92} \\
DeYO   & 51.92\sd{2.94} & 57.34\sd{6.14} & 44.42\sd{3.45} & 37.86\sd{6.62} & & 10.49\sd{1.02} & 12.48\sd{1.56} & 1.89\sd{0.55} & 1.91\sd{0.26} & & 62.91\sd{0.62} & 62.93\sd{1.09} & 60.70\sd{0.54} & 60.60\sd{0.34} \\
ReCAP  & 48.70\sd{0.98} & \underline{58.80}\sd{0.69} & 36.58\sd{6.71} & 43.13\sd{9.33} & & 2.09\sd{0.15} & 7.48\sd{1.58} & 1.06\sd{0.03} & 1.28\sd{0.13} & & \underline{65.00}\sd{0.66} & 64.02\sd{1.56} & 61.37\sd{0.82} & 61.13\sd{0.75} \\
SaTeen & 37.35\sd{2.55} & 52.90\sd{8.45} & 14.27\sd{1.78} & 27.62\sd{7.73} & & 1.92\sd{0.10} & 4.94\sd{0.81} & 1.04\sd{0.01} & 1.33\sd{0.23} & & 64.68\sd{0.81} & \underline{64.97}\sd{1.20} & \underline{61.62}\sd{0.32} & \underline{61.64}\sd{0.56} \\
\rowcolor{grn}OriTTA & \textbf{73.38}\sd{0.08} & \textbf{68.96}\sd{1.51} & \textbf{73.74}\sd{0.17} & \textbf{66.04}\sd{1.43} & & \textbf{62.16}\sd{0.09} & \textbf{61.69}\sd{0.33} & \textbf{58.77}\sd{0.28} & \textbf{57.03}\sd{0.29} & & \textbf{65.58}\sd{0.05} & \textbf{66.23}\sd{0.10} & \textbf{62.91}\sd{0.96} & \textbf{62.00}\sd{0.30} \\
\bottomrule
\end{tabular}}
\vspace{-5mm}
\end{table}
Table~\ref{tab:main} summarizes C10, C100 and IN-ViT by their mean accuracy. OriTTA is the most
accurate method in each of the twelve columns, though the margin depends on how the baselines
fare. On the two CIFAR benchmarks, whose WideResNet normalizes with BatchNorm, several baselines
break down even in the episodic protocol, and in the continual one the strongest of them reaches
only about half of OriTTA's accuracy on C100. The lead ranges from $8$ to $30$ points. On IN-ViT
the picture is different. LayerNorm makes a transformer considerably more robust to single-sample
adaptation, most baselines stay close to their episodic accuracy even in the continual protocol,
and OriTTA's lead narrows to $0.4$ to $1.3$ points.

\paragraph{Continual TTA methods at batch size 1.}
\begin{wraptable}{r}{0.40\linewidth}
\vspace{-\intextsep}
\centering\scriptsize
\setlength{\tabcolsep}{3.5pt}
\caption{CoTTA at batch size 1 under our protocol, shuffled stream.}
\label{tab:ctta_bs1}
\begin{tabular}{l cccc}
\toprule
 & C10 & C100 & IN-RN50 & IN-ViT \\
\midrule
episodic  & 10.31 & 1.14 & 0.12 & 0.12 \\
continual & 10.30 & 0.80 & 0.35 & 0.19 \\
\bottomrule
\end{tabular}
\vspace{-5mm}
\end{wraptable}
The failure at batch size 1 extends to methods designed for continual adaptation. CoTTA~\citep{wang2022continual},
a representative continual method, falls to chance level on every setting when run at batch size 1
(Table~\ref{tab:ctta_bs1}), and the more recent TCA~\citep{ni2025maintaining} likewise reports low
accuracy for continual methods, including its own, at batch size 1.

\subsection{What Each Operation of the Frame Contributes}
\label{sec:analysis}

The frame of Section~\ref{sec:frame} is built from three operations on the source model: centering
the classifier rows, projecting onto the ideal subspace $\mathcal{I}$, and measuring from the
recentered origin $\omega$. Each removes one kind of structure that the classifier does not read,
and how much of that structure a backbone carries can be measured before any test data arrives
(Table~\ref{tab:frame}). This gives a prediction for each operation: it should matter where the
structure it removes is large, and not elsewhere. Table~\ref{tab:ablation} checks the prediction by
removing one operation at a time.

\begin{table}[t]
\centering\footnotesize
\setlength{\tabcolsep}{3.5pt}
\caption{Geometry of the source model under the three operations. \emph{Rows dev.} is the deviation of the classifier rows' mutual cosines from an equiangular
arrangement, $-1/(K-1)$, before and after centering. \emph{pairwise deviation} is the same quantity
for the prototypes $z^a_c$, measured from the origin, after projection, and from $\omega$.
\emph{Axis} is the median cosine between prototype $c$ and its own class axis $v_c$, and
\emph{in cone} the fraction of prototypes the ideal cone admits.}
\label{tab:frame}
\resizebox{\linewidth}{!}{%
\begin{tabular}{l ccc c cc c ccc c cc cc}
\toprule
& \multicolumn{3}{c}{shift} & & \multicolumn{2}{c}{rows dev. $\downarrow$} &
& \multicolumn{3}{c}{pairwise deviation $\downarrow$} & & \multicolumn{2}{c}{axis $\uparrow$} &
\multicolumn{2}{c}{in cone $\uparrow$} \\
\cmidrule(r){2-4}\cmidrule(lr){6-7}\cmidrule(lr){9-11}\cmidrule(l){13-16}
& $\|\omega\|$ & $\|\omega\|/s$ & $\|P\omega\|/s$ & & $w_c$ & $v_c$
& & at $0$ & proj. & at $\omega$ & & at $0$ & at $\omega$ & at $0$ & at $\omega$ \\
\midrule
CIFAR-10-C       &  4.77 & 0.96 & 0.05 & & 0.07 & 0.07 & & 0.58 & 0.38 & 0.36 & & 0.79 & 0.82 & 0.00 & 0.00 \\
CIFAR-100-C      & 12.11 & 0.92 & 0.57 & & 0.10 & 0.10 & & 0.52 & 0.36 & \textbf{0.16} & & 0.67 & \textbf{0.81} & 0.47 & \textbf{0.99} \\
ImageNet-C, RN50 &  6.92 & 0.80 & 0.35 & & 0.65 & \textbf{0.03} & & 0.31 & 0.15 & \textbf{0.07} & & 0.54 & \textbf{0.58} & 1.00 & 1.00 \\
ImageNet-C, ViT  & 33.62 & 0.42 & 0.42 & & 0.06 & 0.06 & & 0.21 & 0.21 & \textbf{0.09} & & 0.69 & \textbf{0.77} & 1.00 & 1.00 \\
\bottomrule
\vspace{-7mm}
\end{tabular}}
\end{table}

\begin{table}[t]
\centering\footnotesize
\setlength{\tabcolsep}{3pt}
\caption{Removing one operation of the frame, episodic. Top: losses computed in the full feature
space, IN-RN50 (episodic, shuffled). Bottom: origin moved from $\omega$ to $0$ and projection to $\mathcal{I}$, with $s$, $\mu_c$, and $\delta$ recomputed, C100 (episodic, class-blocked).}
\label{tab:ablation}
\resizebox{\linewidth}{!}{%
\begin{tabular}{ll ccccccccccccccc| c}
\toprule
\rowcolor{hdr} & & gaus & shot & impu & defo & glas & moti & zoom & snow & fros & fog & brit & cont & elas & pixe & jpeg & Average \\
\midrule
\multirow{3}{*}{\rotatebox{90}{\scriptsize IN-RN50}}
& OriTTA        & 41.11 & 43.15 & 42.22 & 33.53 & 31.93 & 42.38 & 46.33 & 54.32 & 53.27 & 59.04 & 72.80 & 53.79 & 49.64 & 59.77 & 60.62 & \textbf{49.59} \\
& no projection & 38.93 & 40.83 & 39.50 & 31.02 & 27.94 & 39.57 & 43.71 & 53.65 & 52.58 & 48.34 & 72.72 & 52.62 & 45.98 & 57.63 & 60.35 & 47.02 \\
& $\Delta$      & $-2.18$ & $-2.32$ & $-2.72$ & $-2.51$ & $-3.99$ & $-2.81$ & $-2.62$ & $-0.67$ & $-0.69$ & $\mathbf{-10.70}$ & $-0.08$ & $-1.17$ & $-3.66$ & $-2.14$ & $-0.27$ & $-2.57$ \\
\midrule
\multirow{3}{*}{\rotatebox{90}{\scriptsize C100}}
& OriTTA        & 48.07 & 52.29 & 51.20 & 70.98 & 54.74 & 69.73 & 70.81 & 63.13 & 63.30 & 58.71 & 71.93 & 63.51 & 62.84 & 64.83 & 59.24 & \textbf{61.69} \\
& origin at $0$ & 44.78 & 49.69 & 48.92 & 69.94 & 53.38 & 68.74 & 69.25 & 61.74 & 61.54 & 56.09 & 71.23 & 60.11 & 61.74 & 60.29 & 58.45 & 59.73 \\
& $\Delta$      & $-3.29$ & $-2.60$ & $-2.28$ & $-1.04$ & $-1.36$ & $-0.99$ & $-1.56$ & $-1.39$ & $-1.76$ & $-2.62$ & $-0.70$ & $-3.40$ & $-1.10$ & $\mathbf{-4.54}$ & $-0.79$ & $-1.96$ \\
\bottomrule
\end{tabular}}
\vspace{-4mm}
\end{table}
\paragraph{Centering.}
The classifier rows share a component $\bar w$ that adds the same amount to every logit and plays no
part in any decision. On IN-RN50 it carries $81\%$ of a row's norm, and the raw rows are far from
equiangular (deviation $0.65$). Once it is removed the axes $v_c$ form the tightest equiangular set
of the four settings ($0.03$). On the other three backbones $\bar w$ is negligible and centering
changes nothing.

\paragraph{Projection.}
Logit differences depend on a feature only through its component in $\mathcal{I}$, so the
remaining $D-r$ directions carry nothing the classifier reads, while a corruption can still
displace a feature far along them. This is large on IN-RN50, where $1{,}049$ of $2{,}048$
dimensions are invisible, and absent on IN-ViT, where $P=I$. Removing the projection on IN-RN50 costs $2.6$ points, and the loss falls where the argument places it. Without the projection the cone admits samples on the strength of variation the classifier does not use, and the radial term measures a radius it cannot read.

\paragraph{Recentering.}
A subspace fixes directions but not a point, and the cone, the ideal points and every distance the
method computes need one. We take it from the model's own prototypes, and the effect is that from
$\omega$ the prototypes lie along their axes: the cosine to the own axis rises in all four settings,
and the fraction the ideal cone admits changes where it can, from $47\%$ to $99\%$ on C100. Only
the row-space component $P\omega$ acts, and the fraction of the shift it carries grows with $r/D$. Thus, recentering should matter on C100, where it changes which prototypes the cone admits, and little on IN-RN50, where the cone admits all of them either way. Moving the origin back to $0$ costs $2.0$ points on the C100 class-blocked stream. On C10, this shift is only $0.05\,s$, and the row space spans just nine of the 640 feature dimensions. The small ridge weight $\lambda=K/100$ limits the pull toward the centroid of ten prototypes, keeping $\omega$ close to the classifier hyperplanes (see Appendix~\ref{app:lambda}).

\subsection{Collapse Detection}
\label{sec:detector}
\begin{table}[t]
\centering\footnotesize
\setlength{\tabcolsep}{2.4pt}
\caption{SAR with its own recovery, SAR with OriTTA's detector and reset in its place (SAR+RT), and
OriTTA, on IN-RN50, class-blocked. RDumb~\citep{press2023rdumb} restores EATA to its source
weights every $64{,}000$ images. The subscript is the mean number of resets per run, shown only
where nonzero. The total is given with the average.}
\label{tab:reset}
\resizebox{\linewidth}{!}{%
\begin{tabular}{l ccccccccccccccc | c}
\toprule
\rowcolor{hdr} \textit{episodic} & gaus & shot & impu & defo & glas & moti & zoom & snow & fros & fog & brit & cont & elas & pixe & jpeg & Average \\
\midrule
SAR     & 28.53 & 31.11 & 29.44 & 18.56 & 15.14 & 28.83 & 30.53 & 44.16 & 44.36 & 33.30 & 71.93 & 44.67 & 13.87 & 47.73 & 56.22 & 35.89\rst{0} \\
SAR+RT  & 28.54 & 31.15 & 29.42 & 18.56 & 15.13 & 28.83 & 30.54 & 44.15 & 44.32 & 33.31 & 71.93 & 44.67 & 20.47\rst{1} & 47.71 & 56.17 & 36.33\rst{1} \\
\rowcolor{grn}OriTTA  & 40.79 & 42.55 & 41.81 & 33.32 & 31.09 & 41.83 & 45.80 & 53.69 & 52.67 & 57.92 & 72.50 & 53.12 & 48.79 & 58.73 & 60.05 & \textbf{48.98}\rst{0} \\
\midrule
\midrule
\rowcolor{hdr} \textit{continual} & gaus & shot & impu & defo & glas & moti & zoom & snow & fros & fog & brit & cont & elas & pixe & jpeg & Average \\
\midrule
RDumb   & 25.44 & 22.09\rst{1} & 22.14\rst{1} & 15.51\rst{1} & 14.97 & 20.73\rst{1} & 21.44\rst{1} & 28.41\rst{1} & 22.80\rst{1} & 37.82 & 66.62\rst{1} & 32.52\rst{1} & 15.86\rst{1} & 34.16 & 48.97\rst{1} & 28.64\rst{11} \\
SAR     & 28.53 & 41.14 & 44.47 & 15.76 & 19.17 & 11.53\rst{8.5} & 11.17\rst{7.1} & 15.79\rst{6.1} & 18.37\rst{36.8} & 18.04\rst{3564} & 34.59\rst{8} & 20.35\rst{144} & 5.68\rst{3740} & 18.72\rst{33.1} & 20.27\rst{1476} & 21.57\rst{9025} \\
SAR+RT  & 28.54 & 41.14 & 44.46 & 15.74 & 19.17 & 23.90\rst{1} & 33.83 & 43.49 & 46.49 & 53.29 & 61.32 & 46.88 & 17.68\rst{2} & 47.43 & 54.29 & 38.51\rst{3} \\
\rowcolor{grn}OriTTA  & 40.74 & 44.45 & 39.26\rst{0.8} & 25.80\rst{0.2} & 24.91\rst{0.8} & 33.45 & 32.44\rst{0.9} & 52.02\rst{0.1} & 48.77 & 45.97\rst{0.6} & 66.77\rst{0.1} & 45.32\rst{0.1} & 39.96\rst{1} & 41.61 & 48.27 & \textbf{41.98}\rst{4.6} \\
\bottomrule
\end{tabular}}
\vspace{-4mm}
\end{table}
We transplant the detector into SAR~\citep{niu2023towards} in place of its own recovery, which resets the model when the moving average of its entropy falls below $0.2$. SAR's loss, filter and optimizer are unchanged (Table~\ref{tab:reset}). The two mechanisms differ in what they see. In the episodic protocol SAR falls to $13.9\%$ on elastic transform, yet its recovery never fires. OriTTA's detector, which re-classifies a class-balanced bank, fires once there and recovers seven points. In the continual protocol SAR's recovery restores the model saved at the start of the current corruption, which is already degraded, and it fires about nine thousand times per run. With OriTTA's detector, SAR resets three times per run and gains seventeen points, from $21.6\%$ to $38.5\%$ with its loss unchanged. This accuracy is higher than the $35.9\%$ that SAR reaches in the episodic protocol. A reset on a fixed schedule is less effective: RDumb~\citep{press2023rdumb} resets eleven times per run and reaches $28.6\%$, below SAR+RT with three, because it discards adaptation at every interval rather than only when collapse occurs. Moreover, this accuracy is lower than $30.78\%$ that EATA reaches in the episodic protocol. The detector therefore transfers, and the remaining gap to OriTTA is what the frame and the losses contribute. SAR+RT in the continual protocol is also more accurate than SAR in the episodic one: adaptation carries information from one corruption to the next, and a detector that resets only when collapse occurs keeps it, whereas a reset at every corruption boundary would discard it. On OriTTA itself the detector never fires in the episodic protocol and fires $4.6$ times per run in the continual one, against fifteen corruption boundaries (see Appendix~\ref{app:reset_counts} for more details).

\section{Discussion}
\label{sec:discussion}

\textbf{Conclusion.}
We have argued that single-sample test-time adaptation should take its reference from the source
model rather than from the test stream, and that the origin of the feature space is the wrong point
to take it from. OriTTA recovers a frame from the source classifier and one synthetic prototype per
class, adapts inside it with losses that require neither a batch nor the model's confidence, and
detects collapse with a signal that does not depend on the label order of the stream. It is the
most accurate method in every setting we evaluate, its accuracy varies by a fraction of a point
across seeds where recent methods vary by several. Each operation of the
frame removes a kind of structure the classifier does not read, and how much a backbone carries of
it can be measured before adaptation. The accuracy each operation contributes follows that
measurement, which is what allows a single set of constants to serve every dataset, backbone and
stream order.

\textbf{Limitations.}
The frame works best when the ideal subspace is neither the whole feature space nor a small part
of it. When $K-1\ge D$, as for ViT-B/16, the projection has nothing to remove and OriTTA's lead
over baselines that LayerNorm already keeps stable narrows to about a point. When $K$ is small, as
on CIFAR-10, the recentering is nearly invisible to the method and the cone rule, derived for large
row spaces, becomes too narrow. The construction also assumes a linear head whose centered rows are
close to equiangular, which holds for the models we examine but would not for a head trained on a
long-tailed distribution. Finally, our evaluation covers synthetic corruptions only, and natural shifts such as ImageNet-R or ImageNet-Sketch remain to be tested.

\textbf{Outlook.}
Restoring the most recent state the detector judged healthy, rather than the source weights, would
retain adaptation that a full reset discards. The cone rule can be recalibrated for small row spaces, and the frame extends to any head that is linear in a learned representation, including dense prediction.

\subsection*{Reproducibility statement}
The released code contains the method, the anchor-generation script, the anchors used for all reported results, and the stream construction. The baselines were run from the official implementations released with SaTeen, and the streams they see are produced by the same function. Appendix~\ref{app:details}
lists every constant of the method together with the per-setting quantities it derives from the
source model, the learning-rate and batch-size conventions, the anchor-generation procedure, the
baseline configurations and the two adjustments we made to run them under our protocol, and the
hardware. The test streams are a deterministic function of the seed and are shared by all methods,
as described in Appendix~\ref{app:streams}. All results in the paper are averages over the seeds
$2021$--$2030$ unless stated otherwise. Code will be released upon publication.


\subsubsection*{Acknowledgments}
This work was supported by Global - Learning \& Academic research institution for Master's$\cdot$PhD students, and Postdocs (G-LAMP) Program of the National Research Foundation of Korea (NRF) grant funded by the Ministry of Education (No. RS-2025-25442252), the National Research Foundation of Korea (NRF) grant funded by the Korean government (MSIT) (Grant No. RS-2022-NR069041, RS-2025-00561169), Institute of Information \& communications Technology Planning \& Evaluation(IITP) under the Leading Generative AI Human Resources Development(IITP-2027-RS-2026-25546026) grant funded by the Korea government(MSIT).

\bibliography{iclr2027_conference}
\bibliographystyle{iclr2027_conference}

\newpage
\appendix

\section{Implementation Details}
\label{app:details}

\subsection{Backbones and Adapted Parameters}
Table~\ref{tab:app_backbones} lists the four settings. On CIFAR we use the RobustBench checkpoints for common corruptions, WideResNet-28-10 on CIFAR-10 and WideResNet-40-2 trained with AugMix on CIFAR-100, which normalize their input internally and receive raw $[0,1]$ pixels.
On ImageNet we use the ResNet-50 with GroupNorm and the ViT-B/16 (AugReg, ImageNet-21k pretrained,
ImageNet-1k fine-tuned) checkpoints from \texttt{timm}, following SAR~\citep{niu2023towards}, with
ImageNet mean and standard deviation applied externally. For every backbone OriTTA adapts only the
affine parameters ($\gamma$, $\beta$) of the normalization layers. On ViT-B/16 the final LayerNorm
before the classifier is kept frozen. The classifier and all convolutional and attention weights
are frozen. The feature $z$ is the input of the classifier, taken through a forward hook.

\begin{table}[h]
\centering\footnotesize
\caption{Settings. $D$ is the feature dimension, $K$ the number of classes and
$r=\min(K-1,D)$ the dimension of the ideal subspace.}
\label{tab:app_backbones}
\begin{tabular}{l l l c c c c}
\toprule
Setting & Dataset & Backbone & Norm & $D$ & $K$ & $r$ \\
\midrule
C10     & CIFAR-10-C   & WideResNet-28-10 & BatchNorm & 640  & 10   & 9   \\
C100    & CIFAR-100-C  & WideResNet-40-2  & BatchNorm & 128  & 100  & 99  \\
IN-RN50 & ImageNet-C   & ResNet-50        & GroupNorm & 2048 & 1000 & 999 \\
IN-ViT  & ImageNet-C   & ViT-B/16         & LayerNorm & 768  & 1000 & 768 \\
\bottomrule
\end{tabular}
\end{table}

\subsection{Anchor Generation and Its Algorithm}
\label{app:anchors}
For each class $c$ one anchor image $x_c^a$ is synthesized from the source model alone by minimizing $\mathrm{CE}(\ell(f_{\theta_0}(x)),c)+10^{-3}\,\mathrm{TV}(x)$ over the pixels, where $\mathrm{TV}$ is the total variation of the image. The image is initialized from Gaussian noise with standard deviation $0.05$ in normalized pixel space and optimized with Adam at learning rate $0.1$ for $1{,}000$ iterations under cosine decay, with the image clamped to the valid pixel range after every step. Classes are optimized in batches of $50$ with seed $0$.

\paragraph{Sensitivity to the anchor draw.}
The anchors depend on their random initialization, so we regenerate the IN-RN50 anchors from ten
further seeds and rebuild the frame from each set (Table~\ref{tab:app_regen}). Individual anchors
change substantially. A regenerated anchor lies a median of $0.75\,s$ from the released anchor of
its class, and seen from $\omega$ the two have a cosine of $0.72$. The frame does not follow them.
Across the ten sets $\omega$ moves by at most $0.032\,s$, $s$ changes by at most $0.4\%$, and the
geometry of Table~\ref{tab:frame} is unchanged at the reported precision. The frame is built
from the centroid and the median of $K$ anchors, which average out the variation of any single
one, and $\omega$ inherits almost all of the centroid's shift of $0.033\,s$. This is half the shift
caused by raising $\lambda$ from $10$ to $100$, which leaves accuracy unchanged
(Appendix~\ref{app:lambda}).
\begin{table}[h]
\centering\footnotesize
\setlength{\tabcolsep}{6pt}
\caption{Frame of IN-RN50 built from the released anchors and from anchors regenerated with ten
other seeds (mean and standard deviation). Distances are in units of the released $s$.}
\label{tab:app_regen}
\begin{tabular}{l c c}
\toprule
\rowcolor{hdr} & released & regenerated \\
\midrule
$\|\omega\|$ &$ 6.93$ & $6.96 \pm 0.02$ \\
$s$ & $8.63$ & $8.62 \pm 0.02$ \\
$\delta$ & $3.45$ & $3.44 \pm 0.01$ \\
shift of $\omega$ & -- & $0.031 \pm 0.001$ \\
shift of $\bar z^a$ & -- & $0.033 \pm 0.001$ \\
shift of one anchor (median) & -- & $0.75 \pm 0.01$ \\
pairwise deviation at $\omega$ & $0.07$ & $0.07 \pm 0.00$ \\
axis at $\omega$ &$0.58$ & $0.58 \pm 0.00$ \\
in cone &$1.00$ & $1.00 \pm 0.00$ \\
\bottomrule
\end{tabular}
\end{table}

\begin{algorithm}[h!]
\caption{Offline: anchors and frame (once per backbone, no data)}
\label{alg:offline}
\begin{algorithmic}[1]
\Require source model $f_{\theta_0}$ with linear head $(W,b)$, $K$ classes, feature dimension $D$
\State $\bar w \gets \frac1K\sum_c w_c$,\quad $\bar b \gets \frac1K\sum_c b_c$
\State $v_c \gets (w_c-\bar w)/\|w_c-\bar w\|$,\quad $\beta_c \gets (b_c-\bar b)/\|w_c-\bar w\|$ \Comment{Eq.~\ref{eq:V}}
\State $V \gets [v_1;\dots;v_K]$,\quad $P \gets V^{+}V$,\quad $r \gets \operatorname{rank}V=\min(K-1,D)$
\For{$c=1,\dots,K$} \Comment{anchor synthesis, Eq.~\ref{eq:anchor}}
   \State $x \gets$ Gaussian noise ($\sigma=0.05$), clamped to the pixel range
   \For{$1000$ iterations}
    \State $x \gets \mathrm{Adam}\big(x,\ \nabla_x [\mathrm{CE}(\ell(f_{\theta_0}(x)),c) + 10^{-3}\,\mathrm{TV}(x)]\big)$, lr $0.1$, cosine decay; clamp $x$
   \EndFor
   \State $x^a_c \gets x$;\quad $z^a_c \gets f_{\theta_0}(x^a_c)$
\EndFor
\State $\bar z^a \gets \frac1K\sum_c z^a_c$,\quad $\lambda \gets K/100$
\State $\omega \gets (V^{\top}V+\lambda I)^{-1}(\lambda \bar z^a - V^{\top}\beta)$ \Comment{Eq.~\ref{eq:origin}}
\State $s \gets \operatorname{med}_c \|P(z^a_c-\omega)\|$ \Comment{Eq.~\ref{eq:ideal}}
\State $\mu_c \gets \omega + s\,v_c$ for all $c$
\State $d_{\mathrm{nn}} \gets \operatorname{med}_i \min_{j\ne i}\|\mu_i-\mu_j\|$,\quad $\delta \gets 0.333\,d_{\mathrm{nn}}$,\quad $\tau \gets \sqrt{s}$,\quad $\kappa \gets 0.11\sqrt{r}$
\Ensure $\{x^a_c\}$, $\omega$, $P$, $\{v_c\}$, $\{\mu_c\}$, $s$, $\delta$, $\tau$, $\kappa$
\end{algorithmic}
\end{algorithm}

\subsection{Derived Quantities}
\label{app:derived}
Table~\ref{tab:app_derived} gives, for each setting, the quantities that Section~\ref{sec:frame}
derives from the source classifier and the anchors. All are computed once before the stream starts
and never updated.

\begin{table}[h]
\centering\footnotesize
\caption{Quantities derived from the source model. $s$ is the prototype radius, $d_{\mathrm{nn}}$
the median nearest-neighbor distance between ideal points, $\delta=d_{\mathrm{nn}}/3$ the radial
slack, $\tau=\sqrt{s}$ the temperature of $L_{\mathrm{ang}}$, $\kappa=0.11\sqrt{r}$ the cone slope,
and $\lambda=K/100$ the ridge weight in Eq.~\ref{eq:origin}.}
\label{tab:app_derived}
\begin{tabular}{l c c c c c c c c}
\toprule
Setting & $\|\omega\|$ & $\|P\omega\|/s$ & $s$ & $d_{\mathrm{nn}}$ & $\delta$ & $\tau$ & $\kappa$ & $\lambda$ \\
\midrule
C10     & 4.77  & 0.05 & 4.95  & 7.01  & 2.33  & 2.23 & 0.33 & 0.1 \\
C100    & 12.11 & 0.57 & 13.21 & 15.09 & 5.02  & 3.63 & 1.09 & 1   \\
IN-RN50 & 6.92  & 0.35 & 8.62  & 10.35 & 3.45  & 2.94 & 3.48 & 10  \\
IN-ViT  & 33.62 & 0.42 & 79.86 & 87.26 & 29.06 & 8.94 & 3.05 & 10  \\
\bottomrule
\end{tabular}
\end{table}

\subsection{Constants}
\label{app:constants}
Every remaining constant is shared by all settings, protocols and stream orders
(Table~\ref{tab:app_constants}). They were selected on IN-RN50 and are not tuned per setting; the reset threshold $\sigma_{\max}$ was set so that it is not met by two classes merging on a ten-class bank (Appendix~\ref{app:reset_counts}). Learning rates are the only backbone-specific input and are set by
normalization type: $2.5\times10^{-4}/8$ for BatchNorm and GroupNorm, and $2.5\times10^{-4}$ for
LayerNorm.

\begin{table}[h]
\centering\footnotesize
\caption{Constants of the method, shared across all settings.}
\label{tab:app_constants}
\begin{tabular}{l l l}
\toprule
& Value & Role \\
\midrule
$\kappa_0$          & $0.11$ & cone slope per $\sqrt{r}$ (Eq.~\ref{eq:cone}) \\
$t_{\min}$          & $0.5$  & lower bound on the radial coordinate in the cone \\
trust ratio         & $0.2$  & quantile that defines the reliable region (Eq.~\ref{eq:reliable}) \\
warm-up             & $20$   & cone samples per class before the quantile engages \\
$\eta$              & $0.02$ & rate of the quantile tracker (Appendix~\ref{app:quantile}) \\
$w_{\mathrm{ang}}, w_{\mathrm{align}}, w_{\mathrm{rad}}$ & $1, 1, 2$ & loss weights (Eq.~\ref{eq:total}) \\
entropy margin      & $0.4\ln K$ & bank admission threshold \\
snap period, classes & $30$ frames, $20$ & schedule of $L_{\mathrm{align}}$ (Eq.~\ref{eq:align}) \\
probe period        & $200$ frames & how often the bank is re-classified \\
$\sigma_{\max}$     & $0.25$ & reset threshold on the bank share (Eq.~\ref{eq:share}) \\
grace               & $1000$ frames & minimum spacing between resets \\
Adam $(\beta_1,\beta_2)$ & $(0.9, 0.999)$ & no weight decay \\
\bottomrule
\end{tabular}
\end{table}

\subsection{Reliable Region and Quantile Tracking}
\label{app:quantile}
For each class $c$ the reliable region of Eq.~\ref{eq:reliable} uses a radius $q_c$ that tracks the
$0.2$-quantile of $\|\tilde z-\mu_c\|$ over the samples that have entered the cone $\Lambda_c$ with
prediction $c$. Until $20$ such samples have arrived, all of them are treated as reliable and their
distances are collected. $q_c$ is then initialized as the $0.2$-quantile of these twenty distances
(the fourth smallest). Afterwards, each new cone sample of class $c$ updates the radius
multiplicatively,
\begin{equation}
  q_c \leftarrow \begin{cases}
    q_c\,(1-0.8\,\eta) & \text{if } \|\tilde z-\mu_c\|\le q_c,\\
    q_c\,(1+0.2\,\eta) & \text{otherwise},
  \end{cases}
  \qquad \eta=0.02,
\end{equation}
a stochastic approximation of the quantile under which exactly a fraction $0.2$ of samples pass in
equilibrium. Because the update is driven only by samples of class $c$, the radius of a class
that is absent from the stream does not move. The radii are part of the adaptation state and are
cleared on reset.

\subsection{Memory Bank, Alignment and Reset Schedule}
\label{app:reset}
The bank holds one input per class and is initialized with the anchor images. Admission is by
predictive entropy below $0.4\ln K$, evaluated on the input's logits before the current update. The admitted input replaces the bank entry of its predicted class, and when a batch admits several
inputs of the same class the one closest to $\mu_c$ is kept. The geometric gate is not used for
admission because at $K=1000$ it admits too few samples to fill the bank. Admission by entropy can let a confidently mislabeled input into the bank. The detector does not measure this contamination directly but the drift that produces it: as the model moves toward the classes it favors, entries admitted earlier are re-mapped onto them and $\sigma$ rises.

$L_{\mathrm{align}}$ is evaluated every $\lceil 30/B\rceil$ steps on the next $\min\{20,K\}$ classes
in round-robin order, re-passing their bank inputs through the current backbone with gradient
enabled so that the loss adjusts the backbone rather than the stored inputs.

The bank is re-classified without gradient every $\lceil 200/B\rceil$ steps, which is also the
logging cadence. A reset fires when $\sigma>0.25$ and at least $1000$ frames have passed since
initialization or the previous reset. It restores the source affine parameters, re-creates the
optimizer (discarding Adam's moment estimates), re-seeds the bank with the anchors, and clears the
quantiles $q_c$ and their warm-up counters. No smoothing or persistence is applied to $\sigma$. It is recomputed from the whole bank at every probe. Its firing rate in every setting is reported in Appendix~\ref{app:reset_counts}.

\subsection{Batch Sizes Above One}
\label{app:batch}
All schedules of the method are defined in frames, and a schedule of $n$ frames becomes $\lceil n/B\rceil$ steps at batch size $B$, so that the snap, the probe and the grace period occur at approximately the same frame intervals of the stream at any batch size. The learning rate scales as $\sqrt{B}$, the rule appropriate for Adam
\citep{malladi2022sdes}. A linear rule over-adapts at $B\ge 16$ and triggers a collapse reset within
each corruption, which the detector catches with $\sqrt{B}$ no reset is needed. Admission, the quantile
updates and the logging counters are applied per sample within a batch.

\subsection{Test Streams}
\label{app:streams}
For every corruption the stream is a permutation of its images produced by a single function that
both OriTTA and the baseline runner call with the same arguments, so that the two see
byte-identical streams for a given seed. The shuffled stream is \texttt{torch.randperm} under a
generator seeded with the seed. The class-blocked stream permutes the order in which the classes
appear and, within each class block, the order of its images, using the same generator, so that a
new seed redraws both. All fifteen corruptions of a run use the same seed. The ten seeds
$2021$--$2030$ therefore constitute ten independent repetitions of both the stream and the method.
We note that in an earlier version of our code the class-blocked stream ignored the seed and used
a fixed label-sorted order. All class-blocked results in the paper were produced after this was
corrected.

\subsection{Main Algorithm}
\label{app:algorithms}

\begin{algorithm}[h]
\caption{Online adaptation (one stream, batch size $B$)}
\label{alg:online}
\begin{algorithmic}[1]
\Require outputs of Algorithm~\ref{alg:offline}; source affine parameters $\theta_0$
\State $\theta \gets \theta_0$;\quad $m_c \gets x^a_c$,\ $q_c \gets \varnothing$,\ $N_c \gets 0$ for all $c$;\quad $n \gets 0$,\ $n_{\mathrm{rst}} \gets 0$
\For{each batch $x$ of the stream}
   \State $z \gets f_\theta(x)$;\quad $\hat y \gets \arg\max_c \ell_c(z)$;\quad $\tilde z \gets \omega + P(z-\omega)$ \Comment{predict}
   \State $u \gets \tilde z-\omega$;\quad $t_c \gets u\cdot v_c$;\quad $d_{\perp,c} \gets \sqrt{\|u\|^2-t_c^2}$ \Comment{Eq.~\ref{eq:decomp}}
   \State $\mathcal{L} \gets 0$
   \For{each sample $i$ in the batch, with $c=\hat y_i$}
      \If{$t_c > t_{\min}$ \textbf{and} $d_{\perp,c} < \kappa\,t_c$} \Comment{cone, Eq.~\ref{eq:cone}}
         \If{\Call{QuantileGate}{$c,\ \|\tilde z_i-\mu_c\|$}} \Comment{Eq.~\ref{eq:reliable}}
            \State $\mathcal{L} \gets \mathcal{L} + L_{\mathrm{ang}}(\tilde z_i) + 2\,L_{\mathrm{rad}}(\tilde z_i)$
         \EndIf
      \EndIf
   \EndFor
   \If{step $\bmod \lceil 30/B\rceil = 0$} \Comment{alignment, Eq.~\ref{eq:align}}
      \State $\mathcal{L} \gets \mathcal{L} + \frac{1}{|\mathcal S_t|}\sum_{c\in\mathcal S_t} \frac1r\|P(f_\theta(m_c)-\omega)-s\,v_c\|^2$, \ $\mathcal S_t$: next $\min\{20,K\}$ classes
   \EndIf
   \State $\theta \gets \mathrm{Adam}(\theta,\ \nabla_\theta \mathcal{L})$
   \If{step $\bmod \lceil 200/B\rceil = 0$} \Comment{probe, Eq.~\ref{eq:share}}
      \State $\sigma \gets \frac1K \max_j |\{c : \hat y(m_c)=j\}|$ under the current $\theta$
      \If{$\sigma > 0.25$ \textbf{and} $n - n_{\mathrm{rst}} \ge 1000$} \Comment{reset}
         \State $\theta \gets \theta_0$;\ new optimizer;\ $m_c \gets x^a_c$;\ $q_c \gets \varnothing$,\ $N_c \gets 0$;\ $n_{\mathrm{rst}} \gets n$
      \EndIf
   \EndIf
   \For{each sample $i$ with $H(\mathrm{softmax}\,\ell(z_i)) < 0.4\ln K$} \Comment{bank admission}
      \State $m_{\hat y_i} \gets x_i$ \Comment{per class, the sample closest to $\mu_{\hat y_i}$}
   \EndFor
   \State output $\hat y$;\quad $n \gets n + B$
\EndFor
\Statex
\Function{QuantileGate}{$c, \rho$} \Comment{Appendix~\ref{app:quantile}}
   \State $N_c \gets N_c+1$
   \If{$N_c \le 20$}
      \State collect $\rho$; \textbf{if} $N_c=20$ \textbf{then} $q_c \gets$ $0.2$-quantile of the collected values
      \State \Return true
   \ElsIf{$\rho \le q_c$}
      \State $q_c \gets q_c(1-0.8\,\eta)$; \Return true
   \Else
      \State $q_c \gets q_c(1+0.2\,\eta)$; \Return false
   \EndIf
\EndFunction
\end{algorithmic}
\end{algorithm}

\subsection{Baselines}
\label{app:baselines}
All baselines are run from the official implementations released with SaTeen, which bundle TENT,
EATA, SAR, DeYO, ReCAP and SaTeen, under their batch-size-1 hyperparameters: SGD with momentum
$0.9$. The learning rate is $(0.00025/64)\cdot B\cdot 2$ for ResNet at $B<32$ and $0.00025$ (episodic) or
$0.0001$ (continual) at $B\ge32$, and $(0.001/64)\cdot B$ (episodic) or $(0.0005/64)\cdot B$
(continual) for ViT, with an additional factor of $2$ for SAR, DeYO and ReCAP at $B=1$, entropy
margins $0.4\ln K$ (EATA, SAR), $0.5\ln K$ and $0.2\ln K$ (DeYO), $0.2\ln K$ (SaTeen) and $0.8\ln K$, $0.7\ln K$ (ReCAP), ReCAP in its \texttt{recap\_plpd} variant, and EATA without its Fisher regularizer, the variant its authors call ETA. The regularizer serves to
prevent forgetting on in-distribution data, and its Fisher information is estimated from clean
in-distribution samples, which our protocol does not provide. The original paper reports that the two
variants reach comparable accuracy on corrupted data, with ETA slightly ahead, for example $19.4\%$
against $19.7\%$ error on CIFAR-10-C and $54.5\%$ against $54.8\%$ on ImageNet-R
\citep[Tables 3 and 4]{niu2022efficient}. Since we report accuracy on corrupted data only, omitting the
regularizer does not disadvantage EATA at batch size $64$. We refer to this variant as EATA throughout. BatchNorm backbones keep their source running statistics and adapt only the affine parameters, as required at batch size 1. The seed is set once and re-set before each corruption,
as in the released scripts.

Two adjustments were needed. First, ReCAP requires a per-channel feature-variance prior that is
released only for ImageNet. For CIFAR we compute it from the clean CIFAR training features and
calibrate its region scale so that the kernel exponent matches the ImageNet setting, without which
ReCAP degenerates to no adaptation on CIFAR. Second, in the continual protocol the released
scripts re-create the adaptation wrapper at each corruption while the model weights carry over, and
we follow them. As a consequence the reference state that SAR's and ReCAP's recovery mechanisms
restore is the model at the start of the current corruption rather than the source model. OriTTA
receives no corruption-boundary information in either protocol.

RDumb~\citep{press2023rdumb} is ETA restored to its source weights every $64{,}000$ images, the
released interval of $1{,}000$ steps at batch size $64$, with the continual learning rate of EATA.
CoTTA~\citep{wang2022continual} is run with its released hyperparameters at batch size 1.
SAR+RT uses SAR's loss, filter and optimizer unchanged; its bank is seeded with the anchors of the
same backbone, filled by entropy below $0.4\ln K$, and re-classified every $200$ frames, and a
reset to the source weights fires when $\sigma>0.25$ and at least $1{,}000$ frames have passed.
We do not include LAME~\citep{boudiaf2022parameter}, whose correction is computed over the samples
of a batch and reduces to the source prediction at batch size 1, or NOTE~\citep{gong2022note},
which requires retraining the source model with its instance-aware normalization layers.

\subsection{Hardware and Runtime}
\label{app:runtime}
All experiments were run on a single NVIDIA RTX PRO 6000 Blackwell GPU with 96\,GB of memory. Table~\ref{tab:app_runtime} reports the wall-clock time of each method on IN-RN50, episodic, shuffled, at batch size 1. The one-off anchor synthesis, is not included.
\begin{table}[h]
\centering\footnotesize
\setlength{\tabcolsep}{5pt}
\caption{Wall-clock time on IN-RN50, episodic, shuffled, batch size 1.}
\label{tab:app_runtime}
\begin{tabular}{l ccc c}
\toprule
\rowcolor{hdr} & gaus (s) & fog (s) & brit (s) & total (h) \\
\midrule
No adaptation & 304.0  & 301.4  & 300.9  & 1.3  \\
TENT          & 715.5  & 657.3  & 662.3  & 2.8 \\
EATA          & 816.7  & 1097.6 & 1328.5 & 4.5  \\
SAR           & 1961.1 & 3438.8 & 3414.4 & 12.0 \\
DeYO          & 710.6  & 669.4  & 1139.1 & 3.5   \\
ReCAP         & 1065.9 & 1473.2 & 1778.3 & 6.0  \\
SaTeen        & 1120.9 & 1882.6 & 2034.1 & 7.0  \\
\rowcolor{grn} OriTTA & 1175.0 & 1499.0 & 1215.0 & 5.4 \\
\bottomrule
\end{tabular}
\end{table}

Per frame, OriTTA performs one forward and one backward pass on the input, plus on average $2/3$
of an additional forward-backward pass for $L_{\mathrm{align}}$ ($20$ bank inputs every $30$
frames) and a no-gradient re-classification of the $K$ bank inputs every $200$ frames, which is
batched and accounts for a small share of the time. OriTTA is faster than SAR, ReCAP and SaTeen,
which run a second forward and backward pass for sharpness-aware or augmented updates, and slower than TENT and EATA, which update on the input alone, and DeYO, which adds only a forward pass on a patch-shuffled copy.

\section{Further Related Work}
\label{app:related}

This section complements Section~\ref{sec:related} with the lines of work closest to the frame of
OriTTA, which the main text only touches on.

\paragraph{Neural collapse and test-time adaptation.}
Neural collapse~\citep{papyan2020prevalence} describes the terminal phase of training, in which
class means converge to a simplex equiangular tight frame aligned with the classifier rows, and it
has recently been carried into adaptation. NCTTA~\citep{chen2026nctta} observes that the feature of
a single sample aligns with the weight vector of its class, and realigns normalized features with the
normalized classifier weights during adaptation, combined with entropy minimization and with targets
that blend geometric proximity and predictive confidence. TCA~\citep{ni2025maintaining} preserves the
inter-class topology by pushing pseudo-label centroids toward uniformity, and
neural-collapse-inspired alignment has also been used for out-of-distribution
generalization~\citep{chen2024neural}. NEO~\citep{murphy2026neo} assumes, from neural collapse under
cross-entropy, that the source features are centered at the origin, and re-centers target features
there by subtracting their running mean, without updating the model. OriTTA differs from these
methods in two respects. First, it does not measure from the origin. The origin is a property of the
parameterization, and Table~\ref{tab:frame} shows that it lies far from where the classes are
arranged. The classifier rows also share a common component, $81\%$ of a row's norm on IN-RN50,
that no decision reads, so alignment with the raw rows differs from alignment with the centered axes
$v_c$. Second, unlike NEO and TCA, which estimate a center or class centroids from the stream,
OriTTA fixes its frame before any test data. On a class-blocked stream a running mean of the features
describes the class that is currently presented rather than the shift, whereas the frame of OriTTA
does not change.

\paragraph{Centering and angular geometry of features.}
That centering matters for angular comparisons is known from few-shot classification, where
subtracting the mean of the base-class features before normalization improves nearest-centroid
classification~\citep{wang2019simpleshot}. Cosine classifiers with angular
margins~\citep{wang2017normface,deng2019arcface} and classifiers fixed to a simplex
ETF~\citep{zhu2021geometric,yang2022inducing} likewise place the classes on a sphere around the
origin. OriTTA uses the same geometric intuition at test time, with two differences. The center
$\omega$ is recovered from the classifier and synthetic prototypes, since the source data are not
available, and the angular loss is measured from $\omega$ within the subspace the classifier reads.

\paragraph{Reference structures for adaptation.}
Several methods adapt toward a reference derived from the source model or from the stream.
SHOT~\citep{liang2020shot} freezes the source classifier and adapts the feature extractor so that
target features fit it, through information maximization and pseudo-labels over the whole target
set. T3A~\citep{iwasawa2021test} replaces the classifier rows by prototypes of pseudo-labeled test
features. SaTeen~\citep{sateen} maximizes the cross-entropy discrepancy between a sample and a
structure-disrupted copy of it, and minimizes the reconstruction error of test features in a subspace
spanned by incremental PCA on samples it judges reliable. OriTTA shares the premise that a reference
exists, but takes it from the classifier alone. Its subspace is the row space of the centered
classifier rather than one estimated from reliable samples. Because this subspace is fixed before
the stream, it cannot follow the drift it is meant to correct, and it needs no test samples of any
class to be formed.

\paragraph{Synthesizing inputs from the source model.}
Optimizing an input to maximize a class score is an old tool for visualizing
classifiers~\citep{simonyan2014deep}. For data-free knowledge distillation, synthetic inputs have been
made realistic with a generator or by matching the batch-normalization statistics of the
network~\citep{chen2019data,yin2020dreaming}, and source-free domain adaptation has trained a
class-conditional generator together with the source model to synthesize samples for
adaptation~\citep{li2020model}. The anchors of OriTTA serve a different purpose. They are needed only
for where their features land, not for their appearance, so they are synthesized by cross-entropy
with a small total-variation term and no statistical prior, which also makes the procedure identical
for BatchNorm, GroupNorm and LayerNorm backbones. Appendix~\ref{app:anchors} shows that the frame they
define does not depend on the particular draw.

\paragraph{Detecting failure without labels.}
Estimating the accuracy of a model under shift without target labels is a problem of its
own~\citep{garg2022leveraging}, usually addressed with confidences calibrated on labeled source
data. The bank share of Section~\ref{sec:reset} answers a narrower question, whether the predictions
have merged onto a few classes, and needs no source data. Because the bank holds one input per class,
it is also insensitive to the label order of the stream, which a confidence statistic computed over
the stream is not.

\paragraph{Scope.}
Test-time adaptation of vision--language models, for example by tuning prompts on a single test
input~\citep{shu2022tpt}, adapts a different kind of classifier, whose rows are text embeddings, and
is outside the scope of this work. The frame of Section~\ref{sec:frame} requires only a linear head,
and extending it to such models is a direction we leave open.

\section{Firing Rate Across Settings}
\label{app:reset_counts}
Table~\ref{tab:full_resets} reports how often the collapse detector of Section~\ref{sec:reset}
fires in each of the sixteen settings, as the mean number of resets per run over the ten seeds and
per corruption. The detector is armed identically everywhere: it is not disabled in the episodic
protocol, and it receives no information about corruption boundaries in either protocol.

\begin{table}[h]
\centering\footnotesize
\setlength{\tabcolsep}{2.6pt}
\caption{Resets per run, averaged over ten seeds, per corruption. A dot is zero. \emph{Runs} is
the number of the ten seeds in which the detector fired at least once. In the continual protocol
the corruptions are presented from left to right.}
\label{tab:full_resets}
\resizebox{\linewidth}{!}{%
\begin{tabular}{l ll ccccccccccccccc | cc}
\toprule
\rowcolor{hdr} & & & gaus & shot & impu & defo & glas & moti & zoom & snow & fros & fog & brit & cont & elas & pixe & jpeg & Total & Runs \\
\midrule
\multirow{4}{*}{IN-RN50} & episodic & shuffled & $\cdot$ & $\cdot$ & $\cdot$ & $\cdot$ & $\cdot$ & $\cdot$ & $\cdot$ & $\cdot$ & $\cdot$ & $\cdot$ & $\cdot$ & $\cdot$ & $\cdot$ & $\cdot$ & $\cdot$ & 0.0 & 0/10 \\
 & episodic & class-blocked & $\cdot$ & $\cdot$ & $\cdot$ & $\cdot$ & $\cdot$ & $\cdot$ & $\cdot$ & $\cdot$ & $\cdot$ & $\cdot$ & $\cdot$ & $\cdot$ & $\cdot$ & $\cdot$ & $\cdot$ & 0.0 & 0/10 \\
 & continual & shuffled & $\cdot$ & $\cdot$ & 1.0 & $\cdot$ & 1.0 & $\cdot$ & 0.3 & 0.7 & $\cdot$ & 0.1 & $\cdot$ & 0.8 & 0.5 & 0.5 & 0.3 & 5.2 & 10/10 \\
 & continual & class-blocked & $\cdot$ & $\cdot$ & 0.8 & 0.2 & 0.8 & $\cdot$ & 0.9 & 0.1 & $\cdot$ & 0.6 & 0.1 & 0.1 & 1.0 & $\cdot$ & $\cdot$ & 4.6 & 10/10 \\
\midrule
\multirow{4}{*}{IN-ViT} & episodic & shuffled & $\cdot$ & $\cdot$ & $\cdot$ & $\cdot$ & $\cdot$ & $\cdot$ & $\cdot$ & $\cdot$ & $\cdot$ & $\cdot$ & $\cdot$ & $\cdot$ & $\cdot$ & $\cdot$ & $\cdot$ & 0.0 & 0/10 \\
 & episodic & class-blocked & $\cdot$ & $\cdot$ & $\cdot$ & $\cdot$ & $\cdot$ & $\cdot$ & $\cdot$ & $\cdot$ & $\cdot$ & $\cdot$ & $\cdot$ & $\cdot$ & $\cdot$ & $\cdot$ & $\cdot$ & 0.0 & 0/10 \\
 & continual & shuffled & $\cdot$ & $\cdot$ & $\cdot$ & 0.6 & 0.3 & $\cdot$ & $\cdot$ & 0.8 & 0.1 & $\cdot$ & $\cdot$ & 0.3 & 0.9 & $\cdot$ & $\cdot$ & 3.0 & 10/10 \\
 & continual & class-blocked & $\cdot$ & $\cdot$ & 0.3 & 0.7 & $\cdot$ & $\cdot$ & $\cdot$ & 1.0 & $\cdot$ & $\cdot$ & $\cdot$ & 0.7 & 0.3 & $\cdot$ & $\cdot$ & 3.0 & 10/10 \\
\midrule
\multirow{4}{*}{C100} & episodic & shuffled & $\cdot$ & $\cdot$ & $\cdot$ & $\cdot$ & $\cdot$ & $\cdot$ & $\cdot$ & $\cdot$ & $\cdot$ & $\cdot$ & $\cdot$ & $\cdot$ & $\cdot$ & $\cdot$ & $\cdot$ & 0.0 & 0/10 \\
 & episodic & class-blocked & $\cdot$ & $\cdot$ & $\cdot$ & $\cdot$ & $\cdot$ & $\cdot$ & $\cdot$ & $\cdot$ & $\cdot$ & $\cdot$ & $\cdot$ & $\cdot$ & $\cdot$ & $\cdot$ & $\cdot$ & 0.0 & 0/10 \\
 & continual & shuffled & $\cdot$ & $\cdot$ & $\cdot$ & $\cdot$ & $\cdot$ & $\cdot$ & $\cdot$ & $\cdot$ & $\cdot$ & $\cdot$ & $\cdot$ & 0.9 & $\cdot$ & $\cdot$ & $\cdot$ & 0.9 & 9/10 \\
 & continual & class-blocked & $\cdot$ & $\cdot$ & $\cdot$ & $\cdot$ & 0.5 & $\cdot$ & $\cdot$ & $\cdot$ & 0.1 & 0.4 & $\cdot$ & 0.1 & $\cdot$ & 0.1 & 0.1 & 1.3 & 10/10 \\
\midrule
\multirow{4}{*}{C10} & episodic & shuffled & $\cdot$ & $\cdot$ & $\cdot$ & $\cdot$ & $\cdot$ & $\cdot$ & $\cdot$ & $\cdot$ & $\cdot$ & $\cdot$ & $\cdot$ & $\cdot$ & $\cdot$ & $\cdot$ & $\cdot$ & 0.0 & 0/10 \\
 & episodic & class-blocked & $\cdot$ & $\cdot$ & $\cdot$ & $\cdot$ & $\cdot$ & 0.1 & $\cdot$ & $\cdot$ & $\cdot$ & 0.1 & 0.1 & 0.1 & 0.1 & $\cdot$ & $\cdot$ & 0.5 & 4/10 \\
 & continual & shuffled & $\cdot$ & $\cdot$ & $\cdot$ & $\cdot$ & $\cdot$ & $\cdot$ & $\cdot$ & $\cdot$ & $\cdot$ & $\cdot$ & $\cdot$ & $\cdot$ & $\cdot$ & $\cdot$ & $\cdot$ & 0.0 & 0/10 \\
 & continual & class-blocked & $\cdot$ & 0.1 & $\cdot$ & 0.1 & $\cdot$ & $\cdot$ & 0.2 & 0.1 & 0.4 & 0.1 & $\cdot$ & 0.9 & 0.2 & $\cdot$ & 0.1 & 2.2 & 9/10 \\
\bottomrule
\end{tabular}}
\end{table}

\begin{table}[h]
\centering\footnotesize
\setlength{\tabcolsep}{6pt}
\caption{Bank share $\sigma$ in runs without and with a reset, pooled over both stream orders and
ten seeds. The threshold is $0.25$.}
\label{tab:app_sigma}
\begin{tabular}{l l c c}
\toprule
\rowcolor{hdr} & & max $\sigma$, runs without reset & $\sigma$ at firing (median) \\
\midrule
IN-RN50 & episodic  & 0.037 & -- \\
        & continual & --    & 0.258 \\
IN-ViT  & episodic  & 0.022 & -- \\
        & continual & --    & 0.293 \\
C100    & episodic  & 0.150 & -- \\
        & continual & 0.030 & 0.320 \\
C10     & episodic  & 0.200 & 0.300 \\
        & continual & 0.200 & 0.300 \\
\bottomrule
\end{tabular}
\end{table}
Three patterns are visible. In the episodic protocol the detector is silent: over the $10$ seeds of
all four settings it never fires, except on C10 under the class-blocked stream, where it fires in
four of ten runs and $0.5$ times per run. Within a single corruption the losses are stable enough that there is nothing to catch. Table~\ref{tab:app_sigma} gives the largest bank share reached in runs where the detector never fires: on ImageNet it stays below $0.04$, a seventh of the threshold, and on C100 it reaches $0.15$. On C10 the share moves in steps of $1/K=0.1$, healthy runs reach $0.2$, two classes merged into one, and the detector fires at $0.3$, three. The threshold therefore separates healthy runs from collapsing ones in every setting, by a wide margin on ImageNet and by one step of the bank's resolution on C10.

In the continual protocol it fires between $0.9$ and $5.2$ times per run in every setting but C10 under the shuffled stream, where it never fires. The count grows with the number of classes, from $0.9$--$2.2$ on the
CIFAR settings to $3.0$ on IN-ViT and $4.6$--$5.2$ on IN-RN50, and it is concentrated on impulse noise, glass blur, contrast and elastic transform
account for most of the firings on IN-RN50. No run reaches the repeated firing that SAR's own recovery shows under the same protocol (Subsection~\ref{sec:detector}).

Finally, the detector fires in nearly every continual run but never more than a few times: with the
exception of C10 under the shuffled stream, at least nine of the ten seeds see one reset, and no
setting exceeds $5.2$ per run against the $15$ corruption boundaries the stream contains. The detector is therefore neither idle nor hair-triggered. Its rate is similar under the two stream orders on ImageNet and C100, since the bank holds one representative per class regardless of the order; C10 is the exception, where it fires only under the class-blocked stream, whose blocks of one thousand images put the most pressure on the classes absent from it (Appendix~\ref{app:abl_align}).

\section{Ablation Study}
\label{app:ablation}

\subsection{Optimizer Ablation}
\label{app:optim}

Every baseline we compare against optimizes with SGD and momentum $0.9$, whereas OriTTA uses Adam.
To check that the advantage does not come from this choice, we run OriTTA with SGD at the same
momentum on IN-RN50, episodic, class-blocked, over the ten seeds. Table~\ref{tab:app_optim} gives the result.
\begin{table}[h]
\centering\footnotesize
\setlength{\tabcolsep}{2.6pt}
\caption{Adam against SGD, IN-RN50, episodic, class-blocked, ten seeds. The second line of each
method is the standard deviation over seeds.}
\label{tab:app_optim}
\resizebox{\linewidth}{!}{%
\begin{tabular}{l ccccccccccccccc | c}
\toprule
\rowcolor{hdr} & gaus & shot & impu & defo & glas & moti & zoom & snow & fros & fog & brit & cont & elas & pixe & jpeg & Average \\
\midrule
OriTTA + Adam & 40.79 & 42.55 & 41.81 & 33.32 & 31.09 & 41.83 & 45.80 & 53.69 & 52.67 & 57.92 & 72.50 & 53.12 & 48.79 & 58.73 & 60.05 & 48.98 \\
\rowcolor{gray!8} \scriptsize sd & \scriptsize 0.17 & \scriptsize 0.24 & \scriptsize 0.23 & \scriptsize 0.29 & \scriptsize 0.38 & \scriptsize 0.22 & \scriptsize 0.13 & \scriptsize 0.20 & \scriptsize 0.15 & \scriptsize 0.36 & \scriptsize 0.11 & \scriptsize 0.29 & \scriptsize 0.35 & \scriptsize 0.21 & \scriptsize 0.09 & \scriptsize 0.10 \\
\midrule
OriTTA + SGD  & 43.46 & 44.45 & 43.40 & 34.75 & 34.41 & 45.04 & 48.76 & 55.86 & 53.47 & 55.93 & 71.82 & 55.08 & 54.25 & 62.63 & 60.61 & 50.93 \\
\rowcolor{gray!8} \scriptsize sd & \scriptsize 0.20 & \scriptsize 0.24 & \scriptsize 0.27 & \scriptsize 0.27 & \scriptsize 0.37 & \scriptsize 0.28 & \scriptsize 0.18 & \scriptsize 0.19 & \scriptsize 0.24 & \scriptsize 1.03 & \scriptsize 0.19 & \scriptsize 0.28 & \scriptsize 0.39 & \scriptsize 0.23 & \scriptsize 0.18 & \scriptsize 0.32 \\
\midrule
$\Delta$ & $+2.67$ & $+1.90$ & $+1.59$ & $+1.43$ & $+3.32$ & $+3.21$ & $+2.96$ & $+2.17$ & $+0.80$ & $-1.99$ & $-0.68$ & $+1.96$ & $+5.46$ & $+3.90$ & $+0.56$ & $+1.95$ \\
\bottomrule
\end{tabular}}
\end{table}

SGD is ahead by about two points on average and on thirteen of the fifteen corruptions, most on
those where the source model is furthest off and a larger step pays, such as elastic transform and
pixelate. The advantage of OriTTA over the baselines is therefore not an artifact of Adam. With the baselines' own optimizer it is larger still.
We nonetheless use Adam throughout, for three reasons that the same table exposes.

First, stability. The seed-to-seed deviation of the average triples under SGD, and on fog, the
corruption on which the baselines are least stable, it grows from $0.36$ to $1.03$ while the mean
falls by two points. At batch size 1 every update is a single-sample gradient. Adam's normalization
by $\sqrt{v}$ bounds the step taken on any one sample, whereas SGD passes its full magnitude through. Three times the selected SGD rate already triggers resets within a single corruption.

Second, scale invariance. Adam's update is unchanged when the objective is multiplied by a
constant, so the factors $1/r$ and $s^2$ in our losses, which differ by orders of magnitude across
datasets and backbones, never enter the learning rate. This is what allows one rate per
normalization type to serve every setting. Under SGD the rate absorbs the scale of the objective
and would have to be chosen again for each dataset and backbone. The value selected on IN-RN50 would have to be rescaled for every other setting, since the gradient magnitude of each loss depends on $s$ and $r$.

Third, batch size. The $\sqrt{B}$ scaling rule that keeps OriTTA's schedule fixed across batch
sizes (Appendix~\ref{app:batch}) is the rule appropriate for Adam~\citep{malladi2022sdes}. SGD
has no rule that holds down to batch size 1: the released baselines scale their rate linearly with
the batch, and SAR, DeYO and ReCAP then double it at batch size 1 as a separate correction, while
SaTeen does not (Appendix~\ref{app:baselines}). Each such correction is one more setting to be
validated per method and per batch size.

Since the aim of this work is a single procedure with no per-setting tuning, we regard the two
points as the price of that property, and Adam as the appropriate choice.
\begin{table}[h]
\centering\footnotesize
\setlength{\tabcolsep}{2.6pt}
\caption{Removing one loss term at a time for IN-RN50, episodic, class-blocked.}
\label{tab:abl_loss}
\resizebox{\linewidth}{!}{%
\begin{tabular}{l ccccccccccccccc | c}
\toprule
\rowcolor{hdr} & gaus & shot & impu & defo & glas & moti & zoom & snow & fros & fog & brit & cont & elas & pixe & jpeg & Average \\
\midrule
\rowcolor{grn}OriTTA (full) & 40.79 & 42.55 & 41.81 & 33.32 & 31.09 & 41.83 & 45.80 & 53.69 & 52.67 & 57.92 & 72.50 & 53.12 & 48.79 & 58.73 & 60.05 & \textbf{48.98} \\
w/o $L_{\mathrm{ang}}$ & 26.13 & 27.30 & 25.61 & 22.63 & 14.42 & 28.11 & 29.11 & 45.44 & 48.82 & 35.14 & 69.48 & 39.61 & 25.09 & 35.00 & 53.92 & 35.05\,\scriptsize{(-13.92)} \\
w/o $L_{\mathrm{rad}}$ & 40.00 & 41.44 & 40.87 & 32.06 & 30.05 & 40.08 & 44.61 & 52.98 & 51.84 & 57.48 & 71.70 & 52.03 & 47.44 & 57.64 & 58.92 & 47.94\,\scriptsize{(-1.03)} \\
w/o $L_{\mathrm{align}}$ & 40.49 & 41.36 & 40.71 & 33.08 & 30.23 & 40.19 & 44.59 & 52.62 & 51.53 & 58.28 & 72.35 & 51.85 & 47.89 & 58.28 & 59.35 & 48.19\,\scriptsize{(-0.79)} \\
\bottomrule
\end{tabular}}
\vspace{-4mm}
\end{table}

\subsection{Losses Ablation}
\label{app:lossablation}
Table~\ref{tab:abl_loss} removes one term at a time on IN-RN50. The angular loss does most of the
work. Without it accuracy falls by fourteen points, by a third to a half on most corruptions and
least on brightness, frost and JPEG compression, where the source model is already accurate. It is
the only term that assigns a reliable feature to a class and separates it from the others, and
the other two cannot take its place. The radial loss contributes about a point, spread evenly
across the corruptions. The alignment loss contributes the least here. Its role is to hold in place
the classes the stream is not presenting, and on ImageNet a class is absent only while fifty images
of another class pass. Appendix~\ref{app:abl_align} lengthens that interval.

\subsection{Alignment Loss on Long Class Blocks}
\label{app:abl_align}

The alignment loss exists for the classes that the stream is not currently presenting: it pulls a
stored representative of every class back to its ideal point, so that the model does not drift away
from a class while another one occupies the stream. On IN-RN50 this pressure is weak even under
the class-blocked order, because each class arrives as a block of only fifty images, and removing
$L_{\mathrm{align}}$ there costs $0.8$ points (Table~\ref{tab:abl_loss}). The CIFAR benchmarks
lengthen the blocks to one hundred images on C100 and one thousand on C10, and we repeat the
ablation on both, episodic and class-blocked, over ten seeds (Tables~\ref{tab:app_align_sum}
and~\ref{tab:app_align}).

\begin{table}[h]
\centering\footnotesize
\setlength{\tabcolsep}{5pt}
\caption{Removing $L_{\mathrm{align}}$, episodic, class-blocked. \emph{Block} is the number of
consecutive images of one class and \emph{resets} is the mean number of detector firings per run and
the number of runs, out of ten, in which it fired.}
\label{tab:app_align_sum}
\begin{tabular}{l c cc c cc}
\toprule
\rowcolor{hdr} & & \multicolumn{2}{c}{accuracy} & & \multicolumn{2}{c}{resets per run} \\
\rowcolor{hdr} & block & full & w/o $L_{\mathrm{align}}$ & $\Delta$ & full & w/o $L_{\mathrm{align}}$ \\
\midrule
IN-RN50 & 50   & 48.98 & 48.19 & $-0.79$ & 0 \ (0/10) & 0 \ (0/10)\\
C100    & 100  & 61.69 & 60.27 & $-1.41$ & 0 \ (0/10) & 0 \ (0/10) \\
C10     & 1000 & 68.96 & 60.94 & $-8.02$ & 0.5 \ (4/10) & 37.1 \ (10/10) \\
\bottomrule
\end{tabular}
\end{table}

\begin{table}[h]
\centering\footnotesize
\setlength{\tabcolsep}{2.6pt}
\caption{Per-corruption accuracy without $L_{\mathrm{align}}$, episodic, class-blocked, ten
seeds. The line below each result is the standard deviation over seeds.}
\label{tab:app_align}
\resizebox{\linewidth}{!}{%
\begin{tabular}{l ccccccccccccccc | c}
\toprule
\rowcolor{hdr} \textit{C10} & gaus & shot & impu & defo & glas & moti & zoom & snow & fros & fog & brit & cont & elas & pixe & jpeg & Average \\
\midrule
OriTTA & 46.96 & 52.50 & 39.10 & 73.22 & 54.49 & 78.07 & 76.05 & 78.28 & 72.51 & 83.49 & 91.52 & 75.95 & 77.83 & 61.40 & 73.03 & \textbf{68.96} \\
\rowcolor{gray!8} \scriptsize sd & \scriptsize 4.01 & \scriptsize 2.63 & \scriptsize 3.11 & \scriptsize 2.79 & \scriptsize 1.76 & \scriptsize 2.19 & \scriptsize 1.99 & \scriptsize 1.83 & \scriptsize 1.31 & \scriptsize 1.30 & \scriptsize 0.76 & \scriptsize 1.52 & \scriptsize 1.48 & \scriptsize 3.99 & \scriptsize 2.02 & \scriptsize 1.51 \\
w/o $L_{\mathrm{align}}$ & 25.06 & 31.61 & 26.24 & 63.72 & 44.72 & 73.25 & 68.00 & 80.04 & 64.90 & 83.16 & 93.09 & 64.91 & 78.54 & 43.28 & 73.55 & 60.94 \\
\rowcolor{gray!8} \scriptsize sd & \scriptsize 1.40 & \scriptsize 3.66 & \scriptsize 0.77 & \scriptsize 1.39 & \scriptsize 3.19 & \scriptsize 1.67 & \scriptsize 2.72 & \scriptsize 1.10 & \scriptsize 1.84 & \scriptsize 1.41 & \scriptsize 0.83 & \scriptsize 3.02 & \scriptsize 1.40 & \scriptsize 2.56 & \scriptsize 1.72 & \scriptsize 0.71 \\
$\Delta$ & $-21.91$ & $-20.90$ & $-12.86$ & $-9.49$ & $-9.77$ & $-4.82$ & $-8.05$ & $+1.77$ & $-7.60$ & $-0.33$ & $+1.57$ & $-11.04$ & $+0.71$ & $-18.12$ & $+0.52$ & $-8.02$ \\
\midrule
\midrule
\rowcolor{hdr} \textit{C100} & gaus & shot & impu & defo & glas & moti & zoom & snow & fros & fog & brit & cont & elas & pixe & jpeg & Average \\
\midrule
OriTTA & 48.07 & 52.29 & 51.20 & 70.98 & 54.74 & 69.73 & 70.81 & 63.13 & 63.30 & 58.71 & 71.93 & 63.51 & 62.84 & 64.83 & 59.24 & \textbf{61.69} \\
\rowcolor{gray!8} \scriptsize sd & \scriptsize 0.53 & \scriptsize 0.49 & \scriptsize 0.33 & \scriptsize 0.41 & \scriptsize 0.63 & \scriptsize 0.55 & \scriptsize 0.30 & \scriptsize 0.42 & \scriptsize 0.40 & \scriptsize 0.59 & \scriptsize 0.34 & \scriptsize 0.72 & \scriptsize 0.57 & \scriptsize 0.64 & \scriptsize 0.37 & \scriptsize 0.33 \\
w/o $L_{\mathrm{align}}$ & 45.27 & 50.12 & 48.89 & 70.13 & 52.93 & 68.88 & 70.01 & 61.94 & 61.97 & 57.30 & 71.12 & 62.24 & 61.55 & 63.70 & 58.05 & 60.27 \\
\rowcolor{gray!8} \scriptsize sd & \scriptsize 0.79 & \scriptsize 0.53 & \scriptsize 0.52 & \scriptsize 0.44 & \scriptsize 0.50 & \scriptsize 0.57 & \scriptsize 0.34 & \scriptsize 0.39 & \scriptsize 0.53 & \scriptsize 0.52 & \scriptsize 0.39 & \scriptsize 0.94 & \scriptsize 0.59 & \scriptsize 0.60 & \scriptsize 0.40 & \scriptsize 0.37 \\
$\Delta$ & $-2.80$ & $-2.16$ & $-2.31$ & $-0.85$ & $-1.81$ & $-0.84$ & $-0.80$ & $-1.19$ & $-1.33$ & $-1.42$ & $-0.81$ & $-1.27$ & $-1.29$ & $-1.12$ & $-1.19$ & $-1.41$ \\
\bottomrule
\end{tabular}}
\end{table}

\paragraph{The cost grows with the block length.}
On C100, removing $L_{\mathrm{align}}$ lowers accuracy on every one of the fifteen corruptions, by
$1.4$ points on average, and the detector never fires with or without it. On C10 the effect changes
in kind. Without $L_{\mathrm{align}}$ the model collapses repeatedly: the detector fires $37$
times per run instead of $0.5$, in every seed rather than four of ten, and accuracy falls by eight
points. The losses concentrate where the source model is weakest and adaptation matters most, on
the three noise corruptions and on pixelate, by $13$ to $22$ points, while on snow, brightness,
elastic transform and JPEG, where the source model is already accurate, returning to it often
costs nothing. Across the three settings the cost of removing $L_{\mathrm{align}}$ rises with the
length of a class block, from $0.8$ points at fifty images to $1.4$ at one hundred and $8.0$ with
repeated collapse at one thousand. On C10 the alignment loss is also applied most often relative to
the number of classes: with $K=10$, every class is pulled back at every alignment step, whereas on
ImageNet each class is visited once every $1{,}500$ frames.

\paragraph{The detector does not replace the alignment loss.}
The standard deviation on C10 falls from $1.51$ to $0.71$ without $L_{\mathrm{align}}$, but this is
not greater stability: with a reset about every $4{,}000$ frames, the model spends most of the
stream near its source weights, and its accuracy is consistently lower rather than consistently
good. The detector recovers the model after each collapse, yet eight points are lost despite
thirty-seven recoveries. The two mechanisms therefore play different roles: $L_{\mathrm{align}}$
keeps the classes in place and prevents collapse, and the detector restores the model when
collapse occurs nonetheless.

\begin{figure}[h]
\centering
\setlength{\abovecaptionskip}{4pt}
\begin{minipage}{0.49\linewidth}
  \centering
  \includegraphics[width=\linewidth, trim=0 4mm 0 0, clip]{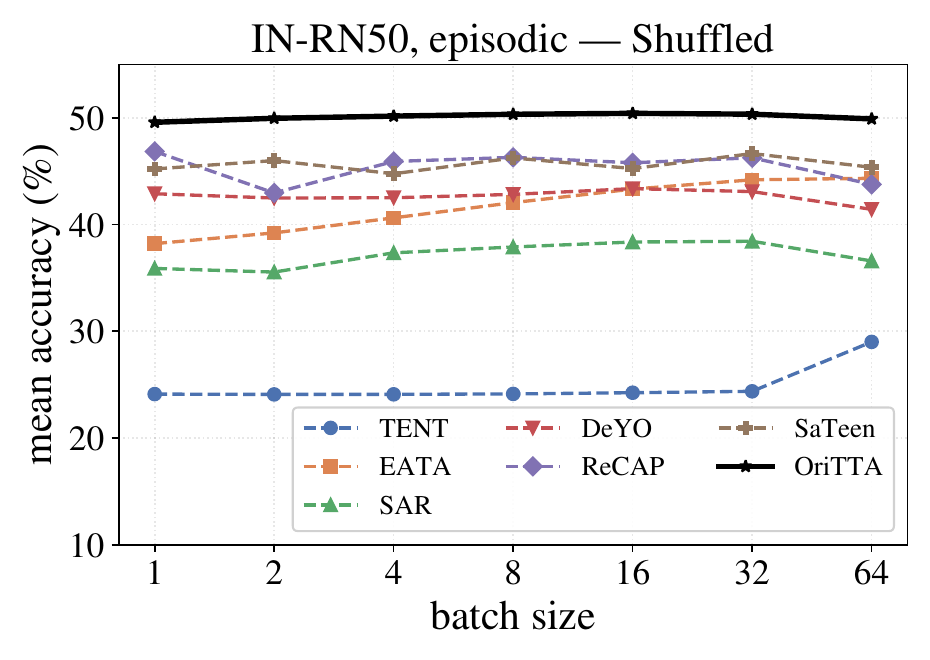}\\[-1mm]
\end{minipage}\hfill
\begin{minipage}{0.49\linewidth}
  \centering
  \includegraphics[width=\linewidth, trim=0 4mm 0 0, clip]{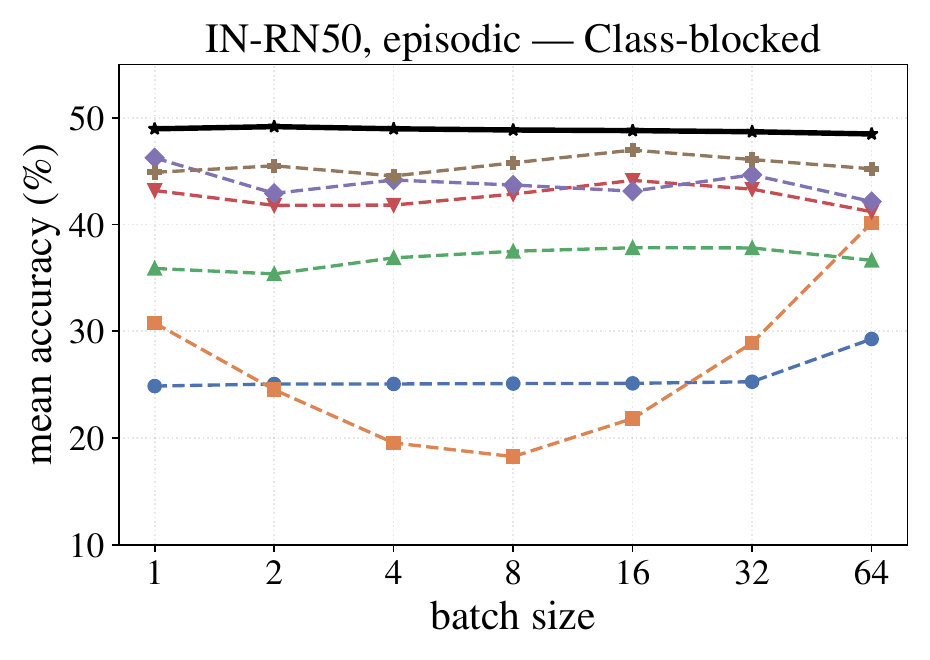}\\[-1mm]
\end{minipage}
\caption{Accuracy against batch size on IN-RN50, episodic, averaged over ten seeds.}
\label{fig:bs}
\end{figure}

\subsection{Batch Size Ablation}
Figure~\ref{fig:bs} varies the batch size from 1 to 64 on IN-RN50 in the episodic protocol mean over ten seeds, with every method's learning rate scaled by its own published rule (Appendix~\ref{app:batch}). OriTTA is the most accurate method at every batch size on both streams, and its accuracy is nearly flat: it moves within $0.8$ points on the shuffled stream and $0.5$ on the class-blocked one. A larger batch does not close the gap. DeYO, ReCAP and SaTeen gain little from larger batches, whereas EATA improves by six points on the shuffled stream but falls below 20\% at batch sizes 4 to 8 on the class-blocked stream. OriTTA remains stable across the evaluated batch sizes and stream orders. Full per-corruption results at batch size 64 are given in Appendix~\ref{app:full}.

\subsection{Ridge Weight $\lambda$}
\label{app:lambda}

The recentered origin (Eq.~\ref{eq:origin}) minimizes the sum of squared distances to the $K$
hyperplanes $H_c$ together with a ridge toward the prototype centroid $\bar z^a$ of weight
$\lambda=K/100$. The rule follows from how the reliability of the two terms changes with $K$.

The hyperplane term is a least-squares system of $K$ equations in $D$ unknowns. When $K$ is small
it is far from determined: on C10 it constrains nine directions of a $640$-dimensional space, so
its solutions form a large affine set that can be satisfied exactly at no cost, and every point of
that set is a point at which the classifier is maximally ambiguous. The centroid of ten anchors
adds little reliable information beyond selecting a point within that set, and a small $\lambda$
lets it do exactly that and no more. When $K$ is large the situation is reversed. A thousand
constraints pin the least-squares solution to a specific point, and nothing in the system requires
that point to lie where features live. On ViT-B/16, where $K-1>D$ and the system is
over-determined, the unregularized solution has norm $1.7$ against a feature scale of $80$ and is
nearly orthogonal to the feature cloud. There the centroid of a thousand prototypes is the more
trustworthy of the two, and a large $\lambda$ lets it dominate. With the per-constraint weight fixed, the effect of the ridge still differs with $K$, because the hyperplane system does: at small $K$ it is underdetermined and its residual vanishes on a whole affine set, so the ridge only selects a point within that set, whereas at large $K$ no point annihilates the residual and the ridge trades against it directly.
\begin{table}[h]
\centering\footnotesize
\setlength{\tabcolsep}{2.6pt}
\caption{C10, episodic, ten seeds, with $\lambda$ set by the rule ($K/100=0.1$) or fixed at the
ImageNet value $10$. The line below each result is the standard deviation over seeds.}
\label{tab:app_lambda}
\resizebox{\linewidth}{!}{%
\begin{tabular}{l ccccccccccccccc | c}
\toprule
\rowcolor{hdr} \textit{shuffled} & gaus & shot & impu & defo & glas & moti & zoom & snow & fros & fog & brit & cont & elas & pixe & jpeg & Average \\
\midrule
$\lambda=K/100$ & 56.14 & 60.55 & 41.11 & 80.25 & 59.54 & 82.02 & 82.34 & 80.75 & 76.99 & 84.44 & 91.32 & 80.58 & 77.84 & 73.40 & 73.43 & \textbf{73.38} \\
\rowcolor{gray!8} \scriptsize sd & \scriptsize 0.30 & \scriptsize 0.34 & \scriptsize 0.26 & \scriptsize 0.40 & \scriptsize 0.40 & \scriptsize 0.31 & \scriptsize 0.19 & \scriptsize 0.19 & \scriptsize 0.31 & \scriptsize 0.24 & \scriptsize 0.09 & \scriptsize 0.19 & \scriptsize 0.08 & \scriptsize 0.55 & \scriptsize 0.16 & \scriptsize 0.08 \\
\midrule
$\lambda=10$ & 43.82 & 48.71 & 29.60 & 67.75 & 50.51 & 71.22 & 70.63 & 72.90 & 68.64 & 76.01 & 86.84 & 67.87 & 68.48 & 58.96 & 63.57 & 63.03 \\
\rowcolor{gray!8} \scriptsize sd & \scriptsize 0.28 & \scriptsize 0.60 & \scriptsize 0.59 & \scriptsize 0.58 & \scriptsize 0.38 & \scriptsize 0.53 & \scriptsize 0.58 & \scriptsize 0.33 & \scriptsize 0.34 & \scriptsize 0.29 & \scriptsize 0.28 & \scriptsize 0.52 & \scriptsize 0.40 & \scriptsize 0.75 & \scriptsize 0.28 & \scriptsize 0.21 \\
\midrule
$\Delta$ & $-12.32$ & $-11.84$ & $-11.51$ & $-12.50$ & $-9.03$ & $-10.80$ & $-11.71$ & $-7.85$ & $-8.35$ & $-8.43$ & $-4.48$ & $-12.71$ & $-9.36$ & $-14.44$ & $-9.86$ & $-10.35$ \\
\midrule
\midrule
\rowcolor{hdr} \textit{class-blocked} & gaus & shot & impu & defo & glas & moti & zoom & snow & fros & fog & brit & cont & elas & pixe & jpeg & Average \\
\midrule
$\lambda=K/100$ & 46.96 & 52.50 & 39.10 & 73.22 & 54.49 & 78.07 & 76.05 & 78.28 & 72.51 & 83.49 & 91.52 & 75.95 & 77.83 & 61.40 & 73.03 & \textbf{68.96} \\
\rowcolor{gray!8} \scriptsize sd & \scriptsize 4.01 & \scriptsize 2.63 & \scriptsize 3.11 & \scriptsize 2.79 & \scriptsize 1.76 & \scriptsize 2.19 & \scriptsize 1.99 & \scriptsize 1.83 & \scriptsize 1.31 & \scriptsize 1.30 & \scriptsize 0.76 & \scriptsize 1.52 & \scriptsize 1.48 & \scriptsize 3.99 & \scriptsize 2.02 & \scriptsize 1.51 \\
\midrule
$\lambda=10$ & 38.04 & 41.09 & 25.39 & 59.53 & 47.97 & 66.38 & 63.86 & 71.50 & 64.49 & 75.12 & 87.09 & 58.99 & 68.62 & 46.33 & 63.38 & 58.52 \\
\rowcolor{gray!8} \scriptsize sd & \scriptsize 5.42 & \scriptsize 3.99 & \scriptsize 2.49 & \scriptsize 2.73 & \scriptsize 3.68 & \scriptsize 2.98 & \scriptsize 2.93 & \scriptsize 2.12 & \scriptsize 2.94 & \scriptsize 2.53 & \scriptsize 1.51 & \scriptsize 4.37 & \scriptsize 3.29 & \scriptsize 3.72 & \scriptsize 2.48 & \scriptsize 2.27 \\
\midrule
$\Delta$ & $-8.92$ & $-11.41$ & $-13.71$ & $-13.69$ & $-6.52$ & $-11.69$ & $-12.19$ & $-6.78$ & $-8.02$ & $-8.37$ & $-4.43$ & $-16.96$ & $-9.21$ & $-15.07$ & $-9.65$ & $-10.44$ \\
\bottomrule
\end{tabular}}
\end{table}

Table~\ref{tab:app_lambda} tests the small-$K$ end of this argument by replacing the rule on C10
with the ImageNet value $\lambda=10$, over ten seeds. Trusting the ten-prototype centroid costs ten
points on both streams and on every corruption, and on the class-blocked stream the seed-to-seed
deviation grows as well. At the large-$K$ end the rule is insensitive: on ImageNet, raising
$\lambda$ from $10$ to $100$ or dropping the hyperplane term altogether moves $\omega$ by less
than $0.06\,s$ and leaves accuracy unchanged.

\section{Full Results}
\label{app:full}

Tables~\ref{tab:full_rn50}--\ref{tab:full_c10} give the accuracy on every corruption for
every setting and protocol, as the mean over ten seeds with the standard deviation over seeds as
a subscript. Within each block, the best result in each column is shown in bold and the second
best is underlined. In the continual protocol the model adapts through the corruptions from left
to right without any reset at corruption boundaries. Table~\ref{tab:full_bs64} gives the full results for IN-RN50 at batch size $64$.

\begin{sidewaystable}
\centering\scriptsize
\setlength{\tabcolsep}{2.5pt}
\caption{Full results on ImageNet-C, ResNet-50-GN.}
\label{tab:full_rn50}
\resizebox{\linewidth}{!}{%
\begin{tabular}{>{\columncolor{hdr}}c l ccccccccccccccc | c}
\toprule
\rowcolor{hdr} & episodic & gaus & shot & impu & defo & glas & moti & zoom & snow & fros & fog & brit & cont & elas & pixe & jpeg & Average \\
\midrule\midrule
& TENT   & 12.35\sd{2.55} & 20.22\sd{2.34} & 13.91\sd{1.47} & 13.16\sd{0.27} & 6.46\sd{3.03} & 18.07\sd{0.97} & 18.23\sd{0.68} & 16.00\sd{0.85} & 20.65\sd{0.70} & 1.48\sd{0.10} & 69.80\sd{0.07} & 42.04\sd{0.25} & 6.22\sd{0.49} & 49.38\sd{0.21} & 53.75\sd{0.09} & 24.11\sd{0.40} \\
& EATA   & 29.49\sd{1.04} & 31.55\sd{1.85} & 30.23\sd{1.76} & 17.93\sd{0.80} & 19.32\sd{0.53} & 30.45\sd{1.00} & 32.62\sd{0.88} & 42.73\sd{1.94} & 41.21\sd{1.60} & 45.55\sd{0.56} & 69.40\sd{0.10} & 43.66\sd{0.53} & 32.98\sd{0.55} & 52.13\sd{0.45} & 54.02\sd{0.35} & 38.22\sd{0.21} \\
& SAR    & 28.63\sd{0.26} & 31.01\sd{0.22} & 29.40\sd{0.32} & 18.46\sd{0.13} & 15.23\sd{0.28} & 28.82\sd{0.26} & 30.43\sd{0.34} & 44.75\sd{0.60} & 44.39\sd{0.12} & 33.60\sd{1.05} & 71.95\sd{0.09} & 44.61\sd{0.13} & 13.03\sd{1.06} & 47.83\sd{0.21} & 56.24\sd{0.04} & 35.89\sd{0.11} \\
& DeYO   & 40.87\sd{0.30} & 43.33\sd{0.18} & 42.08\sd{0.31} & 22.07\sd{0.26} & 23.69\sd{0.17} & 38.63\sd{0.25} & 28.12\sd{8.19} & 52.26\sd{0.31} & 50.21\sd{0.24} & 18.88\sd{25.72} & 73.19\sd{0.07} & 51.27\sd{0.18} & 42.97\sd{0.54} & 57.31\sd{0.18} & 58.54\sd{0.07} & 42.89\sd{1.95} \\
& ReCAP  & \underline{42.07}\sd{0.26} & \underline{44.35}\sd{0.29} & \underline{43.17}\sd{0.25} & 24.69\sd{0.27} & 25.05\sd{0.27} & 41.73\sd{0.30} & \underline{39.64}\sd{2.44} & \underline{54.67}\sd{0.35} & 52.16\sd{0.18} & \underline{46.54}\sd{22.42} & \underline{73.33}\sd{0.09} & 53.52\sd{0.14} & 41.54\sd{13.41} & \underline{60.27}\sd{0.24} & \underline{60.20}\sd{0.16} & \underline{46.86}\sd{1.52} \\
& SaTeen & \textbf{43.20}\sd{0.14} & \textbf{45.44}\sd{0.17} & \textbf{44.25}\sd{0.16} & \underline{25.80}\sd{0.45} & \underline{26.25}\sd{0.50} & \textbf{42.44}\sd{0.33} & 26.68\sd{10.40} & \textbf{55.29}\sd{0.12} & \underline{52.69}\sd{0.15} & 19.30\sd{27.47} & \textbf{73.34}\sd{0.08} & \textbf{54.00}\sd{0.13} & \underline{48.71}\sd{0.44} & \textbf{60.77}\sd{0.22} & 60.13\sd{0.17} & 45.22\sd{2.20} \\
\rowcolor{grn}\cellcolor{hdr}\multirow{-7}{*}{\rotatebox[origin=c]{90}{\textit{shuffled}}} & OriTTA & 41.11\sd{0.20} & 43.15\sd{0.12} & 42.22\sd{0.11} & \textbf{33.53}\sd{0.16} & \textbf{31.93}\sd{0.22} & \underline{42.38}\sd{0.15} & \textbf{46.33}\sd{0.15} & 54.32\sd{0.16} & \textbf{53.27}\sd{0.20} & \textbf{59.04}\sd{0.18} & 72.80\sd{0.06} & \underline{53.79}\sd{0.13} & \textbf{49.64}\sd{0.21} & 59.77\sd{0.17} & \textbf{60.62}\sd{0.11} & \textbf{49.59}\sd{0.05} \\
\midrule
& TENT   & 15.59\sd{3.67} & 22.05\sd{4.03} & 17.00\sd{3.45} & 13.21\sd{0.30} & 7.87\sd{4.97} & 18.40\sd{2.31} & 18.53\sd{0.91} & 17.01\sd{1.52} & 20.43\sd{1.09} & 1.50\sd{0.15} & 69.84\sd{0.13} & 42.12\sd{0.28} & 6.22\sd{0.60} & 49.45\sd{0.37} & 53.80\sd{0.21} & 24.87\sd{0.80} \\
& EATA   & 25.27\sd{0.50} & 25.71\sd{1.05} & 25.28\sd{0.45} & 15.68\sd{0.78} & 14.93\sd{0.60} & 21.35\sd{0.98} & 22.70\sd{0.51} & 33.11\sd{0.70} & 32.90\sd{0.95} & \underline{37.47}\sd{0.51} & 65.32\sd{0.63} & 34.82\sd{0.80} & 18.24\sd{0.76} & 38.49\sd{1.03} & 50.43\sd{0.62} & 30.78\sd{0.19} \\
& SAR    & 28.53\sd{0.18} & 31.11\sd{0.31} & 29.44\sd{0.37} & 18.56\sd{0.44} & 15.14\sd{0.47} & 28.83\sd{0.22} & 30.53\sd{0.22} & 44.16\sd{1.40} & 44.36\sd{0.22} & 33.30\sd{1.52} & 71.93\sd{0.05} & 44.67\sd{0.17} & 13.87\sd{1.72} & 47.73\sd{0.30} & 56.22\sd{0.08} & 35.89\sd{0.15} \\
& DeYO   & 40.89\sd{0.16} & 43.18\sd{0.19} & 42.19\sd{0.24} & 21.84\sd{0.30} & 23.44\sd{0.38} & 38.64\sd{0.50} & 33.34\sd{7.50} & 52.08\sd{0.58} & 50.35\sd{0.39} & 18.67\sd{25.79} & 73.19\sd{0.09} & 51.26\sd{0.17} & 42.93\sd{0.74} & 57.44\sd{0.31} & 58.51\sd{0.14} & 43.20\sd{1.74} \\
& ReCAP  & \underline{42.13}\sd{0.24} & \underline{44.41}\sd{0.35} & \underline{43.39}\sd{0.38} & 24.72\sd{0.47} & 24.72\sd{0.62} & 41.78\sd{0.39} & \underline{39.92}\sd{2.35} & \underline{54.62}\sd{0.40} & 52.25\sd{0.36} & 35.69\sd{27.54} & \textbf{73.44}\sd{0.12} & \underline{53.63}\sd{0.27} & 42.67\sd{11.09} & \underline{60.32}\sd{0.42} & \textbf{60.22}\sd{0.10} & \underline{46.26}\sd{2.10} \\
& SaTeen & \textbf{43.13}\sd{0.15} & \textbf{45.48}\sd{0.20} & \textbf{44.31}\sd{0.29} & \underline{25.58}\sd{0.54} & \underline{25.96}\sd{0.56} & \textbf{42.34}\sd{0.26} & 27.82\sd{8.78} & \textbf{55.28}\sd{0.37} & \textbf{52.74}\sd{0.27} & 13.48\sd{23.72} & \underline{73.41}\sd{0.12} & \textbf{54.00}\sd{0.14} & \textbf{48.85}\sd{0.85} & \textbf{61.00}\sd{0.36} & \underline{60.14}\sd{0.10} & 44.90\sd{1.84} \\
\rowcolor{grn}\cellcolor{hdr}\multirow{-7}{*}{\rotatebox[origin=c]{90}{\textit{class-blocked}}} & OriTTA & 40.79\sd{0.17} & 42.55\sd{0.24} & 41.81\sd{0.23} & \textbf{33.32}\sd{0.29} & \textbf{31.09}\sd{0.38} & \underline{41.83}\sd{0.22} & \textbf{45.80}\sd{0.13} & 53.69\sd{0.20} & \underline{52.67}\sd{0.15} & \textbf{57.92}\sd{0.36} & 72.50\sd{0.11} & 53.12\sd{0.29} & \underline{48.79}\sd{0.35} & 58.73\sd{0.21} & 60.05\sd{0.09} & \textbf{48.98}\sd{0.10} \\
\midrule\midrule
\rowcolor{hdr} & & gaus & shot & impu & defo & glas & moti & zoom & snow & fros & fog & brit & cont & elas & pixe & jpeg & Average \\
\rowcolor{hdr} & continual & \multicolumn{15}{c}{\streamarrow} & \\
\midrule\midrule
& TENT   & 12.35\sd{2.55} & 0.74\sd{0.23} & 0.22\sd{0.05} & 2.16\sd{0.58} & 0.26\sd{0.09} & 0.22\sd{0.02} & 0.31\sd{0.06} & 0.57\sd{0.05} & 0.18\sd{0.02} & 0.20\sd{0.01} & 1.48\sd{0.49} & 0.31\sd{0.15} & 0.10\sd{0.00} & 0.16\sd{0.01} & 0.79\sd{0.20} & 1.34\sd{0.28} \\
& EATA   & 29.49\sd{1.04} & 29.30\sd{3.85} & 25.44\sd{4.01} & 12.94\sd{2.20} & 14.77\sd{2.13} & \underline{18.41}\sd{3.53} & \underline{23.08}\sd{3.19} & \underline{26.36}\sd{4.22} & \underline{27.85}\sd{5.39} & \underline{31.02}\sd{6.70} & \underline{54.08}\sd{2.27} & \underline{24.44}\sd{2.40} & \underline{21.43}\sd{2.36} & \underline{32.12}\sd{6.61} & \underline{38.49}\sd{2.63} & \underline{27.28}\sd{1.91} \\
& SAR    & 28.63\sd{0.25} & 41.23\sd{0.14} & 44.45\sd{0.07} & 15.66\sd{0.25} & \underline{19.02}\sd{0.35} & 9.66\sd{0.78} & 10.66\sd{2.84} & 5.55\sd{6.76} & 3.04\sd{1.35} & 1.21\sd{0.36} & 6.33\sd{0.65} & 1.39\sd{0.73} & 0.48\sd{0.38} & 0.98\sd{0.43} & 1.94\sd{0.70} & 12.68\sd{0.59} \\
& DeYO   & 40.87\sd{0.30} & 48.80\sd{0.09} & \underline{47.90}\sd{0.12} & 10.76\sd{1.03} & 13.74\sd{0.80} & 13.96\sd{4.12} & 6.48\sd{3.75} & 1.53\sd{0.75} & 0.73\sd{0.18} & 0.44\sd{0.11} & 1.76\sd{0.47} & 0.26\sd{0.05} & 0.20\sd{0.03} & 0.19\sd{0.03} & 0.23\sd{0.03} & 12.52\sd{0.55} \\
& ReCAP  & \underline{42.03}\sd{0.25} & \textbf{49.33}\sd{0.16} & \textbf{48.41}\sd{0.17} & \textbf{24.02}\sd{7.68} & 3.50\sd{4.30} & 3.91\sd{8.72} & 2.70\sd{6.65} & 3.55\sd{10.58} & 3.85\sd{11.75} & 5.13\sd{15.80} & 6.98\sd{21.51} & 4.96\sd{15.30} & 1.72\sd{5.02} & 4.31\sd{13.19} & 4.65\sd{14.27} & 13.94\sd{8.28} \\
& SaTeen & \textbf{43.20}\sd{0.14} & \underline{48.90}\sd{0.19} & 47.61\sd{0.20} & 19.60\sd{8.54} & 1.17\sd{1.70} & 0.19\sd{0.07} & 0.19\sd{0.12} & 0.16\sd{0.07} & 0.12\sd{0.02} & 0.11\sd{0.02} & 0.12\sd{0.03} & 0.11\sd{0.02} & 0.11\sd{0.02} & 0.11\sd{0.02} & 0.11\sd{0.02} & 10.79\sd{0.67} \\
\rowcolor{grn}\cellcolor{hdr}\multirow{-7}{*}{\rotatebox[origin=c]{90}{\textit{shuffled}}} & OriTTA & 41.09\sd{0.19} & 43.78\sd{0.74} & 38.40\sd{0.98} & \underline{22.63}\sd{5.79} & \textbf{21.81}\sd{3.38} & \textbf{36.39}\sd{3.28} & \textbf{35.26}\sd{1.66} & \textbf{50.72}\sd{2.73} & \textbf{48.64}\sd{0.59} & \textbf{42.97}\sd{1.25} & \textbf{63.05}\sd{3.33} & \textbf{39.22}\sd{4.98} & \textbf{40.62}\sd{7.53} & \textbf{33.09}\sd{13.57} & \textbf{54.44}\sd{9.66} & \textbf{40.81}\sd{1.02} \\
\midrule
& TENT   & 15.59\sd{3.67} & 1.08\sd{0.43} & 0.27\sd{0.07} & 2.73\sd{0.66} & 0.36\sd{0.11} & 0.27\sd{0.05} & 0.38\sd{0.07} & 0.63\sd{0.10} & 0.19\sd{0.02} & 0.22\sd{0.02} & 1.97\sd{0.52} & 0.41\sd{0.11} & 0.10\sd{0.00} & 0.17\sd{0.02} & 0.97\sd{0.22} & 1.69\sd{0.39} \\
& EATA   & 25.27\sd{0.50} & 11.04\sd{2.07} & 7.49\sd{0.99} & 6.19\sd{1.62} & 5.24\sd{2.38} & 5.26\sd{1.45} & 6.21\sd{2.29} & 4.71\sd{1.22} & 5.98\sd{1.98} & 5.36\sd{2.90} & 12.14\sd{4.26} & 2.29\sd{1.25} & 4.36\sd{2.28} & 5.29\sd{1.61} & 6.47\sd{2.75} & 7.55\sd{1.19} \\
& SAR    & 28.53\sd{0.18} & 41.14\sd{0.14} & 44.47\sd{0.09} & 15.76\sd{0.30} & \underline{19.17}\sd{0.41} & 11.52\sd{0.61} & \underline{11.18}\sd{5.27} & \underline{15.79}\sd{15.25} & \underline{18.37}\sd{19.91} & \underline{18.04}\sd{24.14} & \underline{34.59}\sd{28.87} & \underline{20.35}\sd{23.10} & \underline{5.68}\sd{6.21} & \underline{18.72}\sd{22.66} & \underline{20.27}\sd{24.03} & \underline{21.57}\sd{10.86} \\
& DeYO   & 40.89\sd{0.16} & 48.73\sd{0.11} & \underline{47.86}\sd{0.12} & 11.55\sd{4.05} & 11.50\sd{4.22} & \underline{11.82}\sd{5.74} & 6.89\sd{4.78} & 1.88\sd{1.22} & 0.77\sd{0.39} & 0.47\sd{0.19} & 2.02\sd{1.40} & 0.27\sd{0.09} & 0.20\sd{0.07} & 0.20\sd{0.05} & 0.24\sd{0.10} & 12.35\sd{0.84} \\
& ReCAP  & \underline{42.15}\sd{0.26} & \textbf{49.28}\sd{0.22} & \textbf{48.43}\sd{0.18} & 18.17\sd{10.74} & 3.23\sd{5.37} & 0.81\sd{1.51} & 0.31\sd{0.22} & 0.26\sd{0.13} & 0.17\sd{0.04} & 0.15\sd{0.04} & 0.24\sd{0.08} & 0.14\sd{0.01} & 0.13\sd{0.03} & 0.15\sd{0.02} & 0.17\sd{0.06} & 10.92\sd{1.02} \\
& SaTeen & \textbf{43.13}\sd{0.15} & \underline{48.78}\sd{0.23} & 47.57\sd{0.18} & \underline{24.45}\sd{5.96} & 3.17\sd{4.71} & 3.59\sd{9.03} & 2.04\sd{5.38} & 2.06\sd{5.88} & 0.95\sd{2.65} & 0.24\sd{0.38} & 0.16\sd{0.12} & 0.10\sd{0.01} & 0.12\sd{0.02} & 0.10\sd{0.04} & 0.10\sd{0.02} & 11.77\sd{1.78} \\
\rowcolor{grn}\cellcolor{hdr}\multirow{-7}{*}{\rotatebox[origin=c]{90}{\textit{class-blocked}}} & OriTTA & 40.74\sd{0.16} & 44.45\sd{0.37} & 39.26\sd{1.70} & \textbf{25.80}\sd{6.03} & \textbf{24.91}\sd{5.43} & \textbf{33.45}\sd{5.97} & \textbf{32.44}\sd{5.53} & \textbf{52.02}\sd{2.27} & \textbf{48.77}\sd{1.37} & \textbf{45.97}\sd{4.55} & \textbf{66.77}\sd{5.31} & \textbf{45.32}\sd{4.59} & \textbf{39.96}\sd{4.40} & \textbf{41.61}\sd{4.67} & \textbf{48.27}\sd{4.20} & \textbf{41.98}\sd{0.78} \\
\bottomrule
\end{tabular}}
\end{sidewaystable}

\begin{sidewaystable}
\centering\scriptsize
\setlength{\tabcolsep}{2.5pt}
\caption{Full results on ImageNet-C, ViT-B/16.}
\label{tab:full_vit}
\resizebox{\linewidth}{!}{%
\begin{tabular}{>{\columncolor{hdr}}c l ccccccccccccccc | c}
\toprule
\rowcolor{hdr} & episodic & gaus & shot & impu & defo & glas & moti & zoom & snow & fros & fog & brit & cont & elas & pixe & jpeg & Average \\
\midrule\midrule
& TENT   & 44.66\sd{0.22} & 42.82\sd{0.23} & 45.42\sd{0.27} & 52.51\sd{0.17} & 48.10\sd{0.22} & 55.55\sd{0.14} & 50.49\sd{0.28} & 15.81\sd{0.91} & 17.72\sd{1.17} & 66.51\sd{0.27} & 75.09\sd{0.07} & 64.82\sd{0.12} & 53.07\sd{0.34} & 67.04\sd{0.06} & 64.57\sd{0.09} & 50.95\sd{0.13} \\
& EATA   & 50.24\sd{0.27} & 49.48\sd{0.34} & 51.22\sd{0.21} & 56.69\sd{0.08} & 55.76\sd{0.18} & 61.04\sd{0.16} & \underline{57.94}\sd{0.16} & 64.05\sd{0.21} & 63.74\sd{0.19} & 72.08\sd{0.12} & 77.47\sd{0.07} & 67.50\sd{0.19} & 64.61\sd{0.12} & 71.78\sd{0.09} & 69.11\sd{0.08} & 62.18\sd{0.07} \\
& SAR    & 45.39\sd{0.26} & 42.95\sd{0.47} & 46.23\sd{0.33} & 53.68\sd{0.12} & 50.47\sd{0.22} & 57.64\sd{0.17} & 52.94\sd{0.14} & 58.78\sd{0.33} & 56.00\sd{4.06} & 68.57\sd{0.77} & 75.28\sd{0.94} & 65.65\sd{0.08} & 58.16\sd{0.24} & 68.67\sd{0.87} & 66.41\sd{0.11} & 57.79\sd{0.32} \\
& DeYO   & 53.01\sd{0.21} & 52.75\sd{0.18} & 53.99\sd{0.09} & 57.66\sd{0.08} & 57.44\sd{0.15} & 62.29\sd{0.12} & 49.17\sd{9.60} & 65.76\sd{0.16} & 64.78\sd{0.11} & 72.63\sd{0.11} & 77.75\sd{0.09} & 67.89\sd{0.13} & 66.19\sd{0.14} & 72.55\sd{0.08} & 69.78\sd{0.06} & 62.91\sd{0.62} \\
& ReCAP  & 55.04\sd{0.22} & 55.19\sd{0.34} & 56.05\sd{0.17} & \underline{59.02}\sd{0.16} & 59.45\sd{0.14} & 64.69\sd{0.11} & 56.49\sd{9.32} & 68.12\sd{0.16} & \underline{66.60}\sd{0.21} & \underline{73.83}\sd{0.10} & \textbf{78.43}\sd{0.05} & \underline{68.07}\sd{0.12} & 68.97\sd{0.10} & \underline{74.06}\sd{0.08} & \underline{71.04}\sd{0.11} & \underline{65.00}\sd{0.66} \\
& SaTeen & \underline{55.77}\sd{0.20} & \underline{56.47}\sd{0.17} & \textbf{57.01}\sd{0.21} & \textbf{59.13}\sd{0.08} & \textbf{59.84}\sd{0.15} & \textbf{65.05}\sd{0.13} & 45.92\sd{12.01} & \textbf{68.75}\sd{0.09} & \textbf{66.89}\sd{0.18} & \textbf{73.99}\sd{0.11} & \underline{78.38}\sd{0.05} & 68.04\sd{0.15} & \underline{69.62}\sd{0.15} & \textbf{74.17}\sd{0.12} & \textbf{71.21}\sd{0.12} & 64.68\sd{0.81} \\
\rowcolor{grn}\cellcolor{hdr}\multirow{-7}{*}{\rotatebox[origin=c]{90}{\textit{shuffled}}} & OriTTA & \textbf{55.81}\sd{0.12} & \textbf{56.77}\sd{0.18} & \underline{56.80}\sd{0.10} & 58.64\sd{0.15} & \underline{59.64}\sd{0.13} & \underline{64.92}\sd{0.08} & \textbf{64.64}\sd{0.15} & \underline{68.39}\sd{0.18} & 66.53\sd{0.27} & 73.26\sd{0.09} & 76.90\sd{0.07} & \textbf{68.13}\sd{0.12} & \textbf{70.01}\sd{0.15} & 73.13\sd{0.11} & 70.11\sd{0.15} & \textbf{65.58}\sd{0.05} \\
\midrule
& TENT   & 44.75\sd{0.23} & 42.94\sd{0.37} & 45.57\sd{0.14} & 52.63\sd{0.15} & 48.07\sd{0.09} & 55.66\sd{0.08} & 50.46\sd{0.19} & 16.91\sd{2.45} & 19.09\sd{1.99} & 66.55\sd{0.05} & 75.08\sd{0.14} & 64.90\sd{0.09} & 52.86\sd{0.85} & 67.05\sd{0.16} & 64.61\sd{0.16} & 51.14\sd{0.23} \\
& EATA   & 29.56\sd{0.28} & 27.25\sd{0.97} & 30.81\sd{0.19} & 43.45\sd{0.42} & 36.96\sd{0.45} & 46.07\sd{0.49} & 39.27\sd{0.69} & 44.09\sd{0.78} & 47.04\sd{0.56} & 57.73\sd{0.29} & 69.89\sd{0.23} & 54.69\sd{1.33} & 42.08\sd{0.53} & 58.61\sd{0.50} & 56.93\sd{0.23} & 45.63\sd{0.10} \\
& SAR    & 33.03\sd{2.20} & 30.51\sd{2.78} & 34.86\sd{2.01} & 43.40\sd{0.98} & 37.71\sd{1.71} & 46.28\sd{0.81} & 39.38\sd{0.97} & 38.64\sd{0.91} & 42.71\sd{1.89} & 54.24\sd{0.58} & 63.50\sd{0.21} & 55.24\sd{0.41} & 38.96\sd{0.70} & 54.65\sd{0.29} & 55.63\sd{0.38} & 44.58\sd{0.42} \\
& DeYO   & 53.27\sd{0.19} & 53.07\sd{0.18} & 54.23\sd{0.21} & 57.83\sd{0.25} & 57.68\sd{0.10} & 62.55\sd{0.11} & 47.21\sd{16.81} & 66.09\sd{0.13} & 64.79\sd{0.20} & 72.68\sd{0.12} & 77.77\sd{0.09} & \underline{67.95}\sd{0.16} & 66.36\sd{0.20} & 72.62\sd{0.21} & 69.86\sd{0.15} & 62.93\sd{1.09} \\
& ReCAP  & 55.38\sd{0.38} & 55.72\sd{0.34} & 56.46\sd{0.45} & \underline{59.47}\sd{0.30} & 59.89\sd{0.33} & 65.03\sd{0.22} & 38.03\sd{24.44} & 68.49\sd{0.23} & 66.97\sd{0.32} & 73.99\sd{0.15} & \textbf{78.53}\sd{0.05} & 67.29\sd{2.62} & 69.40\sd{0.18} & \underline{74.29}\sd{0.13} & \underline{71.37}\sd{0.15} & 64.02\sd{1.56} \\
& SaTeen & \underline{56.40}\sd{0.23} & \underline{57.19}\sd{0.22} & \underline{57.42}\sd{0.29} & \textbf{59.73}\sd{0.32} & \underline{60.34}\sd{0.21} & \textbf{65.42}\sd{0.20} & \underline{54.36}\sd{11.76} & \textbf{69.07}\sd{0.13} & \textbf{67.21}\sd{0.30} & \textbf{74.12}\sd{0.11} & \underline{78.52}\sd{0.04} & 58.86\sd{21.65} & \underline{70.02}\sd{0.17} & \textbf{74.38}\sd{0.10} & \textbf{71.51}\sd{0.17} & \underline{64.97}\sd{1.20} \\
\rowcolor{grn}\cellcolor{hdr}\multirow{-7}{*}{\rotatebox[origin=c]{90}{\textit{class-blocked}}} & OriTTA & \textbf{56.41}\sd{0.18} & \textbf{57.36}\sd{0.21} & \textbf{57.47}\sd{0.13} & 59.35\sd{0.26} & \textbf{60.41}\sd{0.15} & \underline{65.29}\sd{0.28} & \textbf{64.85}\sd{0.33} & \underline{68.97}\sd{0.19} & \underline{67.15}\sd{0.14} & \underline{74.05}\sd{0.28} & 77.85\sd{0.15} & \textbf{69.14}\sd{0.24} & \textbf{70.45}\sd{0.20} & 73.77\sd{0.11} & 70.98\sd{0.21} & \textbf{66.23}\sd{0.10} \\
\midrule\midrule
\rowcolor{hdr} & & gaus & shot & impu & defo & glas & moti & zoom & snow & fros & fog & brit & cont & elas & pixe & jpeg & Average \\
\rowcolor{hdr} & continual & \multicolumn{15}{c}{\streamarrow} & \\
\midrule\midrule
& TENT   & 39.90\sd{0.22} & 47.44\sd{0.10} & 52.29\sd{0.09} & 47.79\sd{0.23} & 49.56\sd{0.14} & 54.62\sd{0.10} & 51.26\sd{0.13} & 56.18\sd{0.15} & 59.32\sd{0.15} & 63.91\sd{0.19} & 75.06\sd{0.05} & 61.24\sd{0.15} & 37.04\sd{7.54} & 14.46\sd{19.01} & 7.75\sd{20.28} & 47.85\sd{2.96} \\
& EATA   & 45.83\sd{0.22} & 54.55\sd{0.22} & 56.06\sd{0.13} & 52.55\sd{0.14} & 54.77\sd{0.33} & 57.62\sd{0.21} & 55.17\sd{0.27} & 59.07\sd{0.13} & 61.47\sd{0.22} & 68.02\sd{0.34} & 75.56\sd{0.14} & 62.37\sd{0.33} & 60.23\sd{0.49} & 68.90\sd{0.35} & 66.26\sd{0.29} & 59.90\sd{0.13} \\
& SAR    & 38.79\sd{0.27} & 50.85\sd{0.15} & 55.91\sd{0.07} & 50.59\sd{0.13} & 54.02\sd{0.08} & 58.05\sd{0.08} & \underline{55.18}\sd{0.10} & 59.52\sd{0.07} & 62.75\sd{0.06} & 66.93\sd{0.80} & 75.79\sd{0.21} & 63.05\sd{0.50} & 60.78\sd{0.66} & 67.97\sd{0.91} & 67.77\sd{0.38} & 59.20\sd{0.10} \\
& DeYO   & 49.73\sd{0.16} & 56.47\sd{0.10} & \underline{58.31}\sd{0.05} & \underline{53.95}\sd{0.13} & \underline{57.11}\sd{0.14} & 60.30\sd{0.14} & 36.52\sd{7.43} & 60.17\sd{0.89} & 63.06\sd{0.17} & 69.37\sd{0.16} & 76.73\sd{0.05} & 64.74\sd{0.11} & 64.67\sd{0.19} & 70.68\sd{0.14} & 68.73\sd{0.10} & 60.70\sd{0.54} \\
& ReCAP  & \underline{52.51}\sd{0.33} & \textbf{59.09}\sd{0.10} & \textbf{59.76}\sd{0.12} & \textbf{55.32}\sd{0.30} & \textbf{59.24}\sd{0.12} & \textbf{62.61}\sd{0.21} & 20.24\sd{9.21} & \underline{62.45}\sd{3.96} & \textbf{65.01}\sd{0.58} & \textbf{71.22}\sd{0.15} & \textbf{77.24}\sd{0.09} & \textbf{65.84}\sd{0.14} & \textbf{67.68}\sd{0.20} & \textbf{72.28}\sd{0.09} & \textbf{70.04}\sd{0.10} & 61.37\sd{0.82} \\
& SaTeen & 46.13\sd{0.23} & 55.79\sd{0.08} & 58.15\sd{0.08} & 53.39\sd{0.21} & 56.93\sd{0.14} & 60.53\sd{0.15} & 51.76\sd{3.19} & 61.61\sd{1.18} & 63.87\sd{0.25} & 69.76\sd{0.13} & \underline{77.09}\sd{0.04} & \underline{65.17}\sd{0.21} & 64.34\sd{0.16} & 70.87\sd{0.11} & \underline{68.90}\sd{0.07} & \underline{61.62}\sd{0.32} \\
\rowcolor{grn}\cellcolor{hdr}\multirow{-7}{*}{\rotatebox[origin=c]{90}{\textit{shuffled}}} & OriTTA & \textbf{55.78}\sd{0.10} & \underline{58.03}\sd{0.22} & 57.19\sd{0.48} & 51.37\sd{5.18} & 55.14\sd{2.65} & \underline{60.66}\sd{1.89} & \textbf{60.56}\sd{1.46} & \textbf{63.17}\sd{5.87} & \underline{64.10}\sd{2.09} & \underline{70.71}\sd{0.83} & 75.57\sd{0.38} & 65.13\sd{1.39} & \underline{66.68}\sd{6.03} & \underline{71.20}\sd{0.37} & 68.42\sd{0.16} & \textbf{62.91}\sd{0.96} \\
\midrule
& TENT   & 40.13\sd{0.07} & 47.48\sd{0.18} & 52.29\sd{0.13} & 47.75\sd{0.41} & 49.75\sd{0.23} & 54.64\sd{0.11} & 51.17\sd{0.27} & 56.29\sd{0.14} & 59.36\sd{0.05} & 63.93\sd{0.22} & 75.06\sd{0.02} & 61.34\sd{0.19} & 39.92\sd{7.46} & 19.51\sd{25.62} & 14.19\sd{28.78} & 48.85\sd{4.09} \\
& EATA   & 26.49\sd{0.12} & 30.74\sd{0.44} & 27.36\sd{1.44} & 16.04\sd{2.10} & 6.15\sd{1.90} & 4.16\sd{0.41} & 3.02\sd{0.60} & 3.98\sd{0.57} & 3.46\sd{0.99} & 1.37\sd{0.44} & 6.34\sd{0.58} & 0.49\sd{0.13} & 1.98\sd{0.96} & 1.85\sd{0.83} & 2.47\sd{0.75} & 9.06\sd{0.31} \\
& SAR    & 30.92\sd{1.37} & 38.33\sd{3.45} & 44.52\sd{1.97} & 42.10\sd{0.59} & 41.76\sd{0.44} & 50.25\sd{1.42} & 44.51\sd{1.07} & 47.98\sd{1.47} & 53.06\sd{1.21} & 59.97\sd{0.75} & 74.45\sd{0.70} & 56.31\sd{0.89} & 46.65\sd{0.95} & 62.30\sd{0.69} & 63.05\sd{0.71} & 50.41\sd{0.92} \\
& DeYO   & 49.95\sd{0.30} & \underline{56.57}\sd{0.12} & \underline{58.28}\sd{0.07} & 53.97\sd{0.19} & \underline{57.05}\sd{0.13} & 60.25\sd{0.19} & 34.97\sd{5.42} & 59.70\sd{0.87} & 63.04\sd{0.17} & 69.53\sd{0.13} & 76.75\sd{0.08} & 64.73\sd{0.17} & 64.82\sd{0.24} & 70.62\sd{0.14} & 68.72\sd{0.15} & 60.60\sd{0.34} \\
& ReCAP  & \underline{52.78}\sd{0.49} & \textbf{59.16}\sd{0.16} & \textbf{59.82}\sd{0.15} & \textbf{55.59}\sd{0.33} & \textbf{59.21}\sd{0.20} & \textbf{62.57}\sd{0.21} & 19.77\sd{4.96} & 59.41\sd{6.83} & \underline{64.23}\sd{1.52} & \textbf{71.13}\sd{0.35} & \textbf{77.21}\sd{0.13} & \textbf{65.74}\sd{0.37} & \textbf{67.82}\sd{0.24} & \textbf{72.33}\sd{0.21} & \textbf{70.15}\sd{0.09} & 61.13\sd{0.75} \\
& SaTeen & 46.37\sd{0.37} & 55.79\sd{0.14} & 58.16\sd{0.06} & 53.42\sd{0.39} & 56.89\sd{0.12} & \underline{60.52}\sd{0.15} & \underline{51.97}\sd{5.58} & \underline{61.39}\sd{2.09} & 63.92\sd{0.36} & \underline{69.90}\sd{0.18} & \underline{77.07}\sd{0.02} & \underline{65.12}\sd{0.08} & 64.40\sd{0.32} & \underline{70.83}\sd{0.18} & \underline{68.87}\sd{0.12} & \underline{61.64}\sd{0.56} \\
\rowcolor{grn}\cellcolor{hdr}\multirow{-7}{*}{\rotatebox[origin=c]{90}{\textit{class-blocked}}} & OriTTA & \textbf{56.34}\sd{0.23} & 56.07\sd{0.07} & 54.79\sd{0.53} & \underline{55.22}\sd{0.46} & 56.78\sd{0.58} & 58.58\sd{0.78} & \textbf{55.88}\sd{0.22} & \textbf{64.50}\sd{1.49} & \textbf{64.68}\sd{0.21} & 68.39\sd{0.49} & 74.28\sd{0.09} & 64.03\sd{5.46} & \underline{65.55}\sd{0.39} & 69.28\sd{1.27} & 65.61\sd{0.47} & \textbf{62.00}\sd{0.30} \\
\bottomrule
\end{tabular}}
\end{sidewaystable}

\begin{sidewaystable}
\centering\scriptsize
\setlength{\tabcolsep}{2.5pt}
\caption{Full results on CIFAR-100-C, WRN-28-10.}
\label{tab:full_c100}
\resizebox{\linewidth}{!}{%
\begin{tabular}{>{\columncolor{hdr}}c l ccccccccccccccc | c}
\toprule
\rowcolor{hdr} & episodic & gaus & shot & impu & defo & glas & moti & zoom & snow & fros & fog & brit & cont & elas & pixe & jpeg & Average \\
\midrule\midrule
& TENT   & 4.03\sd{1.15} & 4.60\sd{0.97} & 3.90\sd{0.75} & 53.51\sd{8.67} & 4.68\sd{2.49} & 44.72\sd{15.34} & 58.34\sd{7.00} & 26.88\sd{10.40} & 19.21\sd{7.94} & 6.87\sd{3.52} & \underline{68.55}\sd{0.91} & 2.53\sd{0.81} & 27.61\sd{7.16} & 39.61\sd{17.44} & 22.44\sd{7.88} & 25.83\sd{1.73} \\
& EATA   & \underline{44.73}\sd{1.21} & \underline{47.42}\sd{1.21} & \underline{48.71}\sd{0.81} & \underline{63.27}\sd{0.98} & \underline{46.40}\sd{0.80} & \underline{62.02}\sd{1.15} & \underline{63.46}\sd{1.01} & \underline{54.89}\sd{1.08} & \underline{55.22}\sd{0.77} & \underline{49.52}\sd{1.42} & 66.46\sd{0.65} & \underline{52.04}\sd{1.11} & \underline{52.99}\sd{1.01} & \underline{59.59}\sd{0.91} & \underline{50.59}\sd{0.77} & \underline{54.49}\sd{0.27} \\
& SAR    & 24.67\sd{0.89} & 29.86\sd{1.35} & 29.44\sd{1.00} & 62.24\sd{1.99} & 35.97\sd{0.81} & 60.83\sd{1.04} & 62.11\sd{1.05} & 50.20\sd{1.32} & 46.89\sd{1.85} & 36.85\sd{1.11} & 64.84\sd{0.78} & 28.90\sd{1.29} & 50.81\sd{1.12} & 38.01\sd{2.85} & 47.08\sd{1.20} & 44.58\sd{0.24} \\
& DeYO   & 5.46\sd{1.93} & 5.17\sd{2.84} & 6.98\sd{2.91} & 15.89\sd{2.92} & 2.79\sd{1.47} & 12.37\sd{3.86} & 14.44\sd{4.73} & 13.04\sd{3.39} & 13.24\sd{3.69} & 6.98\sd{3.35} & 25.65\sd{5.01} & 3.35\sd{1.50} & 8.06\sd{2.58} & 15.93\sd{5.48} & 8.00\sd{3.03} & 10.49\sd{1.02} \\
& ReCAP  & 1.44\sd{0.14} & 1.68\sd{0.24} & 1.50\sd{0.41} & 2.88\sd{0.74} & 1.52\sd{0.36} & 2.37\sd{0.70} & 3.10\sd{0.89} & 2.48\sd{0.58} & 2.04\sd{0.43} & 1.30\sd{0.18} & 3.16\sd{0.81} & 1.33\sd{0.16} & 2.16\sd{0.56} & 2.43\sd{0.70} & 1.95\sd{0.34} & 2.09\sd{0.15} \\
& SaTeen & 1.38\sd{0.13} & 1.50\sd{0.13} & 1.76\sd{0.42} & 2.16\sd{0.47} & 1.38\sd{0.24} & 2.33\sd{0.37} & 2.16\sd{0.22} & 2.12\sd{0.27} & 2.01\sd{0.33} & 1.37\sd{0.15} & 3.13\sd{0.32} & 1.30\sd{0.13} & 1.98\sd{0.52} & 2.28\sd{0.52} & 1.91\sd{0.24} & 1.92\sd{0.10} \\
\rowcolor{grn}\cellcolor{hdr}\multirow{-7}{*}{\rotatebox[origin=c]{90}{\textit{shuffled}}} & OriTTA & \textbf{49.66}\sd{0.44} & \textbf{53.17}\sd{0.20} & \textbf{53.11}\sd{0.19} & \textbf{70.64}\sd{0.15} & \textbf{55.62}\sd{0.25} & \textbf{69.51}\sd{0.16} & \textbf{70.61}\sd{0.20} & \textbf{63.31}\sd{0.17} & \textbf{63.29}\sd{0.20} & \textbf{59.11}\sd{0.22} & \textbf{71.84}\sd{0.20} & \textbf{64.81}\sd{0.21} & \textbf{62.74}\sd{0.22} & \textbf{65.62}\sd{0.33} & \textbf{59.42}\sd{0.16} & \textbf{62.16}\sd{0.09} \\
\midrule
& TENT   & 3.28\sd{1.82} & 5.01\sd{2.23} & 4.78\sd{1.72} & 36.34\sd{7.30} & 8.38\sd{3.56} & 28.48\sd{8.12} & 32.73\sd{8.61} & 17.32\sd{4.37} & 16.53\sd{9.61} & 6.87\sd{3.00} & 52.51\sd{9.06} & 9.24\sd{1.93} & 14.53\sd{5.00} & 20.90\sd{6.39} & 15.81\sd{4.63} & 18.18\sd{1.18} \\
& EATA   & \underline{32.80}\sd{2.45} & \underline{38.07}\sd{0.80} & \underline{36.74}\sd{2.29} & 63.44\sd{2.59} & \underline{39.73}\sd{2.76} & 60.96\sd{1.86} & 62.19\sd{1.41} & 52.56\sd{0.52} & \underline{51.89}\sd{1.54} & \underline{43.65}\sd{2.93} & 63.28\sd{2.34} & \underline{40.28}\sd{5.61} & \underline{53.51}\sd{1.54} & \underline{45.24}\sd{3.53} & 50.40\sd{1.22} & \underline{48.98}\sd{0.44} \\
& SAR    & 27.36\sd{1.30} & 32.52\sd{1.54} & 33.40\sd{1.84} & \underline{65.13}\sd{1.53} & 38.63\sd{1.31} & \underline{62.15}\sd{1.94} & \underline{64.33}\sd{1.56} & \underline{52.60}\sd{1.11} & 48.14\sd{1.51} & 39.45\sd{1.79} & \underline{65.72}\sd{0.67} & 33.64\sd{1.20} & 51.95\sd{2.06} & 40.97\sd{1.53} & \underline{50.56}\sd{1.14} & 47.10\sd{0.56} \\
& DeYO   & 8.10\sd{2.17} & 7.67\sd{1.88} & 9.85\sd{3.54} & 17.42\sd{6.80} & 4.02\sd{0.99} & 9.64\sd{2.51} & 20.26\sd{6.92} & 10.43\sd{2.71} & 14.73\sd{2.33} & 10.04\sd{2.52} & 29.72\sd{2.99} & 8.78\sd{2.64} & 8.17\sd{2.45} & 16.55\sd{5.99} & 11.79\sd{3.45} & 12.48\sd{1.56} \\
& ReCAP  & 3.52\sd{1.23} & 4.08\sd{3.63} & 3.98\sd{2.69} & 11.36\sd{9.23} & 2.25\sd{1.36} & 7.05\sd{1.75} & 15.41\sd{9.82} & 7.91\sd{4.67} & 6.86\sd{4.61} & 3.83\sd{2.58} & 24.43\sd{9.66} & 2.81\sd{0.51} & 5.40\sd{2.11} & 7.00\sd{5.92} & 6.37\sd{2.16} & 7.48\sd{1.58} \\
& SaTeen & 3.54\sd{2.14} & 3.84\sd{2.82} & 2.94\sd{1.91} & 6.81\sd{3.58} & 3.20\sd{1.71} & 6.33\sd{1.21} & 6.36\sd{2.36} & 7.07\sd{2.24} & 5.79\sd{1.69} & 2.17\sd{1.19} & 7.97\sd{2.71} & 3.50\sd{2.11} & 4.23\sd{1.21} & 5.14\sd{1.07} & 5.17\sd{1.80} & 4.94\sd{0.81} \\
\rowcolor{grn}\cellcolor{hdr}\multirow{-7}{*}{\rotatebox[origin=c]{90}{\textit{class-blocked}}} & OriTTA & \textbf{48.07}\sd{0.53} & \textbf{52.29}\sd{0.49} & \textbf{51.20}\sd{0.33} & \textbf{70.98}\sd{0.41} & \textbf{54.74}\sd{0.63} & \textbf{69.73}\sd{0.55} & \textbf{70.81}\sd{0.30} & \textbf{63.13}\sd{0.42} & \textbf{63.30}\sd{0.40} & \textbf{58.71}\sd{0.59} & \textbf{71.93}\sd{0.34} & \textbf{63.51}\sd{0.72} & \textbf{62.84}\sd{0.57} & \textbf{64.83}\sd{0.64} & \textbf{59.24}\sd{0.37} & \textbf{61.69}\sd{0.33} \\
\midrule\midrule
\rowcolor{hdr} & & gaus & shot & impu & defo & glas & moti & zoom & snow & fros & fog & brit & cont & elas & pixe & jpeg & Average \\
\rowcolor{hdr} & continual & \multicolumn{15}{c}{\streamarrow} & \\
\midrule\midrule
& TENT   & 4.17\sd{1.50} & 1.12\sd{0.06} & 1.03\sd{0.01} & 1.26\sd{0.16} & 1.00\sd{0.00} & 1.00\sd{0.00} & 1.01\sd{0.01} & 1.00\sd{0.00} & 1.00\sd{0.00} & 1.00\sd{0.00} & 1.01\sd{0.01} & 1.00\sd{0.00} & 1.00\sd{0.00} & 1.00\sd{0.00} & 1.00\sd{0.00} & 1.24\sd{0.11} \\
& EATA   & \underline{45.03}\sd{1.12} & \underline{43.64}\sd{1.45} & \underline{37.68}\sd{2.84} & 44.59\sd{2.21} & \underline{29.14}\sd{2.39} & \underline{37.51}\sd{3.55} & \underline{36.11}\sd{3.65} & \underline{27.68}\sd{5.59} & \underline{29.28}\sd{3.39} & \underline{19.86}\sd{2.71} & \underline{28.33}\sd{5.78} & 10.13\sd{4.99} & \underline{15.79}\sd{2.98} & \underline{19.05}\sd{3.49} & 14.41\sd{1.80} & \underline{29.22}\sd{1.37} \\
& SAR    & 24.12\sd{0.56} & 17.85\sd{6.93} & 14.46\sd{2.88} & \underline{44.81}\sd{8.92} & 12.82\sd{5.62} & 32.99\sd{10.34} & 31.20\sd{13.75} & 18.67\sd{10.20} & 15.17\sd{10.78} & 13.18\sd{8.79} & 24.97\sd{13.20} & \underline{13.86}\sd{8.59} & 12.69\sd{10.40} & 12.98\sd{5.77} & \underline{15.67}\sd{9.95} & 20.36\sd{7.07} \\
& DeYO   & 5.84\sd{1.78} & 4.19\sd{2.74} & 2.26\sd{0.86} & 2.53\sd{1.63} & 1.33\sd{0.37} & 1.41\sd{0.77} & 1.40\sd{0.40} & 1.26\sd{0.30} & 1.19\sd{0.25} & 1.02\sd{0.08} & 1.32\sd{0.27} & 1.04\sd{0.11} & 1.25\sd{0.29} & 1.22\sd{0.23} & 1.12\sd{0.09} & 1.89\sd{0.55} \\
& ReCAP  & 1.46\sd{0.14} & 1.09\sd{0.16} & 1.02\sd{0.04} & 1.25\sd{0.36} & 1.00\sd{0.01} & 0.99\sd{0.03} & 1.00\sd{0.00} & 1.04\sd{0.08} & 1.00\sd{0.00} & 1.00\sd{0.00} & 1.00\sd{0.01} & 1.00\sd{0.00} & 1.00\sd{0.00} & 1.00\sd{0.00} & 1.00\sd{0.00} & 1.06\sd{0.03} \\
& SaTeen & 1.34\sd{0.14} & 1.07\sd{0.17} & 1.01\sd{0.02} & 1.07\sd{0.16} & 1.05\sd{0.09} & 1.03\sd{0.04} & 1.00\sd{0.00} & 1.00\sd{0.00} & 1.00\sd{0.00} & 1.00\sd{0.00} & 1.00\sd{0.00} & 1.00\sd{0.00} & 1.00\sd{0.00} & 1.00\sd{0.00} & 1.00\sd{0.00} & 1.04\sd{0.01} \\
\rowcolor{grn}\cellcolor{hdr}\multirow{-7}{*}{\rotatebox[origin=c]{90}{\textit{shuffled}}} & OriTTA & \textbf{49.63}\sd{0.43} & \textbf{57.69}\sd{0.15} & \textbf{54.24}\sd{0.24} & \textbf{68.21}\sd{0.14} & \textbf{47.86}\sd{2.03} & \textbf{66.47}\sd{0.97} & \textbf{68.61}\sd{0.24} & \textbf{59.51}\sd{0.96} & \textbf{61.51}\sd{0.30} & \textbf{56.09}\sd{0.21} & \textbf{66.12}\sd{0.22} & \textbf{57.01}\sd{3.22} & \textbf{54.51}\sd{3.07} & \textbf{58.49}\sd{1.34} & \textbf{55.54}\sd{1.10} & \textbf{58.77}\sd{0.28} \\
\midrule
& TENT   & 3.28\sd{1.82} & 1.04\sd{0.02} & 1.01\sd{0.01} & 1.27\sd{0.10} & 1.00\sd{0.00} & 1.00\sd{0.00} & 1.01\sd{0.02} & 1.00\sd{0.00} & 1.00\sd{0.00} & 1.00\sd{0.00} & 1.01\sd{0.01} & 1.00\sd{0.00} & 1.00\sd{0.00} & 1.00\sd{0.00} & 1.00\sd{0.00} & 1.17\sd{0.12} \\
& EATA   & \underline{32.80}\sd{2.45} & \underline{32.90}\sd{1.82} & \underline{27.20}\sd{1.75} & 40.04\sd{2.30} & 20.23\sd{2.23} & 24.50\sd{4.31} & 20.46\sd{5.43} & 15.39\sd{3.11} & 11.82\sd{2.49} & 9.85\sd{1.35} & 12.39\sd{2.40} & 3.64\sd{2.34} & 7.09\sd{2.01} & 6.70\sd{3.15} & 5.75\sd{0.83} & 18.05\sd{0.68} \\
& SAR    & 27.36\sd{1.30} & 20.36\sd{10.95} & 18.34\sd{8.35} & \underline{57.65}\sd{5.89} & \underline{20.46}\sd{7.74} & \underline{45.66}\sd{10.03} & \underline{48.01}\sd{9.72} & \underline{29.63}\sd{9.02} & \underline{26.40}\sd{12.71} & \underline{23.65}\sd{6.62} & \underline{45.68}\sd{9.42} & \underline{22.24}\sd{6.54} & \underline{30.45}\sd{5.79} & \underline{24.36}\sd{7.63} & \underline{28.31}\sd{5.89} & \underline{31.24}\sd{6.53} \\
& DeYO   & 8.10\sd{2.17} & 2.71\sd{0.70} & 2.71\sd{1.75} & 1.95\sd{0.47} & 1.89\sd{0.32} & 1.14\sd{0.39} & 1.44\sd{0.15} & 1.22\sd{0.18} & 1.42\sd{0.28} & 1.13\sd{0.31} & 1.25\sd{0.38} & 1.00\sd{0.00} & 0.73\sd{0.37} & 0.88\sd{0.28} & 1.07\sd{0.13} & 1.91\sd{0.26} \\
& ReCAP  & 3.52\sd{1.23} & 0.83\sd{0.63} & 1.28\sd{0.41} & 2.23\sd{1.33} & 1.00\sd{0.00} & 1.00\sd{0.00} & 1.03\sd{0.06} & 1.26\sd{0.37} & 1.00\sd{0.00} & 1.00\sd{0.00} & 1.00\sd{0.00} & 1.00\sd{0.00} & 1.00\sd{0.00} & 1.00\sd{0.00} & 1.00\sd{0.01} & 1.28\sd{0.13} \\
& SaTeen & 3.54\sd{2.14} & 1.14\sd{0.29} & 1.09\sd{0.20} & 1.95\sd{1.14} & 1.18\sd{0.41} & 1.00\sd{0.01} & 1.00\sd{0.00} & 1.00\sd{0.00} & 1.00\sd{0.00} & 1.00\sd{0.00} & 1.73\sd{1.63} & 1.20\sd{0.46} & 1.04\sd{0.09} & 1.00\sd{0.00} & 1.00\sd{0.00} & 1.33\sd{0.23} \\
\rowcolor{grn}\cellcolor{hdr}\multirow{-7}{*}{\rotatebox[origin=c]{90}{\textit{class-blocked}}} & OriTTA & \textbf{48.07}\sd{0.52} & \textbf{55.99}\sd{0.22} & \textbf{50.84}\sd{0.64} & \textbf{67.25}\sd{0.49} & \textbf{46.65}\sd{1.90} & \textbf{66.65}\sd{1.63} & \textbf{68.11}\sd{1.04} & \textbf{57.46}\sd{0.51} & \textbf{57.90}\sd{2.04} & \textbf{53.88}\sd{2.33} & \textbf{66.55}\sd{1.83} & \textbf{55.22}\sd{2.84} & \textbf{56.51}\sd{0.97} & \textbf{50.87}\sd{3.63} & \textbf{53.47}\sd{1.87} & \textbf{57.03}\sd{0.29} \\
\bottomrule
\end{tabular}}
\end{sidewaystable}

\begin{sidewaystable}
\centering\scriptsize
\setlength{\tabcolsep}{2.5pt}
\caption{Full results on CIFAR-10-C, WRN-28-10.}
\label{tab:full_c10}
\resizebox{\linewidth}{!}{%
\begin{tabular}{>{\columncolor{hdr}}c l ccccccccccccccc | c}
\toprule
\rowcolor{hdr} & episodic & gaus & shot & impu & defo & glas & moti & zoom & snow & fros & fog & brit & cont & elas & pixe & jpeg & Average \\
\midrule\midrule
& TENT   & 11.46\sd{0.21} & 12.96\sd{0.68} & 10.94\sd{0.31} & 42.53\sd{27.44} & 16.73\sd{0.79} & \underline{72.56}\sd{17.58} & \underline{66.12}\sd{20.77} & \underline{77.13}\sd{1.20} & 46.58\sd{9.98} & \underline{79.70}\sd{5.12} & \underline{91.16}\sd{0.16} & 23.50\sd{16.37} & 70.43\sd{2.82} & 19.45\sd{2.81} & 59.32\sd{8.62} & 46.70\sd{2.77} \\
& EATA   & 28.07\sd{0.30} & 34.51\sd{0.16} & \underline{27.22}\sd{0.15} & \underline{53.38}\sd{0.25} & \underline{45.87}\sd{0.17} & 65.47\sd{0.20} & 58.17\sd{0.32} & 74.97\sd{0.08} & 58.73\sd{0.13} & 74.11\sd{0.23} & 90.69\sd{0.01} & 53.75\sd{0.34} & \underline{73.47}\sd{0.09} & 41.78\sd{0.26} & \underline{69.75}\sd{0.06} & \underline{56.66}\sd{0.08} \\
& SAR    & 27.14\sd{0.15} & 33.53\sd{0.15} & 25.93\sd{0.27} & 53.31\sd{0.36} & 45.46\sd{0.17} & 65.23\sd{0.10} & 58.07\sd{0.18} & 74.95\sd{0.04} & 58.80\sd{0.10} & 74.01\sd{0.04} & 90.69\sd{0.01} & 53.23\sd{0.27} & 73.46\sd{0.10} & 41.45\sd{0.12} & 69.72\sd{0.08} & 56.33\sd{0.03} \\
& DeYO   & \underline{34.65}\sd{5.30} & \underline{47.15}\sd{4.56} & 19.40\sd{5.32} & 47.17\sd{20.05} & 15.12\sd{1.74} & 46.03\sd{12.28} & 52.38\sd{18.63} & 75.40\sd{3.05} & \underline{63.55}\sd{3.70} & 63.39\sd{7.78} & 90.02\sd{0.43} & \underline{65.94}\sd{7.24} & 41.57\sd{3.30} & \underline{67.51}\sd{4.30} & 49.48\sd{8.06} & 51.92\sd{2.94} \\
& ReCAP  & 22.01\sd{1.86} & 28.92\sd{2.35} & 14.62\sd{1.46} & 40.20\sd{8.94} & 33.60\sd{1.56} & 59.14\sd{1.85} & 40.88\sd{6.91} & 73.64\sd{0.61} & 54.31\sd{1.37} & 66.44\sd{12.00} & 90.69\sd{0.04} & 32.83\sd{6.61} & 63.78\sd{2.24} & 47.77\sd{1.52} & 61.68\sd{3.21} & 48.70\sd{0.98} \\
& SaTeen & 26.22\sd{6.15} & 37.41\sd{8.06} & 11.68\sd{3.20} & 24.33\sd{14.70} & 13.22\sd{3.27} & 27.59\sd{11.22} & 31.10\sd{21.73} & 63.20\sd{5.28} & 57.05\sd{4.50} & 44.53\sd{8.85} & 83.06\sd{1.23} & 26.32\sd{19.35} & 26.12\sd{5.88} & 56.15\sd{4.18} & 32.33\sd{11.88} & 37.35\sd{2.55} \\
\rowcolor{grn}\cellcolor{hdr}\multirow{-7}{*}{\rotatebox[origin=c]{90}{\textit{shuffled}}} & OriTTA & \textbf{56.14}\sd{0.30} & \textbf{60.55}\sd{0.34} & \textbf{41.11}\sd{0.26} & \textbf{80.25}\sd{0.40} & \textbf{59.54}\sd{0.40} & \textbf{82.02}\sd{0.31} & \textbf{82.34}\sd{0.19} & \textbf{80.75}\sd{0.19} & \textbf{76.99}\sd{0.31} & \textbf{84.44}\sd{0.24} & \textbf{91.32}\sd{0.09} & \textbf{80.58}\sd{0.19} & \textbf{77.84}\sd{0.08} & \textbf{73.40}\sd{0.55} & \textbf{73.43}\sd{0.16} & \textbf{73.38}\sd{0.08} \\
\midrule
& TENT   & 12.75\sd{2.11} & 11.46\sd{1.94} & 9.12\sd{0.78} & 42.24\sd{26.23} & 11.59\sd{1.95} & \underline{71.42}\sd{19.76} & \underline{61.65}\sd{24.86} & 71.92\sd{15.59} & 53.95\sd{20.09} & 58.46\sd{30.29} & 91.52\sd{0.44} & 26.27\sd{18.82} & 67.76\sd{15.44} & 16.43\sd{7.70} & 46.34\sd{19.02} & 43.52\sd{6.84} \\
& EATA   & 28.18\sd{0.29} & 34.59\sd{0.14} & 27.22\sd{0.10} & 53.36\sd{0.24} & 45.82\sd{0.20} & 65.25\sd{0.17} & 58.17\sd{0.14} & 74.93\sd{0.05} & 58.89\sd{0.15} & 74.04\sd{0.13} & 90.70\sd{0.02} & 53.47\sd{0.27} & 73.42\sd{0.11} & 41.73\sd{0.14} & 69.73\sd{0.09} & 56.63\sd{0.05} \\
& SAR    & 27.47\sd{0.28} & 34.13\sd{0.77} & 27.54\sd{1.03} & \underline{54.38}\sd{0.83} & 46.11\sd{0.59} & 65.69\sd{0.40} & 59.06\sd{0.41} & 75.19\sd{0.04} & 59.35\sd{0.22} & \underline{74.23}\sd{0.15} & 90.71\sd{0.01} & 54.18\sd{0.52} & \underline{74.08}\sd{0.34} & 41.67\sd{0.40} & 70.58\sd{0.17} & 56.96\sd{0.10} \\
& DeYO   & 44.70\sd{15.15} & 43.05\sd{11.26} & 27.70\sd{11.34} & 50.02\sd{27.06} & 17.17\sd{1.96} & 56.28\sd{13.26} & 57.90\sd{10.31} & \textbf{81.65}\sd{5.34} & 69.83\sd{5.62} & 64.83\sd{10.60} & \underline{93.55}\sd{0.80} & \underline{64.57}\sd{20.32} & 58.34\sd{14.66} & \underline{75.73}\sd{8.21} & 54.79\sd{17.02} & 57.34\sd{6.14} \\
& ReCAP  & 40.34\sd{4.99} & 46.67\sd{4.97} & \textbf{43.52}\sd{9.94} & 48.09\sd{3.19} & \underline{46.84}\sd{2.32} & 63.66\sd{5.85} & 52.12\sd{2.22} & 77.23\sd{1.82} & 62.60\sd{2.49} & 67.24\sd{3.94} & 90.91\sd{0.10} & 46.04\sd{4.00} & 71.34\sd{3.25} & 51.54\sd{2.82} & \textbf{73.84}\sd{3.22} & \underline{58.80}\sd{0.69} \\
& SaTeen & \textbf{52.91}\sd{16.63} & \textbf{67.45}\sd{21.67} & 24.33\sd{9.99} & 29.35\sd{22.67} & 19.76\sd{10.98} & 41.56\sd{20.37} & 39.11\sd{27.30} & \underline{81.02}\sd{8.19} & \textbf{73.45}\sd{17.10} & 42.46\sd{17.97} & \textbf{95.29}\sd{1.87} & 38.16\sd{22.68} & 55.43\sd{27.35} & \textbf{77.96}\sd{11.96} & 55.26\sd{19.99} & 52.90\sd{8.45} \\
\rowcolor{grn}\cellcolor{hdr}\multirow{-7}{*}{\rotatebox[origin=c]{90}{\textit{class-blocked}}} & OriTTA & \underline{46.96}\sd{4.01} & \underline{52.50}\sd{2.63} & \underline{39.10}\sd{3.11} & \textbf{73.22}\sd{2.79} & \textbf{54.49}\sd{1.76} & \textbf{78.07}\sd{2.19} & \textbf{76.05}\sd{1.99} & 78.28\sd{1.83} & \underline{72.51}\sd{1.31} & \textbf{83.49}\sd{1.30} & 91.52\sd{0.76} & \textbf{75.95}\sd{1.52} & \textbf{77.83}\sd{1.48} & 61.40\sd{3.99} & \underline{73.03}\sd{2.02} & \textbf{68.96}\sd{1.51} \\
\midrule\midrule
\rowcolor{hdr} & & gaus & shot & impu & defo & glas & moti & zoom & snow & fros & fog & brit & cont & elas & pixe & jpeg & Average \\
\rowcolor{hdr} & continual & \multicolumn{15}{c}{\streamarrow} & \\
\midrule\midrule
& TENT   & 11.46\sd{0.21} & 10.29\sd{0.03} & 10.00\sd{0.00} & \textbf{77.19}\sd{2.54} & 12.07\sd{0.33} & 22.89\sd{8.62} & 10.95\sd{1.57} & 10.00\sd{0.00} & 10.00\sd{0.00} & 10.00\sd{0.00} & 10.08\sd{0.02} & 10.00\sd{0.00} & 10.00\sd{0.00} & 10.00\sd{0.00} & 10.00\sd{0.00} & 15.66\sd{0.77} \\
& EATA   & 28.07\sd{0.30} & 34.94\sd{0.51} & 27.61\sd{0.21} & 53.38\sd{0.39} & \underline{46.38}\sd{0.52} & \underline{65.59}\sd{0.32} & 58.65\sd{0.56} & \underline{75.29}\sd{0.18} & \underline{59.63}\sd{0.42} & \underline{74.69}\sd{0.41} & \textbf{90.68}\sd{0.06} & \underline{54.39}\sd{0.70} & \underline{73.78}\sd{0.34} & 41.45\sd{0.43} & \underline{69.86}\sd{0.12} & \underline{56.96}\sd{0.23} \\
& SAR    & 27.14\sd{0.15} & 32.70\sd{2.10} & 25.27\sd{1.34} & 53.84\sd{2.15} & 43.80\sd{2.88} & 65.12\sd{2.38} & \underline{59.04}\sd{3.24} & 74.62\sd{0.96} & 57.47\sd{2.09} & 73.85\sd{1.91} & \underline{90.53}\sd{0.18} & 53.13\sd{2.23} & 72.83\sd{1.53} & 41.20\sd{2.28} & 69.33\sd{0.51} & 55.99\sd{0.87} \\
& DeYO   & \underline{34.65}\sd{5.30} & \underline{54.65}\sd{5.92} & \underline{40.36}\sd{3.74} & 31.97\sd{14.39} & 22.38\sd{4.29} & 30.21\sd{13.29} & 45.23\sd{16.27} & 52.48\sd{7.69} & 53.61\sd{5.37} & 49.02\sd{4.92} & 75.18\sd{3.31} & 47.49\sd{10.60} & 38.11\sd{10.07} & \underline{47.95}\sd{5.02} & 43.03\sd{8.53} & 44.42\sd{3.45} \\
& ReCAP  & 22.01\sd{1.86} & 27.88\sd{5.76} & 16.98\sd{1.97} & 32.74\sd{9.29} & 24.03\sd{4.21} & 37.04\sd{14.12} & 35.13\sd{17.24} & 54.28\sd{13.00} & 41.13\sd{15.96} & 44.45\sd{12.35} & 74.68\sd{10.99} & 25.03\sd{10.80} & 33.37\sd{12.29} & 46.43\sd{7.22} & 33.55\sd{11.23} & 36.58\sd{6.71} \\
& SaTeen & 26.22\sd{6.15} & 35.52\sd{13.22} & 19.36\sd{7.04} & 14.52\sd{6.26} & 11.58\sd{1.49} & 10.71\sd{0.90} & 10.30\sd{0.52} & 12.77\sd{4.96} & 10.93\sd{2.19} & 10.90\sd{2.64} & 11.00\sd{1.74} & 10.09\sd{0.24} & 10.14\sd{0.33} & 10.08\sd{0.29} & 9.99\sd{0.02} & 14.27\sd{1.78} \\
\rowcolor{grn}\cellcolor{hdr}\multirow{-7}{*}{\rotatebox[origin=c]{90}{\textit{shuffled}}} & OriTTA & \textbf{56.14}\sd{0.29} & \textbf{69.72}\sd{0.12} & \textbf{53.20}\sd{0.25} & \underline{66.23}\sd{0.84} & \textbf{59.95}\sd{0.27} & \textbf{81.22}\sd{0.30} & \textbf{85.45}\sd{0.21} & \textbf{79.06}\sd{0.25} & \textbf{79.99}\sd{0.34} & \textbf{84.65}\sd{0.12} & 90.41\sd{0.20} & \textbf{81.02}\sd{0.56} & \textbf{75.06}\sd{0.41} & \textbf{69.83}\sd{0.70} & \textbf{74.24}\sd{0.50} & \textbf{73.74}\sd{0.17} \\
\midrule
& TENT   & 12.75\sd{2.11} & 10.35\sd{0.07} & 10.00\sd{0.00} & 21.09\sd{9.47} & 10.54\sd{0.56} & 10.38\sd{0.32} & 10.06\sd{0.05} & 10.29\sd{0.33} & 10.01\sd{0.02} & 10.01\sd{0.02} & 10.14\sd{0.06} & 10.00\sd{0.00} & 10.00\sd{0.00} & 10.01\sd{0.01} & 10.00\sd{0.00} & 11.04\sd{0.78} \\
& EATA   & 28.18\sd{0.29} & 35.27\sd{0.45} & 27.65\sd{0.13} & \underline{53.25}\sd{0.36} & \underline{46.77}\sd{0.45} & \underline{65.30}\sd{0.37} & \underline{58.44}\sd{0.52} & \textbf{75.54}\sd{0.13} & \underline{60.04}\sd{0.42} & \underline{74.80}\sd{0.60} & \textbf{90.71}\sd{0.07} & \underline{53.91}\sd{1.09} & \underline{73.54}\sd{0.53} & 41.21\sd{0.66} & \underline{69.47}\sd{0.19} & \underline{56.94}\sd{0.17} \\
& SAR    & 27.47\sd{0.28} & 33.00\sd{4.21} & 26.86\sd{2.98} & 52.95\sd{2.63} & 43.22\sd{5.70} & 63.18\sd{3.49} & 57.17\sd{2.53} & 73.70\sd{3.64} & 56.61\sd{5.95} & 72.39\sd{3.54} & \underline{90.30}\sd{0.33} & 51.30\sd{5.91} & 70.41\sd{5.25} & 44.48\sd{4.79} & 69.28\sd{1.89} & 55.49\sd{2.39} \\
& DeYO   & \underline{44.70}\sd{15.15} & 56.74\sd{16.70} & \underline{39.40}\sd{8.01} & 20.65\sd{9.89} & 18.43\sd{5.21} & 24.90\sd{9.97} & 25.23\sd{13.52} & 39.69\sd{13.64} & 46.34\sd{9.61} & 37.16\sd{12.91} & 65.04\sd{25.78} & 41.07\sd{24.26} & 30.72\sd{12.39} & 48.71\sd{26.66} & 29.12\sd{16.66} & 37.86\sd{6.62} \\
& ReCAP  & 40.34\sd{4.99} & 45.02\sd{6.64} & 37.07\sd{9.26} & 42.88\sd{13.44} & 26.43\sd{7.18} & 45.98\sd{20.48} & 35.15\sd{18.62} & 55.61\sd{9.83} & 43.49\sd{9.93} & 46.50\sd{14.40} & 74.83\sd{10.42} & 26.10\sd{18.86} & 38.38\sd{19.61} & \underline{50.64}\sd{16.78} & 38.57\sd{27.67} & 43.13\sd{9.33} \\
& SaTeen & \textbf{52.91}\sd{16.63} & \textbf{66.49}\sd{20.39} & 30.83\sd{22.13} & 14.06\sd{11.29} & 9.45\sd{1.45} & 13.29\sd{6.42} & 22.45\sd{19.27} & 31.24\sd{17.72} & 26.02\sd{25.60} & 23.81\sd{17.86} & 42.07\sd{39.52} & 15.46\sd{7.86} & 15.11\sd{9.14} & 31.13\sd{23.25} & 19.99\sd{10.02} & 27.62\sd{7.73} \\
\rowcolor{grn}\cellcolor{hdr}\multirow{-7}{*}{\rotatebox[origin=c]{90}{\textit{class-blocked}}} & OriTTA & 43.35\sd{12.11} & \underline{59.98}\sd{2.57} & \textbf{40.76}\sd{2.51} & \textbf{59.70}\sd{3.12} & \textbf{53.22}\sd{1.75} & \textbf{73.91}\sd{1.96} & \textbf{74.66}\sd{2.46} & \underline{75.17}\sd{1.32} & \textbf{70.08}\sd{2.86} & \textbf{79.38}\sd{3.24} & 87.96\sd{1.48} & \textbf{72.62}\sd{4.21} & \textbf{74.60}\sd{2.65} & \textbf{55.11}\sd{4.31} & \textbf{70.15}\sd{2.09} & \textbf{66.04}\sd{1.43} \\
\bottomrule
\end{tabular}}
\end{sidewaystable}

\begin{sidewaystable}
\centering\scriptsize
\setlength{\tabcolsep}{2.5pt}
\caption{Full results on ImageNet-C, ResNet-50-GN, episodic, at batch size $64$.}
\label{tab:full_bs64}
\resizebox{\linewidth}{!}{%
\begin{tabular}{>{\columncolor{hdr}}c l ccccccccccccccc | c}
\toprule
\rowcolor{hdr} & episodic & gaus & shot & impu & defo & glas & moti & zoom & snow & fros & fog & brit & cont & elas & pixe & jpeg & Average \\
\midrule\midrule
& TENT   & 24.05\sd{1.02} & 27.90\sd{0.49} & 25.13\sd{1.07} & 15.06\sd{0.15} & 9.76\sd{1.90} & 21.87\sd{0.38} & 22.51\sd{0.21} & 26.56\sd{0.71} & 33.03\sd{0.52} & 3.50\sd{0.15} & 69.88\sd{0.05} & 42.07\sd{0.10} & 10.89\sd{0.59} & 48.41\sd{0.14} & 54.49\sd{0.05} & 29.01\sd{0.24} \\
& EATA   & 36.81\sd{0.25} & 38.75\sd{0.29} & 38.15\sd{0.34} & \underline{28.37}\sd{0.47} & \underline{28.01}\sd{0.24} & 35.97\sd{0.20} & \underline{38.62}\sd{0.24} & 48.11\sd{0.47} & 47.90\sd{0.19} & \underline{54.08}\sd{0.30} & 70.41\sd{0.13} & 47.68\sd{0.25} & 40.92\sd{0.47} & 55.21\sd{0.29} & 56.16\sd{0.14} & 44.34\sd{0.10} \\
& SAR    & 32.05\sd{0.29} & 34.27\sd{0.29} & 33.34\sd{0.36} & 18.62\sd{0.25} & 18.84\sd{0.38} & 30.15\sd{0.29} & 30.69\sd{0.35} & 42.11\sd{0.60} & 43.30\sd{0.10} & 30.49\sd{20.61} & 70.31\sd{0.04} & 43.92\sd{0.10} & 16.37\sd{1.91} & 49.07\sd{0.17} & 55.36\sd{0.05} & 36.59\sd{1.41} \\
& DeYO   & 38.12\sd{0.20} & 40.49\sd{0.17} & 39.57\sd{0.37} & 20.82\sd{0.23} & 22.91\sd{0.22} & 35.52\sd{0.10} & 33.22\sd{1.75} & 48.88\sd{0.29} & 47.09\sd{0.18} & 26.25\sd{25.36} & 72.25\sd{0.07} & 48.15\sd{0.09} & 37.25\sd{0.31} & 54.02\sd{0.12} & 56.90\sd{0.04} & 41.43\sd{1.69} \\
& ReCAP  & 39.02\sd{0.26} & 41.18\sd{0.20} & 40.24\sd{0.27} & 22.57\sd{0.43} & 23.72\sd{0.47} & 37.67\sd{0.26} & 38.04\sd{0.45} & 50.19\sd{0.34} & 48.10\sd{0.20} & 42.72\sd{23.04} & \underline{72.60}\sd{0.05} & 48.93\sd{0.11} & 39.46\sd{1.35} & 54.85\sd{0.16} & 57.31\sd{0.04} & 43.77\sd{1.51} \\
& SaTeen & \textbf{43.10}\sd{0.24} & \textbf{45.32}\sd{0.13} & \textbf{44.26}\sd{0.14} & 24.52\sd{0.24} & 25.56\sd{0.25} & \underline{41.53}\sd{0.29} & 27.32\sd{9.30} & \underline{54.17}\sd{0.45} & \underline{51.90}\sd{0.28} & 33.29\sd{30.06} & \textbf{73.17}\sd{0.05} & \underline{52.78}\sd{0.19} & \underline{45.78}\sd{0.35} & \underline{58.86}\sd{0.23} & \underline{59.06}\sd{0.16} & \underline{45.37}\sd{1.89} \\
\rowcolor{grn}\cellcolor{hdr}\multirow{-7}{*}{\rotatebox[origin=c]{90}{\textit{shuffled}}} & OriTTA & \underline{41.68}\sd{0.19} & \underline{43.83}\sd{0.23} & \underline{42.65}\sd{0.26} & \textbf{34.58}\sd{0.30} & \textbf{33.38}\sd{0.26} & \textbf{43.15}\sd{0.16} & \textbf{46.82}\sd{0.22} & \textbf{54.67}\sd{0.30} & \textbf{53.24}\sd{0.20} & \textbf{57.84}\sd{0.34} & 72.44\sd{0.11} & \textbf{53.61}\sd{0.23} & \textbf{50.68}\sd{0.48} & \textbf{59.29}\sd{0.08} & \textbf{60.20}\sd{0.21} & \textbf{49.87}\sd{0.09} \\
\midrule
& TENT   & 24.11\sd{3.22} & 27.84\sd{2.15} & 25.75\sd{2.08} & 15.02\sd{0.26} & 11.00\sd{2.65} & 22.02\sd{0.90} & 22.90\sd{0.37} & 27.18\sd{0.99} & 33.30\sd{0.50} & 3.50\sd{0.31} & 69.85\sd{0.09} & 42.30\sd{0.19} & 11.36\sd{0.66} & 48.52\sd{0.19} & 54.54\sd{0.15} & 29.28\sd{0.55} \\
& EATA   & 33.09\sd{0.97} & 34.90\sd{0.55} & 33.67\sd{0.78} & 22.89\sd{0.32} & 24.00\sd{0.41} & 32.57\sd{0.28} & 34.62\sd{0.63} & 42.54\sd{1.19} & 42.63\sd{0.92} & \underline{49.34}\sd{1.06} & 68.97\sd{0.38} & 43.62\sd{0.55} & 34.86\sd{0.71} & 51.35\sd{0.42} & 53.52\sd{0.54} & 40.17\sd{0.21} \\
& SAR    & 32.62\sd{0.44} & 34.61\sd{0.32} & 33.57\sd{0.29} & 18.48\sd{0.42} & 18.64\sd{1.63} & 30.17\sd{0.32} & 30.44\sd{0.52} & 39.49\sd{1.18} & 43.27\sd{0.12} & 35.62\sd{17.48} & 70.29\sd{0.04} & 44.09\sd{0.19} & 13.74\sd{1.75} & 49.34\sd{0.32} & 55.42\sd{0.08} & 36.65\sd{1.17} \\
& DeYO   & 38.15\sd{0.29} & 40.41\sd{0.27} & 39.51\sd{0.26} & 20.37\sd{0.31} & 22.67\sd{0.34} & 35.42\sd{0.36} & 32.44\sd{3.68} & 48.09\sd{0.36} & 46.91\sd{0.34} & 26.58\sd{26.21} & 72.22\sd{0.08} & 48.17\sd{0.10} & 36.39\sd{0.73} & 53.95\sd{0.26} & 56.87\sd{0.13} & 41.21\sd{1.79} \\
& ReCAP  & 39.11\sd{0.23} & 41.04\sd{0.31} & 40.35\sd{0.43} & 21.72\sd{0.38} & 23.18\sd{0.79} & 37.38\sd{0.38} & \underline{35.99}\sd{4.20} & 49.36\sd{0.52} & 47.88\sd{0.39} & 27.48\sd{26.94} & \underline{72.59}\sd{0.07} & 48.98\sd{0.10} & 35.22\sd{7.66} & 54.82\sd{0.31} & 57.29\sd{0.13} & 42.16\sd{2.08} \\
& SaTeen & \textbf{42.60}\sd{0.20} & \textbf{44.71}\sd{0.16} & \textbf{43.83}\sd{0.31} & \underline{23.94}\sd{0.43} & \underline{25.04}\sd{0.59} & \underline{40.97}\sd{0.34} & 25.98\sd{9.97} & \underline{53.68}\sd{0.43} & \underline{51.65}\sd{0.16} & 36.92\sd{29.59} & \textbf{73.24}\sd{0.10} & \textbf{52.51}\sd{0.16} & \underline{45.44}\sd{0.75} & \textbf{58.80}\sd{0.35} & \underline{59.14}\sd{0.12} & \underline{45.23}\sd{2.07} \\
\rowcolor{grn}\cellcolor{hdr}\multirow{-7}{*}{\rotatebox[origin=c]{90}{\textit{class-blocked}}} & OriTTA & \underline{40.66}\sd{0.26} & \underline{42.50}\sd{0.32} & \underline{41.64}\sd{0.27} & \textbf{33.24}\sd{0.33} & \textbf{31.05}\sd{0.43} & \textbf{41.38}\sd{0.37} & \textbf{45.15}\sd{0.32} & \textbf{54.02}\sd{0.16} & \textbf{52.64}\sd{0.10} & \textbf{55.77}\sd{0.35} & 72.31\sd{0.19} & \underline{52.39}\sd{0.30} & \textbf{48.59}\sd{0.18} & \underline{57.36}\sd{0.22} & \textbf{59.76}\sd{0.21} & \textbf{48.56}\sd{0.14} \\
\bottomrule
\end{tabular}}
\end{sidewaystable}

\end{document}